\documentclass[conference]{IEEEtran}
\pdfoutput=1 

\author{\IEEEauthorblockN{Mercedes Garcia-Salguero\IEEEauthorrefmark{1} 
and Javier Gonzalez-Jimenez\IEEEauthorrefmark{2}} \\
\IEEEauthorblockA{Machine Perception and Intelligent Robotics (MAPIR) Group, System Engineering and Automation Department, \\ 
University of Malaga, Campus de Teatinos, 29071 Malaga, Spain \\
Email: \IEEEauthorrefmark{1}mercedesgarsal@uma.es,
\IEEEauthorrefmark{2}javiergonzalez@uma.es}}

\usepackage{multirow}

\usepackage{amsmath,amssymb,amsfonts, amsthm, bm}
\usepackage{mathtools}
 
\usepackage{graphicx}
\usepackage{comment}
\usepackage{xcolor}

\usepackage{comment}

\usepackage{cite}
\usepackage{mathtools}
\usepackage{epsfig}
\usepackage{algorithmic}
\usepackage{textcomp}
\usepackage{subcaption}
\usepackage[linesnumbered,ruled,vlined]{algorithm2e}

\usepackage{GeneralNotation}

\newcommand{\vHatLambda}{\hat{\bm{\lambda}}}

\newcommand{\dataMatrixQ}{\matQ }

\newcommand{\fRt}{f(\rot, \trans)}
\newcommand{\fE}{f(\boldsymbol{E})}

\newcommand{\obsi}{\boldsymbol{f}_i}
\newcommand{\obsip}{\boldsymbol{f}_i'}

\newcommand{\trans}{\boldsymbol{t}}

\newcommand{\Essential}{\boldsymbol{E}}

\newcommand{\Me}{\mathcal{M}_{\matE}} 

\newcommand{\maxParallax}{||\trans||_{\text{max}}\xspace}

\newcommand{\cpp}{\textsc{C++}}

\newcommand{\RPp}{RPp}
\newcommand{\RPpExt}{Relative Pose problem}
\newcommand{\eightpt}{\textsc{8-pt} \xspace}
\newcommand{\sevenpt}{\textsc{7-pt} \xspace}
\newcommand{\fivept}{\textsc{5-pt} \xspace}
\newcommand{\opengv}{\textsc{opengv} \xspace}

\newcommand{\rotSphereSpace}{\rotSpace \times \sphere}

\newcommand{\EIG}{\textsc{EIG}}
\newcommand{\BA}{\textsc{BA}}
\newcommand{\BARED}{\textsc{BARED}}
\newcommand{\OURS}{\textsc{OURS}}
\newcommand{\ETH}{\textsc{ETH3D}}
\newcommand{\TUM}{\textsc{TUM}}
\newcommand{\CVPR}{\textsc{CVPR08}}
\newcommand{\OURSCERT}{\textsc{OURS-CERT}}
\newcommand{\FIVER}{\textsc{RANSAC-5PT}}
\newcommand{\SEVENR}{\textsc{RANSAC-7PT}}
\newcommand{\EIGHTR}{\textsc{RANSAC-8PT}}

\newcommand{\opLoss}{\rho}
\newcommand{\opLossC}{\rho_{\overline{c}}}
\newcommand{\opGNC}{\rho_{ \mu}}

\newcommand{\outlierProc}{\Psi_{\overline{c}}(w_i)}
\newcommand{\outlierProcMu}{\Psi_{\overline{c}}(w_i, \mu)}

\newcommand{\cSq}{\overline{c}^2}

\providecommand{\vCross}[1]{\ensuremath{\operatorname{vec}([{#1}]_{\times})}}

\newcommand{\gradfRt}{ \nabla  \fRt}
\newcommand{\gradfR}{ \nabla_{\rot} \fRt}
\newcommand{\gradft}{ \nabla_{\trans} \fRt}

\newcommand{\RGradfRt}{ \text{grad} \fRt}
\newcommand{\RGradfR}{ \text{grad}_{\rot} \fRt}
\newcommand{\RGradft}{ \text{grad}_{\trans} \fRt}

\newcommand{\ProjRt}{ \text{Proj}_{\rotSphereSpace}}
\newcommand{\ProjR}{ \text{Proj}_{\rot}}
\newcommand{\Projt}{ \text{Proj}_{\trans}}

\newcommand{\TM}{ \textit{TM}}
\newcommand{\TanR}{ \text{T}_{\rot}}
\newcommand{\Tant}{ \text{T}_{\trans}}

\newcommand{\HessfRt}{\nabla^2 \fRt}
\newcommand{\HessfR}{\nabla^2_{\rot} \fRt}
\newcommand{\HessRR}{\nabla_{\rot \rot} \fRt}
\newcommand{\HessRt}{\nabla_{\rot \trans} \fRt}

\newcommand{\Hessft}{\nabla^2_{\trans} \fRt}
\newcommand{\Hesstt}{\nabla_{\trans \trans} \fRt}
\newcommand{\HesstR}{\nabla_{ \trans \rot} \fRt}

\newcommand{\RHessRt}{\text{Hess}_{\rot, \trans}}
\newcommand{\RHessR}{\text{Hess}_{\rot}}
\newcommand{\RHesst}{\text{Hess}_{\trans}}

\newcommand{\RHessfRt}{\RHessRt \fRt}
\newcommand{\RHessfR}{\RHessR \fRt}
\newcommand{\RHessft}{\RHesst \fRt}

\newcommand{\HessLambda}{\matM(\bm{\lambda})}
\newcommand{\HessLambdaHat}{\matM\bm{(\hat{\lambda})}}
\newcommand{\VR}{\dot{\rot}}
\newcommand{\Vr}{\dot{\rVec}}
\newcommand{\Vt}{\dot{\trans}}

\newcommand{\fHatE}{ f(\hat{\Essential})}
\newcommand{\dHatR}{ d_{\text{R}}(\vHatLambda)}
\newcommand{\fHatR}{ f_{\text{R}}(\vHatX)}
\newcommand{\dOptR}{ d^{\star}_{\text{R}}}
\newcommand{\fOptR}{ f ^{\star}_{\text{R}}}

\newcommand{\SetEssentialMatrices}{\mathbb{E}} 
\newcommand{\SetRelaxedEssentialMatrices}{\mathbb{E}_\text{R}}

\newcommand{\constrSet}{\mathcal{C}}
\newcommand{\constrSetRelaxed}{\constrSet_{\text{R}}}

\newcommand{\vHatX}{\hat{\boldsymbol{x}}}
\newcommand{\hatE}{\hat{\Essential}}
\newcommand{\hatLambda}{\hat{\lambda}}

\newcommand{\HessE}{{\matM^{\eVec} \bm{(\hat{\lambda})}}}
\newcommand{\HessT}{{\matM^{\trans} \bm{(\hat{\lambda})}}}

\newcommand{\Mt}{\matM_{\trans}}
\newcommand{\Mr}{\matM_{\rot}}
\newcommand{\Mtr}{\matM_{\trans. \rot}}
\newcommand{\Mrt}{\matM_{\rot. \trans}}

\newcommand{\MtrRT}{\Mtr(\rot,\trans)}
\newcommand{\MrtRT}{\Mrt(\rot,\trans)}

\newcommand{\tauGap}{\tau_{\text{gap}}}
\newcommand{\tauMu}{\tau_{\mu}}

\renewcommand{\suppl}{\emph{Online Resource}}

\usepackage{hyperref}
\hypersetup{
    colorlinks=true,
    linkcolor=blue,
    filecolor=magenta,      
    urlcolor=cyan,
}

\begin{document}

\title{Fast and Robust Certifiable Estimation 
of the Relative Pose Between Two Calibrated Cameras
}

\maketitle
\begin{abstract}

This work contributes an efficient algorithm to compute the Relative Pose problem (RPp) between calibrated cameras and certify the optimality of the solution, 
given a set of pair-wise feature correspondences affected by noise and probably corrupted by wrong matches. 
We propose a family of certifiers that is shown to increase the ratio of 
detected optimal solutions. This set of certifiers is incorporated  into a fast essential matrix estimation pipeline that, given any initial guess for the RPp, refines it iteratively on the product space of 3D rotations and 2-sphere. 
In addition, this fast certifiable pipeline is integrated into a robust framework that combines
Graduated Non-convexity and the Black-Rangarajan duality 
between robust functions and line processes. 
 
We proved through extensive experiments on 
synthetic and real data 
that the proposed framework  
provides a fast and robust relative pose estimation. 
We make the code publicly available 
\url{https://github.com/mergarsal/FastCertRelPose.git}.

\end{abstract}

\section{Introduction}
\begin{figure}[t]
    \centering
    \includegraphics[width=0.9\linewidth]{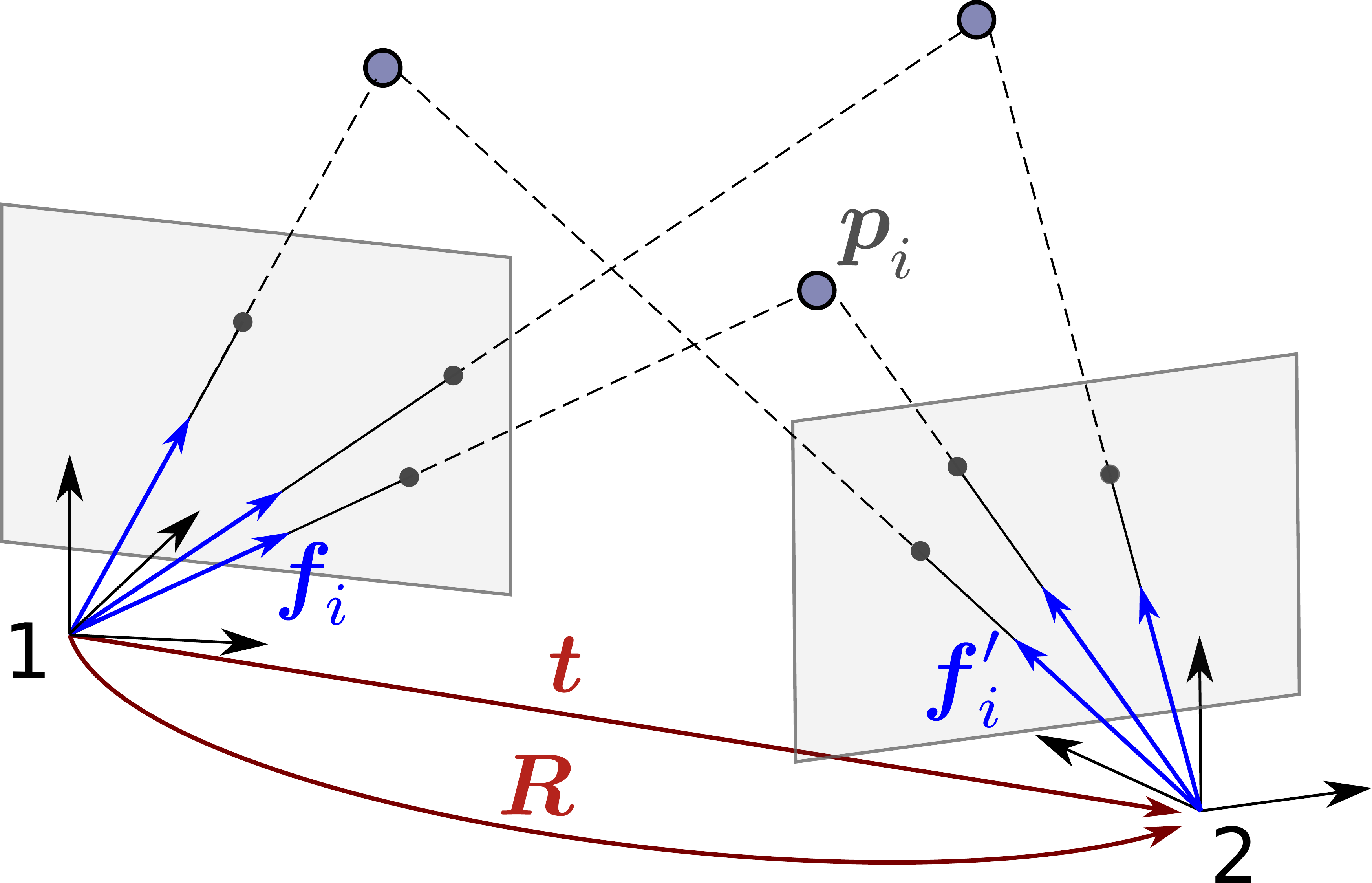}
    \caption{The \RPp\xspace seeks the relative pose, 
    orientation $\rot$ 
    and translation $\trans$ \uts 
    between two calibrated, central cameras 
    given a set of $N$ pair-wise feature correspondences. 
    Figure from~\cite{garcia2020certifiable}.}
    \label{fig:def-relpose}
\end{figure}

The \RPpExt \xspace(\RPp) consists of 
finding the relative rotation $\rot$ 
and translation $\trans$ 
between two central, 
calibrated cameras 
given a set of $N$ pair-wise feature correspondences 
$(\obsi, \obsip)$, 
as shown in Figure~\eqref{fig:def-relpose}. 
Since the scale cannot be recovered 
for this type of configurations, 
the translation component 
is estimated only \uts~\cite{hartley2003multiple}.

\begin{figure}[b]
    \centering
    \includegraphics[width=0.9\linewidth]{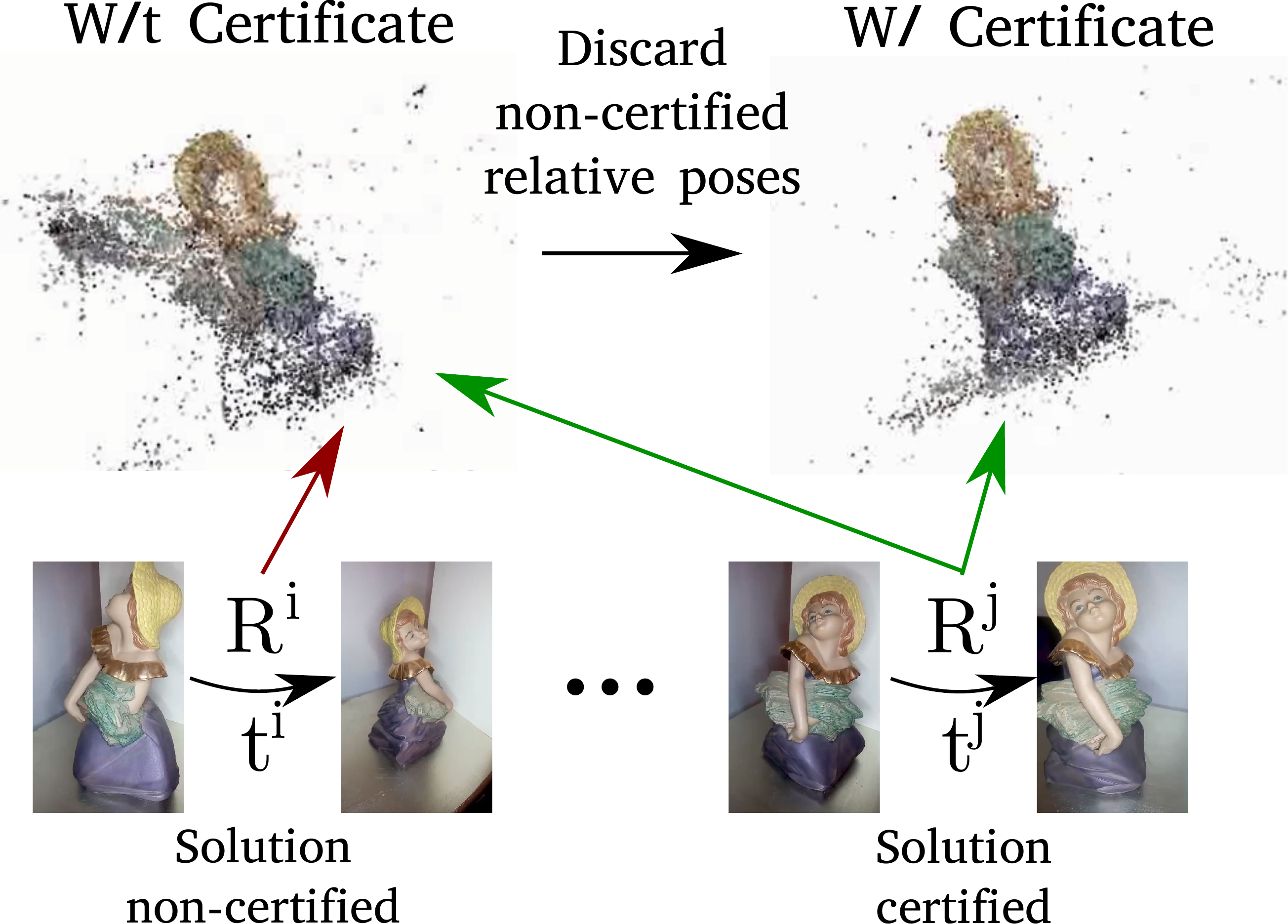}
    \caption{Suboptimal solutions for the \RPp~
    may lead to inaccurate solutions 
    or even the complete failure of those applications 
    that employed the \RPp~as fundamental block. 
    The left reconstruction 
    (\textsc{w/t certificate}) 
    is performed with all the solutions 
    for the \RPp, 
    while the right reconstruction 
    (\textsc{w/ certificate}) 
    only employs 
    those solutions certified as optimal. 
    Some of the images employed for the reconstruction 
    are included below. }
    \label{fig:intro-doll}
\end{figure}

Estimating the relative pose 
from visual data 
is the cornerstone of tasks such as visual odometry
~\cite{gomez2016robust, nister2004visual, scaramuzza2011visual}, 
and more complex computer vision problems 
\eg, Simultaneous Localization and Mapping (SLAM)
~\cite{taketomi2017visual, kerl2013dense, gomez2019pl} 
and Structure-from-Motion (SfM)
~\cite{schonberger2016structure, wu2013towards, westoby2012structure}. 
The gold-standard approach for \RPp~
poses it as a 2-view Bundle Adjustment (BA) 
that minimizes the reprojection error
~\cite{hartley2003multiple, kneip2013direct, hartley2007global}. 
This is a non-convex problem that typically presents 
multiple local minima. 
Since the algorithms employed for its resolution 
are iterative, 
they are prone to get stuck in suboptimal solutions, 
which may lay arbitrarily far away from 
the global minimum. 
The lack of quality of these solutions 
may hinder the subsequent blocks 
of the applications that leverage them, 
leading to inaccurate solutions 
or even to the complete failure of the application. 
An example of the last case is shown in Figure~\eqref{fig:intro-doll}, 
where the left reconstruction employs all the solutions, 
while the right reconstruction \emph{only} employed 
those solutions certified as global optima 
(we return to this certification later on this introduction). 
Therefore, special attention must be given to this 
fundamental and early block on these complex applications. 

In order to try to avoid these suboptimal solutions 
for the \RPp, a good initialization is usually required
~\cite{spetsakis1992optimal}. 
If this initial guess is within the attraction basin 
of the optimal solution, 
the iterative method will return it with high probability
~\cite{kneip2013direct}. 
Although, in general, the quality of the initial guess 
cannot be measured, 
it has been reported in the literature that 
simplifications of the problem 
lead to good initializations. 
A common simplification for the \RPp~
relies on 
the algebraic error and the so-called 
\textit{essential matrix} $\Essential$~\cite{hartley2003multiple}, 
which is a $3 \times 3$ matrix 
that encapsulates all the geometric information 
about the relative pose (rotation and translation) 
and it is defined as 
$\Essential \defexp \cross{t} \rot$, 
where $\cross{t}$ denotes 
the matrix form for the cross-product 
for a $3$D vector such that 
$\trans \times \bullet = \cross{t} \bullet$ 
(we define its explicit form later 
in Equation \eqref{eq:cross-t}). 

The essential matrix constrains 
the pair-wise correspondences 
$(\obsi, \obsip)$ 
via the \emph{epipolar constraint}, 
formally defined as 
$\obsi^T \Essential \obsip = 0$. 
Here the observations $\obsi, \obsip$ 
are 3D vectors, 
which can be expressed in homogeneous coordinates 
(last entry to one)  
or normalized 
(Euclidean norm to one). 
While this relation holds 
for all the correspondences 
when they are noiseless, 
in real scenarios 
$\obsi^T \Essential \obsip = \epsilon_i$, 
with $\epsilon_i \neq 0$, 
usually known as the \emph{epipolar error}. 
A common approach leverages this relation 
in order to find the ``best`` essential matrix 
in terms of this error, 
\ie 
we seek the $\Essential$ that 
minimizes $\epsilon_i^2$ for all the correspondences, 
$\sum_{i=1}^N \epsilon_i^2$~\cite{hartley2003multiple, Zhao2019}. 
Although more complex errors are employed in the literature, 
see for example~\cite{ma2001optimization, helmke2007essential}, 
this is already a complex, non-convex problem, 
hence presenting multiple local minima. 

The relative pose, 
and hence the essential matrix, 
has five degrees of freedom, 
three for the 3D rotation, three for the 3D translation 
and one less due to the scale ambiguity. 
Since the epipolar constraint is linear 
in the entries of $\Essential$, 
at least five correspondences are required to estimate it 
(except for degenerate cases~\cite{hartley2003multiple, decker2008dealing}). 
This is known as the minimal solver 
(the five-point algorithm, \fivept)
~\cite{nister2004efficient, stewenius2006recent}. 
This solver can be embedded into robust paradigms, 
such as RANSAC
~\cite{lui2013iterative,botterill2011refining}, 
in order to gain robustness against wrong correspondences, 
\ie outliers.

Although the \fivept is the minimal solver for 
this problem, 
another common approach is the eight-point algorithm, \eightpt
~\cite{faugeras1990motion, hartley2003multiple, ma2012invitation}. 
This method employs eight points, 
and although it was initially devised for the 
\emph{fundamental matrix}, 
the equivalent version of the essential matrix for 
uncalibrated cameras, 
it can be adapted to the calibrated case in a straightforward manner. 
The main advantage of this solver is its simplicity, 
since the solution is obtained by 
first computing the Singular Value Decomposition 
of the data matrix 
(formed by the pair-wise correspondences) and 
then projecting the 
right singular vector associated 
with the smallest singular value  
onto the space of essential matrices~\cite{hartley2003multiple, ma2012invitation}. 
Further, this approach admits an extension 
that includes all the correspondences, 
known as the Direct Linear Transformation (DLT). 
However, the solution found by these methods 
is only the global optimum in theory  
when correspondences are noiseless
~\cite{lee2020geometric}. 
These (suboptimal) solutions 
from the closed-form solvers  
are usually refined 
by iterative algorithms  
that respect the intrinsic constraints 
of the essential matrices, 
see~\eg~\cite{tron2017space, garcia2020certifiable, ma2001optimization}. 
However, since the problem is still non-convex, 
these iterative algorithms may still return suboptimal solutions
~\cite{spetsakis1992optimal}. 

Some works have proposed certifiable solvers 
for the non-minimal \RPp, 
that is, 
solvers that \textit{guarantee} 
that the returned solution  is 
the global optimum, 
among others~\cite{briales2018certifiably, Zhao2019, hartley2007global}. 
They make use of off-the-shelf optimization tools  
that have polynomial or even exponential time complexity. 
Although fast implementations has been reported, 
see~\cite{Zhao2019}, 
they are still slow for the computer vision standards ($30$ fps). 
On the bright side, 
these solvers are not the only ones that are able to certify optimality. 
Banderia~\cite{bandeira2016note} characterized the so-called Certifiable Algorithms 
that, for a given solution to a non-convex problem 
(obtained by any means) 
and for most real-world problem instances, 
are able to certify if the solution is optimal, 
provided some conditions are fulfilled. 
Such certifiable algorithms have been reported previously 
in the literature for other problems, 
such as rotation averaging~\cite{eriksson2018rotation}, 
point set registration
~\cite{iglesias2020global, briales2017convex} or 
pose graph optimization
~\cite{briales2016fast, carlone2015lagrangian, carlone2015duality}.
The main advantage of these approaches 
is that they are usually faster 
in terms of certification than 
the above-mentioned certifiable proposals. 
Further, since the solution to the non-convex problem 
can be estimated by any iterative method, 
the estimation stage is 
also faster than the exponential tools, 
see \eg~\cite{tron2017space}.
%
%
In~\cite{garcia2020certifiable}, 
we proposed for the first time 
this kind of certifiable algorithm for the \RPp, which was demonstrated to be faster than 
the state-of-the-art solver 
under a \matlab~implementation.

Nevertheless, 
the above-mentioned approaches 
(with some exceptions, like~\cite{Zhao2019}) 
do not consider explicitly the 
presence of bad correspondences (outliers), 
which strongly affect the accuracy of the solution. 
Outliers are common in real-world problem instances 
due to the nature of the 
feature detection algorithms 
and matching processes. 
The classic approach to detect these bad correspondences 
relies on 
the embedding of a minimal solver 
into a RANSAC framework
~\cite{botterill2011refining, kneip2013direct, Ra2008RANSAC}. 
The returned set of inliers 
is then employed to estimate the 
essential matrix 
by any of the previously mentioned methods. 
To overcome some of the limitations that 
such strategies present, 
attention has been recently directed to 
tools known as Graduated Non-convexity (GNC) ~\cite{blake1987visual} 
and the Black-Rangarajan (BR)~\cite{black1996unification} 
duality between robust estimations and line processes. 
In combination, they permit to 
estimate both the set of inliers and 
the solution to a given (non-convex) problem. 
The potential of these tools 
has been reported in, 
for example, 
~\cite{Zhao2019, yang2020teaser, yang2020graduated},
demonstrating to outperform RANSAC-based strategies in accuracy  
while being robust up to a higher percentage of outliers.

\smallskip 
\textbf{Contributions}: 
In this work, 
we extend our proposal in~\cite{garcia2020certifiable} 
for the \RPp \xspace 
with two significant improvements: 
a family of optimality certifiers  
and a robust paradigm for the non-minimal solver 
based on GNC and BR. 

More precisely, the main technical contributions of the present work are: 
\begin{enumerate}
 \item We define in Section~\eqref{sec:proposed-pipeline} 
    the set of optimality certifiers associated 
    with the different relaxations 
    of the set of essential matrices. 
    This family of certifiers is shown 
    to detect (certify) more optimal solutions, 
    overcoming 
    in most problem instances 
    the limitations associated 
    with the different relaxations 
    and their tightness.
    \item We provide in Section~\eqref{sec:essential-manifold} 
    the quadratic Euclidean model of the 
    \RPp~whose objective function is the squared, normalized epipolar error, 
    defined over 
    the product space of 3D rotation matrices and the 2-sphere. 
    A preconditioner that decreases the convergence time 
    of the iterative method is also provided. 
    \item We illustrate in Section~\eqref{sec:GNC} 
    the embedding of the \RPp~into the robust framework
    formed by the combination of GNC and BR, 
     providing also a general library for its utilization 
    (not restricted to the \RPp) in 
    \url{https://github.com/mergarsal/GNCSO.git}. 
    \item We perform extensive experiments in Section~\eqref{sec:experiments} 
    on both synthetic and real data 
    and compare our proposal against the state-of-the-art solvers 
    under a wide variety of configurations. 
    We also report the influence of our method in the performance of 2-view BA 
    against the common initializations. 
\end{enumerate}
As an additional contribution, 
we release the code for this version of our work in \\
\url{https://github.com/mergarsal/FastCertRelPose.git}. 

Notice that, although we show in Section~\eqref{sec:experiments} 
that strong duality usually holds for the \RPp~
and the chosen relaxations 
(from which our certifiers are derived), 
a formal proof of this behavior 
is not available yet. 

Last, we want to point out that while our proposal 
estimates the essential matrix, 
the relative rotation and translation \uts 
can be extracted from it 
by any classic computer vision algorithm~\cite{hartley2003multiple}.

\section{Related Work} 
\label{sec:related}
Here we summarized briefly the solvers proposed in the literature 
for the Relative Pose problem. 
For a more detailed list, 
we refer the reader to~\cite{garcia2020certifiable}.

\subsection{Non-certifiable approaches}

\smallskip 
\textbf{Closed-form solvers}

The essential matrix has five degrees of freedom 
(three from 3D rotation, three from 3D translation 
and one less because the scale ambiguity) 
and therefore, only five correspondences 
(except for degenerate cases~\cite{hartley2003multiple}) 
are required for its estimation. 
This is the so-called minimal problem and 
since it provides us with an efficient hypothesis generator, 
it can be embedded into a RANSAC framework to gain robustness against wrong correspondences, 
\ie outliers~\cite{kneip2013direct, botterill2011refining}. 
However, this minimal problem involves the resolution of a 
nontrivial (tenth degree) polynomial~\cite{stewenius2006recent, nister2004efficient}, 
and the tools to solve it 
(\eg polynomial ideal theory 
or Gröbner basis) 
are not always numerically stable. 
Hence, some previous works have proposed alternative approaches, 
see \eg~\cite{kukelova2007two, kukelova2008polynomial}.

In post of simplicity, the $8$-point algorithm by Faugeras \etal~\cite{faugeras1990motion}, 
which was originally proposed for the fundamental matrix, 
is usually adapted to the case of the essential matrix, 
\ie calibrated cameras. 
Of significant importance is 
the extension of the $8$-point algorithm 
that includes all the correspondences, 
usually known as the 
Direct Linear Transform (DLT)
~\cite{hartley2003multiple}.

\smallskip 
\textbf{Iterative solvers}

The above-mentioned solvers typically provide suboptimal solutions for the non-minimal N-point problem and 
therefore it is a common practice to refine these initial estimates by local, iterative methods~\cite{botterill2011refining}. 
Contrary to optimization problems on flat (Euclidean) spaces, these local optimization methods must respect the intrinsic constraints of the search space, 
\ie the solution must be always an essential matrix. 
Different parameterizations for the set of essential matrices have been proposed, 
see~\eg~\cite{lui2013iterative, ma2001optimization, helmke2007essential, tron2017space}. 
As it was shown in \cite{helmke2007essential}, 
these parameterizations may lead to different performances 
and convergence rates for non-linear optimization methods. 
The main difficulty relies on the peculiarities of the epipolar constraints 
and the symmetry between the two views, 
which are often not taking into account. 
The characterization proposed in~\cite{tron2017space} 
leverages the concept of quotient manifold~\cite{absil2009optimization} 
in order to model all these peculiarities.  
Nevertheless, simple characterizations 
(\eg, the product space of 3D rotations and 2-sphere) 
may still work in most 
problem instances under some conditions, 
for example, a good initial guess.

\subsection{Certifiable approaches}

\textbf{Branch-and-Bound}

Despite its attractiveness as fast solvers, 
the above-mentioned methods do not guarantee 
nor certify if the retrieved solution is optimal. 
In fact, finding said optimal solutions with guarantees for non-convex problems, 
such as the Relative Pose problem, is in general a hard task. 
Nonetheless, approaches with 
exponential time complexity in worst-case scenarios 
that rely on Branch-and-Bound global optimization 
have been proposed for that.
In \cite{yang2014optimal}, 
the authors incorporated the presence of outliers 
as an inlier-set maximization problem. 
In \cite{hartley2007global}, 
it was first proposed the estimation of the 
essential matrix under a $L_{\infty}$ cost function.
In \cite{kneip2013direct}, 
an eigenvalue formulation equivalent to the algebraic error was proposed 
and solved in practice by an efficient Levenberg-Marquardt scheme 
and a globally optimal Branch-and-Bound.
However, Branch-and-Bound has a slow performance.

\smallskip
\textbf{Convex Relaxation}

A different approach that also certifies the optimality of the solution  
consists of the re-formulation of the original problem 
as a Quadratically Constrained Quadratic Program (QCQP).
These problems are still NP-hard to solve in general; 
nonetheless we can relax these QCQPs 
into Semidefinite Relaxation Programs (SDPs) 
that can be actually solved in polynomial time 
by off-the-shelf tools, 
like SDPT3~\cite{toh1999sdpt3} or 
SeDuMi~\cite{sturm1999using}. 

One of these convex relaxations 
that 
have been shown 
to perform well in practice 
is known as 
the Shor's relaxation~\cite{boyd2004convex}. 
If the relaxation happens to be tight, 
one can recover the solution to the original problem 
with an optimality certificate. 
This relaxation has been shown to 
remain tight for 
a wide variety of problems 
under some conditions and noise regimens, 
for example 
in generalized essential matrix estimation~\cite{zhao2020certifiably}, 
$n$-view triangulation~\cite{aholt2012qcqp} or 
the Wahba problem with outliers~\cite{yang2019quaternion}, 
among others.

This was the approach followed 
in~\cite{briales2018certifiably}, 
in which the authors posed the \RPp\xspace
as an optimization problem that minimizes 
an objective function equivalent to the squared epipolar error 
directly over the rotation and translation spaces. 
In~\cite{Zhao2019} 
the author proposed another Shor's relaxation 
for the Relative Pose problem 
that minimizes the squared epipolar error 
but over the set of normalized essential matrices, 
leading to a small relaxation that 
was solved in roughly $6~ms$ under a \cpp~implementation.  
Although these approaches are faster than those 
based on BnB, 
they are still slow for the computer vision standard.

\smallskip 
\textbf{Certifiable algorithms}

Although tractable, solving these convex problems 
from scratch may not be the most efficient way to address them.  
As an alternative approach one may found the so-called Fast Certifiable Algorithms
recently characterized and motivated in \cite{bandeira2016note}. 
These algorithms typically leverage the existence of an optimality certifier
which, given a solution obtained by any means,
may be able to certify its optimality.
A straightforward approach to get that certificate 
is through the resolution of the \textit{dual problem}~\cite{boyd2004convex} from scratch, 
whose optimal cost value always provides a lower bound 
on the optimal objective for the original problem. 
In many real-world problem instances this bound is \emph{tight}, 
meaning both cost values are the same up to some accuracy and 
one can certify optimality from it. 
Under some conditions, 
we can even extract the solution to our original problem 
from the solution of
the dual, 
see~\eg~\cite{giamou2019certifiably} for extrinsic calibration of sensors 
or \cite{briales2017convex} for registration of points, lines and planes. 
However, this naive approach would still be as slow 
as directly solving the problem via its convex relaxation, 
since for the \RPp\xspace as posed in~\cite{Zhao2019, briales2018certifiably} 
both problems are instances of Positive Semidefinite problems (SDP) 
which can be solved up to arbitrary accuracy in polynomial time.

However, it is also possible to leverage the dual problem  
to derive a faster certification algorithm. 
This approach usually leads to a closed-form expression 
for dual candidates, 
that is, 
points that are feasible for the dual problem 
(within the domain of the dual function). 
This expression depends on the optimal solution 
for the original (primal) problem, 
and 
if the candidate is indeed feasible, 
it allows us to certify the optimality of the original 
solution. 
For QCQP, 
the feasibility of the dual candidates 
is reduced to a spectral analysis of the Hessian 
of the Lagrangian~\cite{boyd2004convex}. 
This has been the approach followed 
for other problems, 
for example, 
rotation averaging~\cite{eriksson2018rotation}, 
point cloud registration with missing data~\cite{iglesias2020global} 
and outliers~\cite{yang2020teaser}, 
pose graph optimization~\cite{briales2017cartan, rosen2019se} or 
SLAM~\cite{briales2016fast, carlone2015lagrangian, carlone2015duality}. 

Our previous work~\cite{garcia2020certifiable} 
proposed for the first time this kind of 
optimality certification 
for the Relative Pose problem. 
It was shown that the certifiable pipeline 
works in most problem instances 
under a wide variety of regimes. 
Here, we improve this certifier, 
extend it to consider bad matches, 
\ie outliers 
and evaluate the new proposal 
against the state-of-the-art solvers 
for the \RPp\xspace
in extensive experiments on 
both synthetic and real data.

\section{Notation}
In order to make clearer the mathematical formulation in the paper, 
we first introduce the notation used hereafter. 
Bold, upper-case letters denote matrices, \eg $\boldsymbol{E, C}$; 
bold, lower-case denotes (column) vectors, 
\eg $\trans, \xVec$; 
and normal font letters denote scalars, \eg $a, b$.
We reserve $\lambda$ for the Lagrange multipliers (Section \eqref{sec:closed-form-one}) and $\mu$ for eigenvalues. 
Additionally, we will denote with $\Reals{n \times m}$ 
the set of $n \times m$ real-valued matrices, 
$\symm{n} \subset \Reals{n \times n}$ the set of symmetric matrices of dimension $n \times n$ and 
$\symmPlus{n}$ the cone of positive semidefinite (PSD) matrices of dimension $n \times n$. 
A PSD matrix will be also denoted by $\succeq$ , 
\ie, $\matA \succeq 0 \Leftrightarrow \matA \in \symmPlus{n}$. 
We denote by $\kron$ the Kronecker product and 
by $\matI_n$ the (square) identity matrix of dimension $n$.
The operator $\VEC{\matB}$ vectorizes the given matrix $\matB$ column-wise. We denote by $\cross{t}$ the matrix form for the cross-product with a 3D vector $\trans = [t_1, t_2, t_3]^T$, \ie, $\trans \times (\bullet) = \cross{t} (\bullet)$ with
\begin{equation}
    \cross{t} =
    \begin{bmatrix}
    0 & -t_3 & t_2 \\
    t_3 & 0 & -t_1 \\
    - t_2 & t_1 & 0
    \end{bmatrix}.
    \label{eq:cross-t}
\end{equation}
Last, we employ the subindex $R$ through this document 
to indicate a relaxation of the element \wrt the element without subindex. 
For example, $\SetRelaxedEssentialMatrices$ 
stands for the set that is a relaxation of 
$\SetEssentialMatrices$ 
and therefore a superset of the latter, 
\ie $\SetRelaxedEssentialMatrices \supset \SetEssentialMatrices$.

We define the 2-sphere as 
$\sphere \doteq \{ \trans \in \Reals{3} | \trans^T \trans = 1 \}$
and the set of (3D) rotation matrices  as 
$\rotSpace \doteq \{ \rot \in \Reals{3 \times 3} | \rot^T \rot = \iden{3},\ \det(\rot) = +1$ \}.

\section{\RPp\xspace Formulation and Closed-form Dual Candidate} \label{sec:problemformulation}
We consider the central calibrated Relative Pose problem
in which one seeks the relative rotation $\rot$ and translation $\trans$ 
between two cameras,
given a set of $N$ pair-wise feature correspondences 
between the two images coming from these distinctive viewpoints, 
see Figure~\eqref{fig:def-relpose}.
In this work, the pair-wise correspondences are defined as pairs of (noisy) unit bearing vectors $(\obsi, \obsip)$ 
which,  if the were a correct correspondence, should point from their respective camera centers towards the same 3D scene point.
A traditional way to face this problem is by introducing the essential matrix $\Essential = \cross{t} \rot $~\cite{longuet1981computer, hartley2003multiple},
a $3 \times 3$ matrix 
that encapsulates the geometric information 
about the relative pose between two calibrated views. 
The essential matrix relates each pair of corresponding points through the \textit{epipolar constraint} $\obsi^T \Essential \obsip = 0$, 
provided that the observations are noiseless.
With noisy data, however, the equality does not hold and
$\obsi^T \Essential \obsip  = \epsilon_i$ defines 
what is called the \emph{normalized algebraic error}.

In this work we pose the \RPp\xspace as an optimization problem 
over the set of essential matrices 
that minimizes the squared algebraic error $\epsilon ^2$, 
as it has been performed previously in the literature, \eg,~\cite{ma2001optimization, Zhao2019, helmke2007essential}. 
We write the cost function in terms of $\Essential$ as 
a quadratic form via the positive semi-definite (PSD) 
matrix
$ \matC = \sum_{i=1} ^N \matC_i$, with $\matC_i =  (\obsip\kron \boldsymbol{f}_i)(\obsip \kron \obsi)^T \in \symmPlus{9}$ 
and so, the problem reads:

\begin{equation}
    \tag{O} \label{eq:originalproblem}
    \fOpt = \min_{\Essential \in \SetEssentialMatrices} \sum_{i = 1} ^N (\obsi^T \Essential \obsip ) ^2 =
    \min_{\Essential \in \SetEssentialMatrices} \underbrace{ \VEC{\Essential}^T \matC \VEC{\Essential}}_{\fE}.
\end{equation}
See~\cite[App.~A]{garcia2020certifiable} for a formal proof of the equivalence between the objective functions.
In~\eqref{eq:originalproblem}, $\SetEssentialMatrices$ stands for the set of (normalized) essential matrices, typically defined as
\begin{equation}
\label{eq:Me:[t]xR}
    \SetEssentialMatrices \doteq \{ \Essential \in \Reals{3 \times 3} \;|\; \Essential = \cross{t} \rot , \rot\in \rotSpace,\ \trans \in \sphere \}.
\end{equation}
Note that in \eqref{eq:Me:[t]xR} the translation is identified with points in the 2-sphere
due to the scale ambiguity for central cameras.

\subsection{A Relaxation of the Set of Essential Matrices}
For the purpose of this work, 
a minimal (quadratic) representation is preferred. 
We leverage the one proposed by Faugeras \etal in~\cite{faugeras1990motion}:
\begin{equation}
    \SetEssentialMatrices \doteq \{ \Essential \in \Reals{3 \times 3} \;\vert \; \Essential \Essential ^T = \cross{t}\cross{t}^T , \trans^T \trans = 1\}
    . \label{eq:Me:EEt}
\end{equation}
This parameterization, recently exploited by Zhao in~\cite{Zhao2019}, 
features a low number of variables $(12)$ and constraints $(7)$, 
and was proved to be useful in practice. 
Despite its advantages, 
this parameterization 
still does \emph{not} allow for the development 
of a \emph{fast optimality certifier} for the \RPp\xspace
in the fashion of that proposed for problems like Pose Graph Optimization, as in \cite{briales2016fast,carlone2015lagrangian}. 

To overcome this, 
in~\cite{garcia2020certifiable} 
we proposed a relaxed version 
$\SetRelaxedEssentialMatrices$ 
of the essential matrix set 
$\SetEssentialMatrices$ 
by dropping one the constraints 
in \eqref{eq:Me:EEt}, 
concretely  $\eVec_1^T \eVec_2 + t_1 t_2 = 0$ .  
This relaxed set is then defined as:
\begin{equation}
    \SetEssentialMatrices \subset \SetRelaxedEssentialMatrices \doteq \{ \Essential \in \Reals{3 \times 3} \;|\; h_i(\Essential,\trans)=0, \forall h_i \in \constrSetRelaxed;\ \trans \in \Reals{3}\},
    \label{eq:Me:min}
\end{equation}
with $\constrSetRelaxed$ the \emph{relaxed constraint set} defined as
\begin{equation}
\constrSetRelaxed \equiv \begin{cases}
   h_1 \equiv \boldsymbol{t}^T \boldsymbol{t} - 1 = 0\\
   h_2 \equiv \eVec_1^T \eVec_1 - (t_2^2 + t_3^2) = 0\\
   h_3 \equiv \eVec_2^T \eVec_2 - (t_1^2 + t_3^2) = 0\\
   h_4 \equiv \eVec_3^T \eVec_3 - (t_1^2 + t_2^2) = 0\\
   h_5 \equiv \eVec_1^T \eVec_3 + t_1 t_3 = 0\\
   h_6 \equiv \eVec_2^T \eVec_3 + t_2 t_3 = 0
\end{cases},
\label{eq:set-relaxed-essential}
\end{equation}
and the rows of $\Essential$ have been denoted by 
$\eVec_i \in \Reals{3}, \forall i \in \{1, 2, 3\}$.
A formal proof of how $\SetRelaxedEssentialMatrices$~in~\eqref{eq:Me:min} 
defines a strict superset of $\SetEssentialMatrices$ is provided in
~\cite[App.~B]{garcia2020certifiable}.

We could have discarded any other constraint (except $h_1$) and
hence, obtained other similar yet different relaxed constraint sets.
For the sake of simplicity, we will restrict ourselves to the
set in \eqref{eq:set-relaxed-essential} 
and we provide the general form of the certifier 
in Section~\eqref{sec:family-certifiers}. 
The following development can be trivially
adapted to any other relaxation of $\SetEssentialMatrices$.

\smallskip
\textbf{Relaxed Formulation of the Relative Pose Problem}

With this relaxed set at hand, we define a \emph{relaxed} version \eqref{eq:relaxedproblem}
of the original \RPp\xspace in \eqref{eq:originalproblem}:
\begin{equation}
    \tag{R} \label{eq:relaxedproblem}
    \fOptR =
    \min_{\Essential \in \SetRelaxedEssentialMatrices} \VEC{\Essential}^T \matC \VEC{\Essential}.
\end{equation}
Since problem \eqref{eq:relaxedproblem} is a relaxation of \eqref{eq:originalproblem},
the inequality $\fOptR \leq \fOpt$ holds with equality
only if the solution to \eqref{eq:relaxedproblem} is also an essential matrix,
and hence also feasible for \eqref{eq:originalproblem}.
Interestingly enough though,
we observed that equality holds ($\fOptR = \fOpt$) in many problem instances in practice,
meaning that the relaxed problem \eqref{eq:relaxedproblem}
is very often a tight relaxation of the original problem \eqref{eq:originalproblem}.
We have no theoretical proof as to why the behavior above holds so often,
and our support to this claim is fundamentally empirical
(given by extensive experiments in Section~\eqref{sec:experiments}).

\subsection{Closed-form Expression for Dual Candidates} \label{sec:closed-form-one}
To derive our certifier, 
we first re-formulate the relaxed problem~\eqref{eq:relaxedproblem} 
as a standard instance of QCQP by writing explicitly the constraints in $\constrSetRelaxed$. 
For that, we define the 12-D vector $\xVec = [\eVec^T, \trans^T ] ^T$, with $\eVec = \VEC{\Essential}$ and hence:
\begin{align}
    \fOptR =
    \min_{\xVec \in \Reals{12}} \xVec ^T  \dataMatrixQ \xVec  \nonumber \quad
    \text{subject to }
    \xVec^T \matA_i \xVec = c_i, i = 1, \dots, 6
    \tag{P-R}\label{eq:primalproblem}
\end{align}
where $\{\matA_i\}_{i=1} ^6$ are the $12 \times 12$-symmetric 
corresponding matrix forms of the quadratic constraints,
so that 
$h_i(\Essential,\trans) \equiv \xVec^T \matA_i\xVec - c_i = 0, c_i \in \Reals{}$ 
with $c_1 = 1$ and $c_i = 0$ for $i = 2, \dots, 6$, 
and $\dataMatrixQ $ is the 
$12 \times 12$-symmetric data matrix of compatible dimension with $\xVec$ padded with zeros.

Problem \eqref{eq:primalproblem} is exactly equivalent to the relaxed Problem \eqref{eq:relaxedproblem}.
Nonetheless, Problem \eqref{eq:primalproblem} is still a 
Quadratically Constrained Quadratic Program (QCQP),
in general NP-hard to solve.
However, it allows us to derive an optimality certifier by exploiting the so-called \textit{Lagrangian dual problem}~\cite{boyd2004convex},
that takes the form:

    \begin{align}
    \dOptR
    =  \max_{\bm{\lambda} } \lambda_1  \tag{D-R} \label{eq:dualproblem}
    \quad
    \text{subject to  }  \HessLambda\succeq 0  \nonumber
    \end{align}

where $ \HessLambda \doteq \dataMatrixQ  - \sum_{i=1} ^6 \lambda_i \matA_i$ is the so-called \textit{Hessian of the Lagrangian}
and $\bm{\lambda} = \{\lambda_i\}_{i=1} ^6$ are the \textit{Lagrange multipliers}.
The derivation of the problem is given in~\cite[App.~C]{garcia2020certifiable}.

The dual problem \eqref{eq:dualproblem} presents a relaxation of the primal program \eqref{eq:primalproblem} and
hence, its objective cost $\dHatR$ provides us with a lower bound for the objective of the latter $\fHatR$,
principle known as \textit{weak duality}~\cite{boyd2004convex}.
Consider a feasible dual point $\vHatLambda$,
\ie a point that fulfills the constraints required by the dual problem,
for \eqref{eq:dualproblem} and
a feasible primal point $ \vHatX$ for the relaxed primal in \eqref{eq:primalproblem}.
The objective costs attain by this pair of primal/dual points 
($\fHatR, \dHatR$, respectively)
are related by the following chain of inequalities: 
\begin{equation}
   \dHatR \stackrel{a}{\leq} \dOptR \stackrel{b}{\leq} \fOptR \stackrel{c}{\leq}\fOpt \stackrel{d}{\leq}\fHatE, \label{eq:chain-ineq-original}
\end{equation}
where
(a) and (d) hold by definition of optimality, 
(c) stands since \eqref{eq:primalproblem} is a relaxation of \eqref{eq:originalproblem}
and (b) holds because, at the same time, 
the dual problem is a relaxation of the primal 
\eqref{eq:primalproblem} (\emph{weak duality}).

From~\eqref{eq:chain-ineq-original}
we can see that the dual problem allows to certify if a given primal feasible point $\vHatX$ is indeed optimal.
If for a pair of primal/dual feasible points we have that $\fHatR = \dHatR$ (up to some accuracy),
then the chain of inequalities becomes tight
and $\fOpt = \fHatR$, which shows that $\vHatX$ is the optimal solution.
In these cases, we say that there is \textit{strong duality} and
that the \textit{duality gap} $\fOptR - \dOptR$ is zero.

Although the dual problem \eqref{eq:dualproblem} 
is a SDP and can be solved by off-the-shelf solvers
(\eg, SeDuMi~\cite{sturm1999using} or SDPT3~\cite{toh1999sdpt3}) 
in polynomial time,
a faster optimality verification 
can be derived. 
This certification 
relies 
on a closed-form expression for dual candidates $\vHatLambda$,
thus avoiding the resolution of the SDP from scratch.
This closed-form expression is derived from the requirement 
that every primal solution $\vHatX$ 
must fulfill the KKT conditions~\cite[Sec.~5.5]{boyd2004convex}.

\begin{theorem}[Verification of Optimality~{\cite[Th.~5.2]{garcia2020certifiable}}]
    \label{theorem:verification}
    Given a putative primal solution $\vHatX$ for Problem \eqref{eq:primalproblem}, if there exists a solution $\vHatLambda$ to the linear system
    \begin{equation}
        \matJ (\vHatX) \vHatLambda = \dataMatrixQ  \vHatX ,
        \label{eq:systeminlambdahat}
    \end{equation}
    and $\HessLambdaHat\succeq 0$, 
    then we have strong duality and the putative solution $\vHatX$ is indeed optimal. 
    Here  $\matJ (\vHatX) \in \Reals{12 \times 6}$ 
    is the full-rank Jacobian of the constraints 
    evaluated at the potentially optimal point $\vHatX$ 
    for the relaxed problem \eqref{eq:primalproblem} 
    whose explicit form can be found in \cite{garcia2020certifiable} 
    and in the \suppl~Section~(A).
\end{theorem}

The restriction of Theorem \eqref{theorem:verification} 
to essential matrices is summarized as follows:
\begin{corollary} \label{cor:verification-essential}
~\cite[Cor.~5.2.1]{garcia2020certifiable}
Given a potentially optimal solution $\hatE$ for problem \eqref{eq:originalproblem} and its equivalent form $\vHatX = [\VEC{\hatE}^T, \hat{\trans}^T]^T$ where $\hat{\trans}$ is the associated translation vector, if there exists a \emph{unique} solution $\vHatLambda$ to the linear system in \eqref{eq:systeminlambdahat} and it is dual feasible (\ie $\HessLambdaHat\succeq 0$), then we can state that: (1) strong duality holds between problems \eqref{eq:primalproblem} and \eqref{eq:dualproblem}; (2) the relaxation carried out in \eqref{eq:primalproblem} is \emph{tight}; and (3) the potentially optimal solution  $\hatE $ is optimal for both \eqref{eq:primalproblem} and \eqref{eq:originalproblem}.
\end{corollary}

We want to point out that 
Theorem \eqref{theorem:verification} and
Corollary \eqref{cor:verification-essential} 
can only either certify the given primal solution is indeed optimal,
or it is inconclusive about its optimality.
The reason for the latter is associated either with a suboptimal solution 
and/or a loose relaxation
(either the one carried out for the primal in 
\eqref{eq:primalproblem} and/or
the one for the dual in \eqref{eq:dualproblem}).

\begin{remark}
In this work we propose a set of six relaxations 
of the original problem 
\eqref{eq:originalproblem} 
obtained by dropping only 
\emph{one} constraint 
from the original set in 
\eqref{eq:Me:EEt}. 
These relaxations 
allow to compute 
a unique dual candidate 
$\vHatLambda$ in closed-form 
from \eqref{eq:systeminlambdahat} 
since the Jacobian is full rank. 
We can, 
nevertheless, 
discard more constraints 
and 
still derive closed-form expressions  
for dual candidates 
and the associated 
fast certification algorithms 
akin to Theorem~\eqref{theorem:verification}. 
These problems, 
though, 
are looser relaxations  
of the original~\eqref{eq:originalproblem},    
and their 
dual problems may not provide with a 
tight lower bound, 
that is,  
the certifier associated to that relaxation will not 
be able to certify the optimal solution 
to the original problem. 
While the computation cost for these smaller certifiers 
could be potentially lower 
(the linear system in \eqref{eq:systeminlambdahat} 
has fewer unknowns $\vHatLambda$), 
we do not know a priori 
whether any of these relaxations 
perform well  
in terms of certification of essential matrices.  
Therefore, 
the tightest relaxations  
for the parameterization 
of the essential matrix set 
in \eqref{eq:Me:EEt} 
that still assure 
closed-form expressions 
with unique solution 
for dual candidates 
are obtained by dropping only 
\textit{one} constraint of the set.  
\end{remark}

\begin{figure}[t!]
    \centering
    \includegraphics[width=0.9\columnwidth]{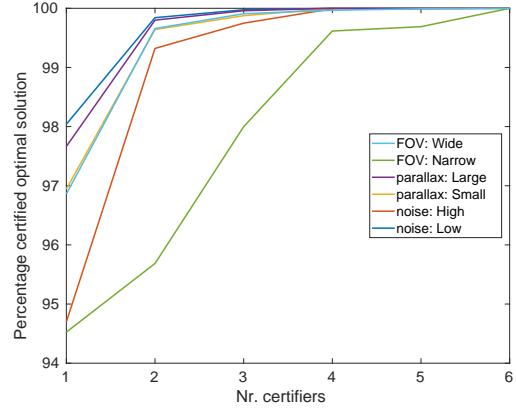}
    \caption{Cumulative percentage of cases with certified solutions
    as a function of the number of computed certifiers
    for regimes with different noise, parallax and field of view.
    ``Parallax: Small`` and ``FOV: Wide`` overlap in the figure.}
    \label{fig:perc-certifiers-basic}
\end{figure}

\subsection{Family of Certifiers from different Closed-form Dual Candidates} \label{sec:family-certifiers}

The employed relaxation of the
feasible set of essential matrices in \eqref{eq:set-relaxed-essential} was necessary in order to
assure the uniqueness of the Lagrange multipliers. 
We can, nevertheless, obtain similar yet different relaxations
by dropping one of the other five different constraints in 
$\Essential \Essential^T$.
These relaxations would yield distinct 
optimality certifiers 
for the same problem 
whose candidates to Lagrange multipliers 
are also computed in closed-form.
If (at least) one of them fulfills the conditions for optimality
(see Theorem \eqref{theorem:verification}),
then the solution is optimal and the associated underlying relaxation is tight.
On the other hand, if none of them fulfill these conditions,
this does not imply that the
solution is suboptimal: all the relaxations could be not tight 
at the same time as well.
In its general form, we can write the closed-form expression for the multipliers as:
\begin{equation}
    \dataMatrixQ \vHatX = \sum_{i=2}^7 \theta_i \hatLambda_i \matA_i \vHatX +
    \hatLambda_1 \matA_1 \vHatX = \matJ(\vHatX, \bm{\theta}) \vHatLambda, \label{eq:gen-certifier}
\end{equation}
where $\theta_i \in \{0, 1 \}, i=2, \dots, 7$, with $\sum_i \theta_i = 5$
and $\bm{\theta} \defexp [\theta_2, \dots, \theta_7]$. 
The variable $\theta_i$ ``activates`` the $i$-th constraint 
(with matrix form $\xVec^T \matA_i \xVec = c_i, \ i= 2, \dots, 7$) 
when it is equal to one.
In other terms,
for any combination of $\bm{\theta}$, 
one of the constraints in 
\eqref{eq:Me:EEt} (except $h_i$) 
is always deactivated.
The column of the full Jacobian 
associated with that multiplier is zero for any point $\vHatX$
and it can be dropped.
Hence, the matrix $\matJ(\vHatX, \bm{\theta}) $ has always linearly independent columns
(following a similar proof to that provided in~\cite[App.~D]{garcia2020certifiable}).

In the worst case, we compute six different certifiers: 
solve Equation \eqref{eq:gen-certifier} 
with the six different combinations for $\bm{\theta}$); 
and 
six eigenvalues of a $3 \times 3$ matrix (closed-form)
and six eigenvalues of a $9 \times 9$ matrix.
In practice, however, this is not necessary.
Figure~\eqref{fig:perc-certifiers-basic} depicts the cumulative percentage of cases
with certified optimal solutions (Y-axis)
as a function of the number of computed certifiers (X-axis)
for different level of noise, parallax and Field of View
(Section~\eqref{sec:experiments} explains the details about the experiments).
Note that at least $99\%$ of optimal cases are detected
with only $2$ certifiers for most of the problems
(except narrow field of views, \FOV),
while the $100\%$ is detected with $4$ certifiers.
For narrow \FOV~$70-90\ \degrees$, however,
one may need to compute all of them.

This redundant estimation of potential dual candidates increases
the computation time required by the verification stage.
We observe that estimating the Lagrange multipliers
and checking the spectral conditions takes
approximately the same amount of time
that the initialization by the \eightpt algorithm
(singular vector estimation and projection).
We can, of course, compute only one certifier.
As it is shown in Figure~\eqref{fig:perc-certifiers-basic},
we will still detect more than $95\%$ of the optimal cases in general.
The choice is left to the user and the application.

\subsection{Exploiting structure: separating and ordering}

Once the candidate has been computed, 
Theorem~\eqref{theorem:verification} 
requires to analyze the 
eigenvalues of the Hessian of the Lagrangian $\HessLambdaHat$. 
This matrix is block-diagonal with two (squared) blocks 
$\HessE, \HessT$ of size 9 and 3, respectively.
Both blocks must be positive semi-definite (PSD) for the full Hessian to be PSD,
since the spectrum of a block-diagonal matrix is the union of the spectra of its blocks.
We can then translate the original condition $\HessLambdaHat \succeq 0$
to the two smaller conditions $\HessE \succeq 0$ and $\HessT \succeq 0$,
which must hold at the same time.
Further, and although our justification here is purely empirical,
we observe that those cases for which the Hessian was not PSD
were associated with $\HessT$ being not PSD.
Therefore and in practice,
we check first the PSD condition for this
$3\times 3$ matrix
(whose eigenvalues have a closed-form expression).
If it is fulfilled, we then verify the condition for $\HessE$.
If both blocks are PSD, then the Hessian is PSD.

 \begin{algorithm}[t!]
    \caption{Optimality Certification} \label{alg:certification}
    \hspace*{\algorithmicindent} {\textbf{Input}: Compact data matrix $\matC$; primal solution $\vHatX$}\\
    \hspace*{\algorithmicindent} {\textbf{Output}: Optimality certificate \textsc{isOpt} $\in \{\text{True}, \text{unknown}\}$} \\

    Compute $\fHatR$ from \eqref{eq:primalproblem}\;
    idxRelaxation $\leftarrow$ 1 \; 
        
        \Repeat {\textsc{isOpt} is \textsc{True} or \text{idxRelaxation} is $6$}
        {
        \tcp{For the i-th relaxation}
         {Compute $\vHatLambda$ by solving the linear system in \eqref{eq:systeminlambdahat} and set $\dHatR =  \hat{\lambda}_1 $\; }
        
        Compute min. eigenvalue $\mu_{t}$ of $\HessT$\;

            \eIf {$\mu_t \leq \tauMu$ or $|\fHatR - \dHatR| > \tauGap$} 
            {  \tcp{Dual candidate is not feasible}
                 {\textsc{isOpt} $=$ unknown\; 
                 idxRelaxation $\leftarrow$ idxRelaxation + 1 \;
                  }
            }
            {
              {Compute min. eigenvalue $\mu_{E}$ of $\HessE$\;}
                \eIf {$\mu_E \leq \tauMu$ or $|\fHatR - \dHatR| > \tauGap$ }
                {
                    \tcp{Dual candidate is not feasible}
                     {\textsc{isOpt} $=$ unknown\; 
                     idxRelaxation $\leftarrow$ idxRelaxation + 1 \;
                     }
                }
                {
                    \tcp{Dual candidate is feasible}
                    {\textsc{isOpt} $=$ True\;}
                }
            }
        }

    \end{algorithm}

\medskip
With these improvements at hand, 
we can now provide the optimality certification algorithm 
in Algorithm~\eqref{alg:certification}. 
To universalize the Algorithm,
let us denote by $\vHatX = [ \VEC{\hatE}^T, \hat{\trans}^T]^T$ 
the feasible solution for \eqref{eq:primalproblem} or for \eqref{eq:originalproblem}.
Further, for any $\hatE \in \SetEssentialMatrices$, the attained objective value in \eqref{eq:originalproblem} $\big( \fHatE \big) $ agrees with the objective value in \eqref{eq:primalproblem} $\big( \fHatR \big)$ since $\SetEssentialMatrices \subset \SetRelaxedEssentialMatrices$;
hence we employ $\fHatR$ to denote the corresponding objective value in both cases without confusion.
Recall that our certification has two possible outcomes,
either \textsc{Positive} (the solution is optimal) or
\textsc{unknown} (the certification is inconclusive).
From a practical point of view, we write the conditions 
$\HessE, \HessT \succeq 0$ 
as their smallest eigenvalues
$\mu_t, \mu_{E}$ being greater than a negative threshold $\tauMu$. 
Further, 
while our closed-form certifier 
assumes that strong duality holds, 
due to rounding errors 
the dual gap $\fHatR - \dHatR$ 
may not be exactly zero. 
We assure strong duality by applying a (positive) threshold $\tauGap$
to the absolute value of the dual gap $|\fHatR - \dHatR|$,
which allow us to accommodate numerical errors. 
In practice, we fix the tolerances to $\tauMu = -0.02$ and $\tauGap = 10^{-14}$. 
If any of the minimum eigenvalues are negative and/or 
the dual gap is greater than zero 
(considering the tolerance in both cases),
the certification procedure is inconclusive.

\section{Proposed Fast Certifiable Pipeline} \label{sec:proposed-pipeline}
Rather than solving the original problem \eqref{eq:originalproblem}
via its convex SDP relaxation,
in this work we propose to solve the \RPp\xspace through an iterative method that respects the intrinsic nature of the essential matrix set,
but comes with no optimality guarantees, and
to certify \textit{a-posteriori} the optimality of the solution leveraging our fast optimality certifier
in Algorithm~\eqref{alg:certification}.
Current iterative methods work well and usually converge to the global optima 
in addition to be faster than the methods employed in convex programming,
\eg, Interior Point Methods (IPM).
Following, we enumerate and briefly explain the three major stages in which the proposed pipeline is separated: 
\begin{enumerate}
    \item \textbf{Initialization}: 
    We start by generating an initial guess
    with any standard algorithm, 
    \eg the \eightpt algorithm~\cite{hartley2003multiple}. 
    \item \textbf{Refinement (optimization on manifold)}:
    We seek the solution to the original primal problem \eqref{eq:originalproblem}
    by refining the initial guess with a local iterative method
    that operates within the essential matrix manifold $\Me$
    (always fulfilling constraints).
    \begin{align}
     \label{eq:optim-on-manifold}
     \hat{\Essential} = \argmin_{\Essential \in \Me}  \VEC{\Essential}^T \matC \VEC{\Essential}.
    \end{align}
    Recall that the matrix $\dataMatrixQ \in \symmPlus{12}$ 
    is the data matrix $\matC \in \symmPlus{9}$ padded with zeros.
    \item \textbf{Verification of optimality}: 
    The candidate solution $\hat{\Essential}$ returned 
    by the iterative method 
    can be certified as the global optimum  
    with the proposed Algorithm \eqref{alg:certification}
    if the underlying dual problem~\eqref{eq:dualproblem} is tight.
\end{enumerate}

Note that the above-mentioned pipeline estimates the essential matrix, which encodes the relative pose.
Both the rotation and translation \uts can be extracted from it by any classic computer vision algorithm~\cite{hartley2003multiple}.

\subsection{Refinement of the initial estimation: optimization on manifold}
\label{sec:essential-manifold}
Riemannian optimization toolboxes decouple the optimization problem into
manifold (domain), problem description and solvers,
thus making it quite straightforward to implement an iterative solver for \eqref{eq:optim-on-manifold} as proposed above.

\smallskip
\textbf{Domain}:
Current tools for optimization on manifolds, such as \manopt~\cite{manopt},
provide 
a wide variety on pre-defined manifolds and
operators which alleviate the programming effort required for its use.
Although the implementation is pretty simple 
when the employed manifold is supplied by the tool
(as it happens in~\cite{garcia2020certifiable}), 
this is not always the case.
In those situations, the user needs to either implement it or employ a different
but equivalent parameterization of the manifold whose components are provided by the library.

In the \RPp\xspace the essential matrix manifold can be simplified
in most cases to the direct product space $\rotSphereSpace$
by leveraging the standard definition given in \eqref{eq:Me:[t]xR}.
This product is composed of two well-known spaces, $\rotSpace, \sphere$,
which are usually provided by the optimization toolboxes.
In any case, they can be also easily implemented,
so is their direct product (see \eg \cite{ma2001optimization}).

Despite its simplicity, however,
the above-mentioned product is indeed a four-fold covering of the essential matrix manifold $\Me$
(ambiguity which is well-known \cite{hartley2003multiple, briales2018certifiably}).
Nevertheless, this does not suppose a major issue when we are only required
to estimate one single essential matrix at a time (manifold optimization);
we refer the reader to \cite{tron2017space} for a full characterization of $\Me$
and to \cite{dubbelman2012manifold} for a detailed explanation of those cases
in which this simplification cannot be employed directly (manifold statistics).

Although simple, this re-parameterization of the search space requires to adapt the objective function to the new set of variables,
which may not be straightforward.
Further, we also need to provide the gradient and the Hessian
(if the chosen solver employs second-order information)
\wrt this new parameterization.
Thanks to the decoupling of the problem into domain and model description,
the well-known nature of the rotation and sphere groups and
the simplification of the product of manifolds,
there exist simple relations between the ambient gradient and Hessian-vector product
and their Riemannian counterparts.
First, let the objective function $\fRt$ be defined in terms of 
the rotation and translation $(\rot, \trans)$
(to be defined later).
The Riemannian gradient $\RGradfRt$ and Hessian-vector product $\RHessfRt$ 
defined on the space $\rotSphereSpace$ can
be treated element-wise, that is: \\ 
$\RGradfRt = (\RGradfR, \RGradft) $,
where the two terms are the Riemannian gradients 
\wrt $\rot$ and $\trans$, respectively.
Similarly, the Riemannian Hessian-vector product is defined as \\ 
$\RHessfRt = (\RHessfR, \RHessft)$, 
with the same notation than with the gradient.
Hence, we only need to treat each component individually.

\smallskip
\textbf{Gradient}:
By considering the objective function $\fRt$ as a function restricted 
to the embedded submanifold 
$\rotSphereSpace \subset \Reals{3 \times 3} \times \Reals{3}$,
we can define each term in the Riemannian gradient \\ 
$\RGradfRt = (\RGradfR, \RGradft) $
as the \emph{orthogonal projection} to the tangent space
of the corresponding Euclidean gradient 
($\gradfR, \gradft$)\cite[Eq.~3.37]{absil2009optimization}, \ie,
\begin{align}
    \RGradfR &= \ProjR(\gradfR) \nonumber \\
    \RGradft &= \Projt(\gradft)
\end{align}

The orthogonal projection operator $\ProjR(\bullet)$ 
onto the tangent space of $\rotSpace$ ($\TanR $) 
at the point $\rot$ 
is defined as~\cite{absil2009optimization}
\begin{align}
    &\ProjR: \TanR (\Reals{3 \times 3}) \longrightarrow \TanR (\rotSpace) \\
    & \ProjR(\matX) = \rot \text{skew}(\rot^T \matX),
\end{align}
where $\text{skew}(\matA)$ extracts the skew-symmetric part of the matrix $\matA$, 
\ie $\text{skew}(\matA) = \frac{1}{2} ( \matA - \matA^T)$. 
Similarly for the translation term, 
the orthogonal projection operator $\Projt(\bullet)$ 
onto the tangent space of 
$\sphere$ ($\Tant$)
at the point $\trans$ is 
\begin{align}
    &\Projt: \Tant (\Reals{3}) \longrightarrow  \Tant (\sphere) \\
    & \Projt(\xVec) = \xVec - (\trans^T \xVec) \trans.
\end{align}

\smallskip
\textbf{Hessian-vector product}:
Similarly, the two components of the Riemannian Hessian-vector product \newline
$\RHessfR$ and $\RHessft$ are computed as
the orthogonal projections of the ambient directional derivative
of the gradient vector fields $\RGradfR$, \newline
$\RGradft$ in the direction of
$\dot{\rot}, \dot{\trans}$, 
respectively~\cite[Eq.~5.15]{absil2009optimization}:

\begin{align}
    \RHessR (\rot,\trans) [\VR,\Vt] &= \ProjR \Big(  \HessfR[\VR, \Vt] - \nonumber\\
    & \qquad \VR^T \text{sym} \big ( \rot \gradfR \big ) \Big) \\
    \RHesst (\rot, \trans) [\VR, \Vt] &= \Projt \Big(  \Hessft [\VR, \Vt]\Big) - \nonumber \\
    & \qquad \big (\trans^T \gradft \big )\Vt,
\end{align}
where $\text{sym}(\matA)$ extracts the symmetric part of the matrix $\matA$ \ie, $\text{sym}(\matA) = \frac{1}{2} (\matA + \matA^T)$.

\smallskip
\textbf{Quadratic Model of the Problem}:
With this in mind, we show next how to write the original objective function in \eqref{eq:primalproblem}
in terms of the elements $(\rot, \trans) \in \rotSphereSpace$ and provide the Euclidean gradient and Hessian-vector product. 
The following Theorem allows us to express the original quadratic cost function
in terms of the points $(\rot, \trans) \in \rotSphereSpace$ in two different and equivalent ways.

\begin{theorem}[Equivalent expressions for the cost function in \eqref{eq:primalproblem}]
    \label{th:eqexpressions}
    The original cost function has the two following equivalent forms as functions of $\rot, \trans$
    \begin{equation}
        \frac{1}{2}\eVec^T \matC \eVec  = \frac{1}{2}\trans^T \Mt \trans = \frac{1}{2}\rVec ^T \Mr \rVec  
    \end{equation}
    where
    $\eVec = \VEC{\Essential} \in \Reals{9}$,
    $\rVec = \VEC{\rot} \in \Reals{9}$,
    \begin{align}
        \Mr(\trans) &\defexp (\iden{3} \kron \cross{t}) ^T \matC (\iden{3} \kron \cross{t})
        \in \symm{9} \\
        \Mt(\rot) &\defexp  \matB^T (\rot^T \kron \iden{3})^T\matC(\rot^T \kron \iden{3})\matB
        \in \symm{3},
    \end{align}
    being $\matB$ a sparse matrix such that $\vCross{\trans} = \matB \trans$ holds for any $\trans \in \Reals{3}$.
    Their full development \emph{from} the original quadratic cost function, as well as their relation with previous works \cite{briales2018certifiably}, \cite{kneip2013direct} are given in the \suppl~Section~(B).
\end{theorem}

From Theorem \eqref{th:eqexpressions}, we can derive the (Euclidean) gradient and Hessian required
to define the quadratic model of the problem in a straightforward manner.
\begin{theorem}[Euclidean Gradient and Hessian in $\Reals{3 \times 3} \times \Reals{3} \supset \rotSphereSpace$]
    \label{th:quadmodel}
    From Theorem \eqref{th:eqexpressions}, the (Euclidean) gradient $\gradfRt \in \Reals{12}$ and
    Hessian-vector product $\HessfRt[\VR, \Vt] \in \Reals{12}$ are defined as:
    \begin{align}
        \gradfRt &=
        \begin{pmatrix}
        \gradfR \\
        \gradft
        \end{pmatrix}
        =
        \begin{pmatrix}
        \Mr & \zeros{9 \times 3} \\
        \zeros{3 \times 9} & \Mt
        \end{pmatrix}
        \begin{pmatrix}
        \rVec \\
        \trans
        \end{pmatrix} \\
    \HessfRt [\VR, \Vt] &=
        \begin{pmatrix}
        \HessfR[\VR, \Vt]  \\
        \Hessft [\VR, \Vt]
        \end{pmatrix}
        \nonumber \\
        &=
        \begin{pmatrix}
        \Mr & \Mtr \\
        \Mtr^T & \Mt
        \end{pmatrix}
        \begin{pmatrix}
        \Vr \\
        \Vt
        \end{pmatrix}
    \end{align}
    where $\Mr, \Mt$ are the symmetric matrices defined above. 
    $\Reals{9 \times 3} \ni \Mtr \doteq \MtrRT$ is provided in the 
    \suppl~Section~(C).
\end{theorem}

\smallskip
\textbf{Solver}:
We choose here an iterative \emph{truncated-Newton Riemannian trust-region} (RTR) solver~\cite{absil2009optimization}.
RTR has shown before~\cite{briales2017cartan, rosen2019se} a very good trade-off
between a large basin of convergence and superlinear convergence speed.

In order to speed up the converge of the algorithm and inspired
by \cite{briales2017cartan} and \cite{rosen2019se},
we propose a suitable Hessian preconditioner for the relative pose problem
which is proved through extensive experiments in Section \eqref{sec:experiments}
to reduce the number of iterations required by the solver to converge.
The use of a preconditioner is based on the dependence of the 
convergence rate of the conjugate method when solving the linear system $\matA \xVec = \bVec$ 
with the condition number of the matrix $\matA$. 
The \textit{Riemannian} preconditioner 
$\matP$ must be a linear, symmetric and positive definite operator 
from the tangent space of the manifold to itself.
We propose to use the preconditioning operator similar to Jacobi and of the form
\begin{equation}
    \matP(\matU) \doteq \ProjRt(\frac{1}{\alpha} \matU),
\end{equation}
with $\alpha$ is a positive scalar related 
to the eigenvalues of the data matrix $\matC$.
Since the data matrix cannot be zero and it is PSD,
this preconditioner is, by definition, 
linear, symmetric and positive definite.
It is also defined from and onto the tangent space since 
$\matU$ belongs to this tangent space
and the result is projected onto the tangent space.
We propose the value for $\alpha = \lambda_1 + \lambda_2 + \lambda_3$, 
which are the three largest eigenvalues of the data matrix $\matC$.
Although this preconditioner is heuristic,
it is showcased later on Section~\eqref{sec:experiments} 
that it does reduce the computation time
for the iteration method,
specially when the number of correspondences is large,
as it is usually the case in practical applications.

\section{Robust Resolution of the Relative Pose Problem} \label{sec:GNC}

Section~\eqref{sec:proposed-pipeline} outlines 
a complete certifiable pipeline
for the essential matrix estimation whose core lies
on the local refinement stage.
This refinement is, however, very sensitive to bad correspondences,
\ie outliers;
in the presence of only one outlier, 
the solution will be biased. 
Instead of embedding and 
adjusting our pipeline into a RANSAC paradigm,
we leverage the tools known as 
Graduated Non-convexity (GNC)~\cite{blake1987visual} and
the Black-Rangarajan duality~\cite{black1996unification} 
to detect and discard outliers. 
This scheme has been previously employed, 
see \eg~\cite{Zhao2019, yang2020graduated}, 
showing that it is able to estimate  
better solutions than RANSAC-based paradigms, 
specially for problem instances 
where the percentage of outliers is large.  
In this Section 
we summarize this scheme 
and specialized it for the Tukey's biweight 
loss function~\cite{black1996unification}. 
The approach is extensible to other functions, 
some of them can be also found in previous works
(\eg, Welsch function in~\cite{Zhao2019}
and Truncated Least Square and Geman-McClure in~\cite{yang2020graduated}).

We first introduce a 
robust loss function $\opLoss(\bullet)$ 
into the cost function $\fE$ of the original problem \eqref{eq:originalproblem}.
The robust variant of \eqref{eq:originalproblem} is written as
\begin{equation}
    \label{eq:robust-original-problem}
    \min_{\Essential \in \SetEssentialMatrices} \sum_{i = 1} ^N \opLoss \big( \epsilon_i \big ),
\end{equation}
where recall that $\epsilon_i$ was the epipolar error 
for the $i$-th correspondence.

\begin{figure}
 \begin{subfigure}{0.49\linewidth}
     \centering
     \includegraphics[width=0.98\linewidth]{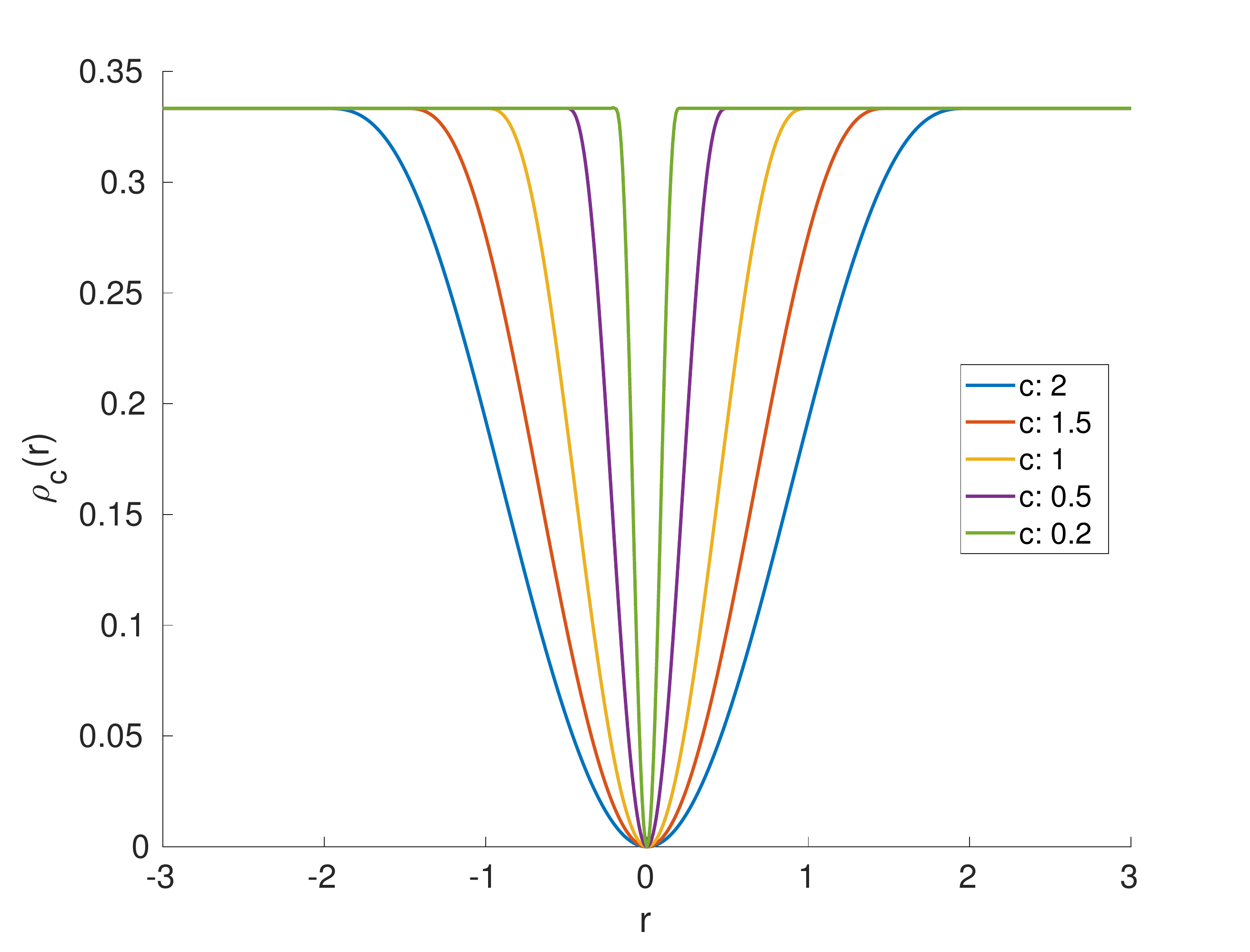}
     \caption{}
     \label{fig:tukey-loss-c-hat}
\end{subfigure}
\begin{subfigure}{0.49\linewidth}
     \centering
     \includegraphics[width=0.99\linewidth]{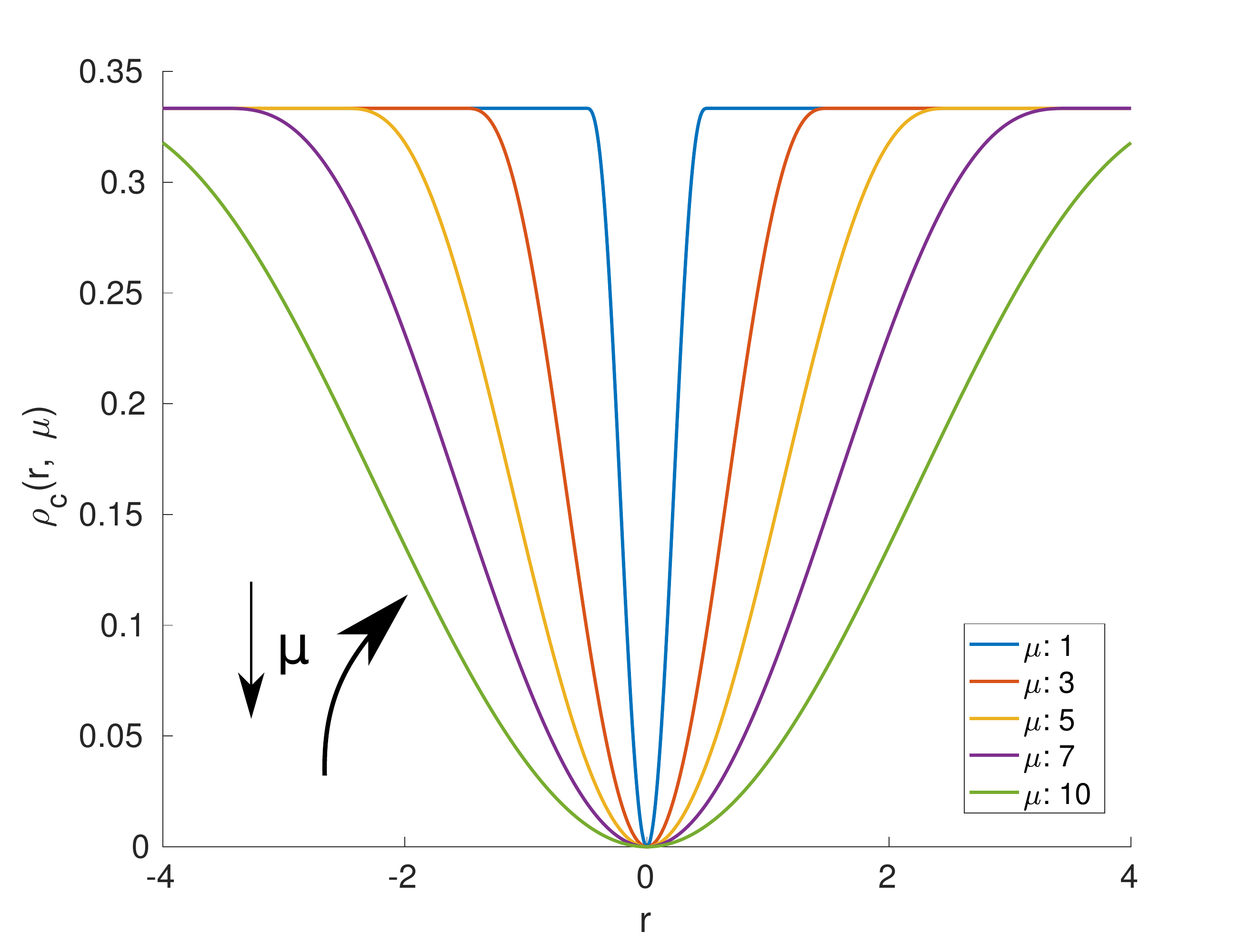}
     \caption{}
     \label{fig:tukey-loss-mu}
\end{subfigure}
\caption{Tukey's biweight function for different $\overline{c}$ \eqref{eq:tukey-fcn} (a);
and its associated GNC functions for varying 
$\mu$ \eqref{eq:tukey-fcn-mu} (b). 
In (b), we depict $\overline{c} = 0.5$.
}
    \label{fig:tukey-loss}
\end{figure}

It is known that the chosen loss function $\opLoss$ 
has a great impact on the performance of the algorithm.
Thus, although convex functions exist, such as Huber or $\ell_1$,
these have a low breakdown point and are still sensible to gross outliers; 
non-convex loss functions are usually preferred. 
Following the approach in~\cite{yang2020graduated},
we define the Tukey's biweight  loss function $\rho(\bullet)$ 
in terms of a threshold $\overline{c} \in \Reals{}$ as

\begin{equation}
    \label{eq:tukey-fcn}
    \opLossC (r) =
    \begin{cases}
    & \frac{r^2}{\cSq} - \Big(\frac{r^2}{\cSq}\Big)^2 + \frac{1}{3} \Big( \frac{r^2}{\cSq} \Big) ^3, \quad | r | \leq \overline{c} \\
    & \frac{1}{3}, \quad \text{ otherwise }
    \end{cases},
\end{equation}
where $\overline{c}$ determines the shape of the function,
as it is shown in Figure~\eqref{fig:tukey-loss-c-hat}.
The variable $\overline{c}$ thus controls what points we are considering as inliers;
for $\overline{c}$ large, all the points are considered as such (including outliers),
while in the opposite extreme, 
only these points with very low residuals are considered as inliers.

Since the loss function is non-convex,
Problem \eqref{eq:robust-original-problem} is prone to get trapped in local minima.
The first step towards the efficient resolution of 
Problem \eqref{eq:robust-original-problem}
leverages
the Black-Rangarajan duality
between robust and line processes~\cite{black1996unification}.
By introducing a set of slack variables or \emph{weights}, $w_i \in [0, 1]$,
one for each correspondence and
an \emph{outlier process} term $\outlierProc$,
we can re-formulate any robust estimation problem
as
\begin{equation}
    \label{eq:GNC}
    \min_{\substack{\Essential \in \SetEssentialMatrices, \\ w_i \in [0, 1], i=1, \dots N}}
    \sum_{i = 1} ^N w_i \epsilon_i^2 + \outlierProc.
\end{equation}
The term $\outlierProc$ that penalizes the weight $w_i$ 
usually admits an analytical expression 
for most common loss functions 
(see \cite{black1996unification}).

However, \eqref{eq:GNC} is still non-convex.
To avoid this behavior, we leverage the tool known as Graduated Non-convexity (GNC)
that allows to optimize a generic non-convex function
by solving a sequence of surrogate problems of increasing non-convexity.
This is achieved by introducing a surrogate function $\opGNC$ controlled
by a parameter $\mu$ such that, for some $\mu$ the function is convex
and, in the limit, we obtain the original non-convex function $\opLoss(\bullet)$.
We start by solving the initial convex problem and
update the parameter $\mu$,
thus increasing the non-convexity of the next problem to solve,
till the original problem is attained.
The variable estimated for the $i$-th problem 
is employed as initialization for the $(i+1)$-th problem.

For the Tukey's biweight function in \eqref{eq:tukey-fcn},
the surrogate function with the control parameter $\mu$ is given as
\begin{equation}
    \label{eq:tukey-fcn-mu}
    \opLossC (r, \mu) =
    \begin{cases}
    & \frac{r^2}{\mu  \cSq} - \Big(\frac{r^2}{\mu  \cSq}\Big)^2 + \frac{1}{3} \Big( \frac{r^2}{\mu \cSq} \Big) ^3, \quad | r | \leq \sqrt{\mu} \overline{c} \\
    &\frac{1}{3},  \quad \text{ otherwise }
    \end{cases}.
\end{equation}

For large $\mu$, the function $\opLossC (r, \mu)$ is convex and 
returns the original function \eqref{eq:tukey-fcn} 
when $\mu = 1$ 
(see Figure~\eqref{fig:tukey-loss-mu}).
The outlier process $\outlierProc$ in
\eqref{eq:GNC} for any GNC surrogate with parameter $\mu$ is defined as
\begin{equation}
    \outlierProcMu = \frac{\mu \cSq} {3} (1 - \sqrt{w_i}) ^2 (1 + 2 \sqrt{w_i}).
\end{equation}

\smallskip
\textbf{Overview of the Robust Estimation}

With these tools at hand,
we are now able to solve the non-convex problem \eqref{eq:robust-original-problem}.
At each outer iteration, we update the control parameter $\mu$ and optimize
\begin{equation}
    \min_{\Essential \in \SetEssentialMatrices} \sum_{i = 1} ^N \opLossC \big( \epsilon_i , \mu  \big ).
\end{equation}
with $\opLossC(\epsilon_i, \mu)$ given by \eqref{eq:tukey-fcn-mu}.
By the Black-Rangarajan duality, we re-formulate this surrogate problem
as the joint optimization over $\Essential$ and $ \{w_i\}_{i=1}^N$:
\begin{equation}
    \label{eq:GNC-Tukey}
    \min_{\substack{\Essential \in \SetEssentialMatrices, \\ w_i \in [0, 1], i=1, \dots N}}
    \sum_{i = 1} ^N w_i \epsilon_i^2 + \outlierProcMu.
\end{equation}

Problem \eqref{eq:GNC-Tukey} is actually solved by alternating optimization.
At each inner iteration, we optimize over $\Essential$ with the previous $\{w_i\}$
and then, we optimize over $\{w_i\}$  with fixed $\Essential$.
See that in the first optimization, the second term in \eqref{eq:GNC-Tukey}
does not depend on $\Essential$ and can be dropped.
Hence, this first problem is the weighted version of the non-minimal presented in
\eqref{eq:optim-on-manifold}.
The optimization over $\{w_i\}$ with fixed $\Essential$ 
can be solved in closed-form for the loss function considered here.
The weight $w_i$ is computed as:
\begin{equation}
    w_i
    =
    \begin{cases}
    0 & \quad \frac{\epsilon_i^2}{\mu \overline{c}^2} > 1 \\
    \Big ( 1 - \frac{\epsilon_i^2}{\mu \overline{c}^2}  \Big) ^2 & \quad \text{ otherwise }
    \end{cases}.
    \label{eq:closed-form-weights}
\end{equation}
Note that $w_i \in [0, 1]$ since $\frac{\epsilon_i^2}{\mu \overline{c}^2} \geq 0$.

This process, which is summarized in Algorithm~\eqref{alg:GNC},
is repeated for decreasing values of $\mu$ until it is equal to one.
This decreasing rate is given by the parameter $\tau_{\mu} > 1.0$.
Although Algorithm~\eqref{alg:GNC} only requires as input the set of correspondences,
we can also provide the initialization for the local, iterative optimization algorithm
that updates the variable $\Essential$
and the initial weights associated the correspondences.

 \begin{algorithm}[t!]
    \caption{Robust Estimation of the Essential Matrix} \label{alg:GNC}
    \KwData{Set of correspondences $\{(\obsi, \obsip)\}_{i=1}^N$}
    \KwResult{ Estimated essential matrix $\Essential$; set of inliers $\mathcal{I}$ (weights);
     \textsc{isValid} $\in \{\text{True}, \text{False}\}$}
      Initialize $\mu = 6000$\;

    \tcp{Outer iteration}
    \Repeat {$\mu = 1$ or convergence}{
        \tcp{Inner iteration}
        \Repeat {convergence or max. iterations}
        {
              Update $\Essential$ with weighted problem \eqref{eq:optim-on-manifold} \;
             Update weights $w_i, i=1, \dots, N$ in closed-form \eqref{eq:closed-form-weights} \;
         }
        $\mu \leftarrow \mu / \tau_{\mu}$ \;
    }
     {Set inliers: $\mathcal{I} = \{i \ \vert w_i > \tau_{w}, i=1, \dots, N \}$} \;

    \eIf{$|\mathcal{I}| \geq N_0$}
    { {Refine estimation \eqref{eq:optim-on-manifold} with set of inliers } \;
     {\textsc{isValid} $=$ True} \;
    }
    {\tcp{Number of inliers is not enough}
     {\textsc{isValid} $=$ False        }\;
    }
    \end{algorithm}

\smallskip
\textbf{Implementation details}:
In practice, we fix the rate for $\mu$ in $\tau_{\mu} = 1.10$
and stop the algorithm if the difference between two consecutive
cost values for \emph{outer} iterations lies below $10^{-6}$.
The same threshold is employed to break the inner loop,
although it does not stop the estimation.
We allow two iterations for the inner loop and 500 for the outer.
We define the shape parameter as $\overline{c}^2 = 10^{-5}$
and consider as inlier any correspondence whose weight is greater than $\tau_{w} = 0.9$.
In the last step, we refine the estimated essential matrix
only with the set of inliers.
To avoid ill-posed problems,
we required that at least there exist $N_0 = 12$ inliers. 
We apply our certifier to this last problem and 
solution defined only with the inliers.

A fast, \cpp~implementation of our proposal 
is publicly available in \\ 
\url{https://github.com/mergarsal/FastCertRelPose.git}. 
Bindings for \matlab and \python 
are also provided.
As a by-product of our contribution,
we also release a lightweight template library in \\ 
\url{https://github.com/mergarsal/GNCSO.git}  \\ 
for the robust algorithm in \eqref{alg:GNC},
that currently\footnote{Date of this document}
includes Tukey's biweight, Welsch (Leclerc), Geman-McClure and TLS loss functions,
and admits any non-minimal o minimal solver for the variable estimation step.
This library also wraps around the \emph{Optimization} library by Rosen \etal \cite{rosen2019se}
that allows to perform local optimization on Riemannian manifold and convex spaces.

\section{Experimental Validation} \label{sec:experiments}
In this last Section, we empirically showcase the utility of the proposed certifiable pipeline
and the improvements presented in this work
through an extensive set of experiments 
on both synthetic (Section \eqref{sec:synt})  
and real data (Section \eqref{sec:exp-real}). 
Before that, 
we compare the proposed certifiable 
pipeline   
against the state-of-the-art  
certifiable approaches  
proposed for the \RPp 
\xspace (Section \eqref{sec:comp-cert}). 
All the experiments were performed 
in the same machine (PC) with: 
CPU intel i7-4702MQ, 2.2GHz 
and RAM 8 GB.

\subsection{Comparison against certifiable solvers} 
\label{sec:comp-cert}
In this Section we compare 
the performance in 
terms of accuracy and computational cost 
of the proposed solver 
against the state-of-the-art certifiable approaches. 

As we introduce in Section~\eqref{sec:related}, 
certifiable solvers for the \RPp \xspace 
exist that rely on 
Branch-and-Bound optimization and 
convex relaxations, 
the latter solved via Interior Point Algorithms (IPM) 
with off-the-shelf tools. 
The first set of solvers 
present worst-case exponential time complexity 
due to their exploration 
of the feasible set. 
As it has been reported in 
the original papers \cite{hartley2007global, kneip2013direct}, 
and recent works \cite{Zhao2019}, 
this leads to computational times 
of seconds 
or even minutes. 
We will not examine further 
this type of solvers 
since 
the other approaches based on convex relaxations 
have been shown to be faster 
with polynomial time complexity. 
The complexity depends 
on the number of variables and constraints, 
which can be empirically observed 
when comparing, 
for example, 
\cite{briales2018certifiably} 
($40$ variables and $536$ constraints) 
and \cite{Zhao2019} 
($12$ variables and $7$ constraints).  
We perform a set of experiments 
with both solvers 
under a \matlab implementation 
(using \sdpt~\cite{toh1999sdpt3} as IPM), 
and observe that 
\cite{briales2018certifiably} takes  1-2 seconds, 
while \cite{Zhao2019} remains under 0.5 seconds. 
For the same set of experiments and under a matlab implementation, 
our proposal roughly takes $50$ milliseconds 
(including variable estimation). 

\begin{figure}
\begin{subfigure}{0.48\textwidth}
     \includegraphics[width=0.9\textwidth]{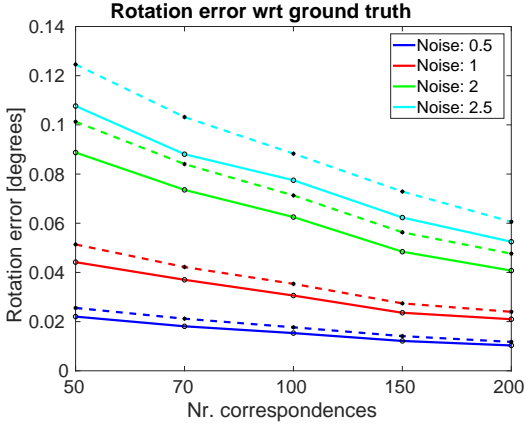}
     \caption{Error in rotation (degrees) between 
     the returned solution by our solver 
     (solid line) 
     and the SDP 
     (dashed line) 
     \wrt the ground truth 
     in terms of the number of correspondences 
     (X-axis). 
     Legend shows the noise level 
     for each graphic.}
     \label{fig:error-ess-cert}
\end{subfigure}
\begin{subfigure}{0.48\textwidth}
     \includegraphics[width=0.9\textwidth]{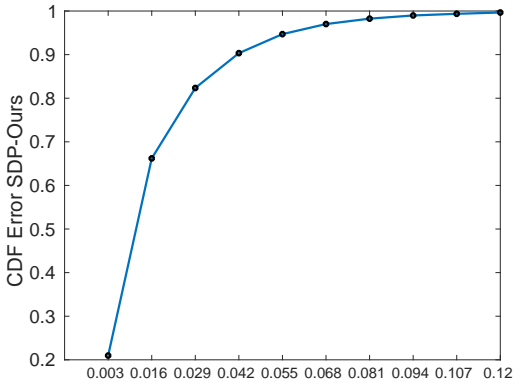}
     \caption{Probability distribution of the error 
     between the rotation returned by our solver 
     and the one returned by the SDP 
     for all the experiments 
     and number of correspondences.}
     \label{fig:cdf-cert}
\end{subfigure}
\caption{Comparison of our solver and the SDP in \cite{Zhao2019}. 
Figures show the results (see captions) 
for the experiments 
with noise $\sigma = \{0.5, 1.0, 2.0, 2.5\}~\pixels$ 
and correspondences $N = \{50, 70, 100, 150, 200\}$. }
\end{figure}

From the above discussion 
we can derive 
that the faster certifiable solver 
is the SDP proposed in \cite{Zhao2019}. 
We next perform a comparison 
against this SDP;  
we employ the released code by the author
\footnote{
\url{https://github.com/jizhaox/npt-pose}
}. 
We perform a set of experiments 
on synthetic data 
(see Section~\eqref{sec:exp-synt-params} 
for a detailed explanation of the 
data generation). 
We observe, 
as it was pointed out by the author, 
that 
the relaxation does not perform 
always well with low number of correspondences 
and/or large noise. 
For the sake of clarity, 
we restrict the noise to 
$\sigma = \{0.5, 1.0, 2.0, 2.5\}$ pixels 
and number of correspondences in 
$N = \{50, 70, 100, 150, 200\}$; 
for each configuration, 
we generate $500$ random problem instances. 
We measure the distance between the returned 
rotation matrices 
\wrt the ground truth 
in terms of geodesic distance. 
Figure~\eqref{fig:error-ess-cert} 
shows the mean errors 
for each configuration of parameters. 
We also measure the distance between 
the solutions 
\wrt each other 
and plot the probability distribution 
of errors in Figure~\eqref{fig:cdf-cert}. 
Observe that for the considered range 
of observations-noise, 
both certifiable solvers 
return similar solutions, 
while our proposal attains lower errors 
in rotation. 
This difference can be partially 
explained by the 
procedure that the SDP 
employs to extract the solution, 
that may have associated some rounding errors. 
While these differences are small, 
the computational time required by 
each solver does differ: 
the SDP requires 
$5$ milliseconds 
and 
the proposed pipeline 
takes only $500$ microseconds 
in average. 
That is, 
while the solution obtained 
by our proposal is similar 
in all occasions 
(if not better), 
reflecting in turn that 
the iterative solver does attain 
the global optimum repeatedly, 
the certifiable pipeline  
is at least ten times 
faster than the SDP.

From a implementation point of view, 
we should also remark  
that our certification algorithm 
does not require specific tools, 
as the convex relaxations do. 
The proposed solver 
allows to estimate 
the variable by any means
and 
then, 
it certifies the given solution. 
The latter requires the resolution 
of a linear system in six variables 
and the computation of two eigenvalues 
(one of them can be computed in closed-form), 
operations that any mathematical library 
provide with.

\begin{figure*}
\begin{subfigure}{0.24\textwidth}
     \hspace*{-1.2cm}
     \includegraphics[width=\BigfigureSize]{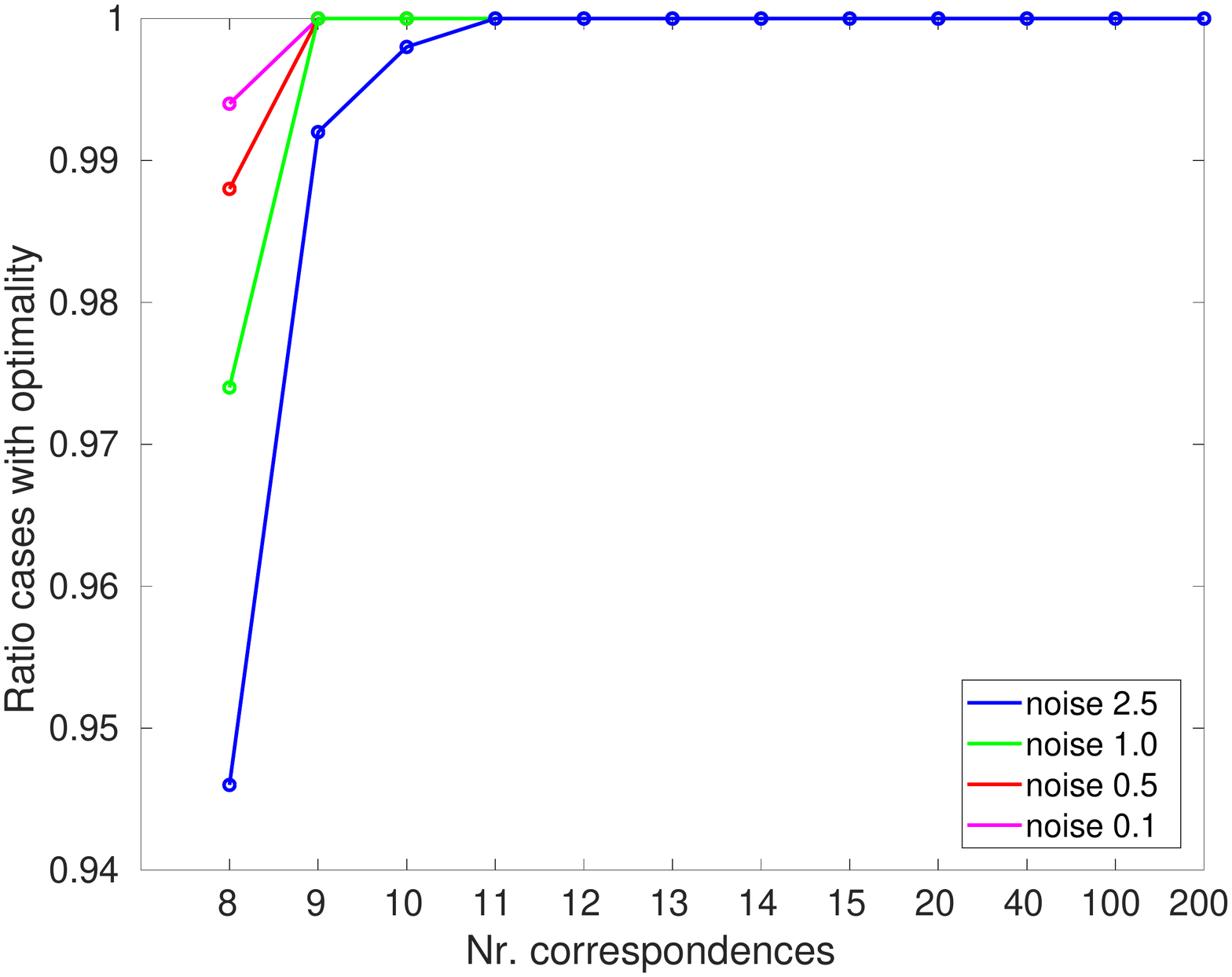}
     \caption{\hspace*{1.6cm}}
     \label{fig:opt-cases-noise}
\end{subfigure}
\begin{subfigure}{0.24\textwidth}
     \hspace*{-0.7cm}
     \includegraphics[width=\BigfigureSize]{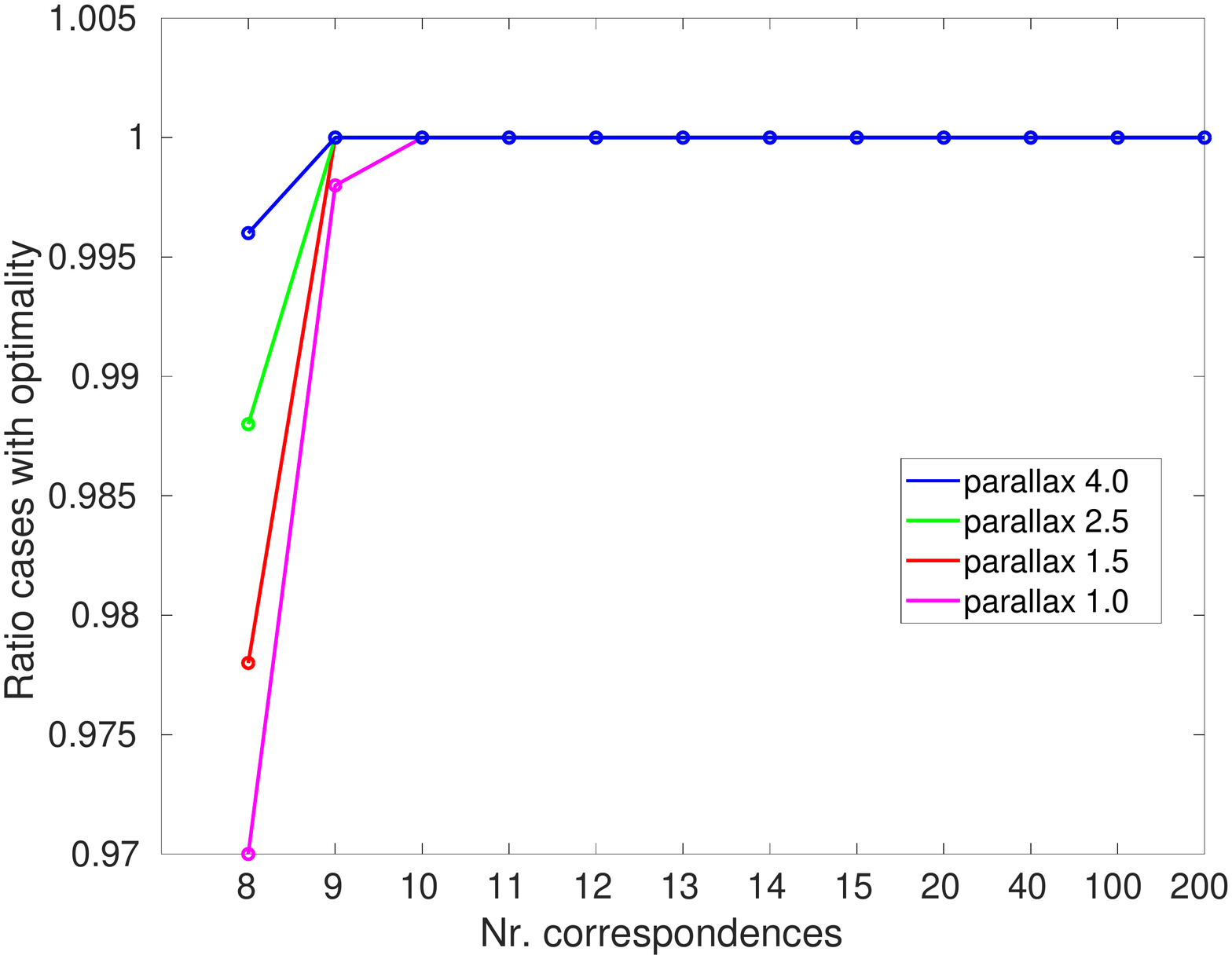}
     \caption{\hspace*{-0.2cm}}
     \label{fig:opt-cases-parallax}
\end{subfigure}
\begin{subfigure}{0.24\textwidth}
     \hspace*{-0.1cm}
     \includegraphics[width=\BigfigureSize]{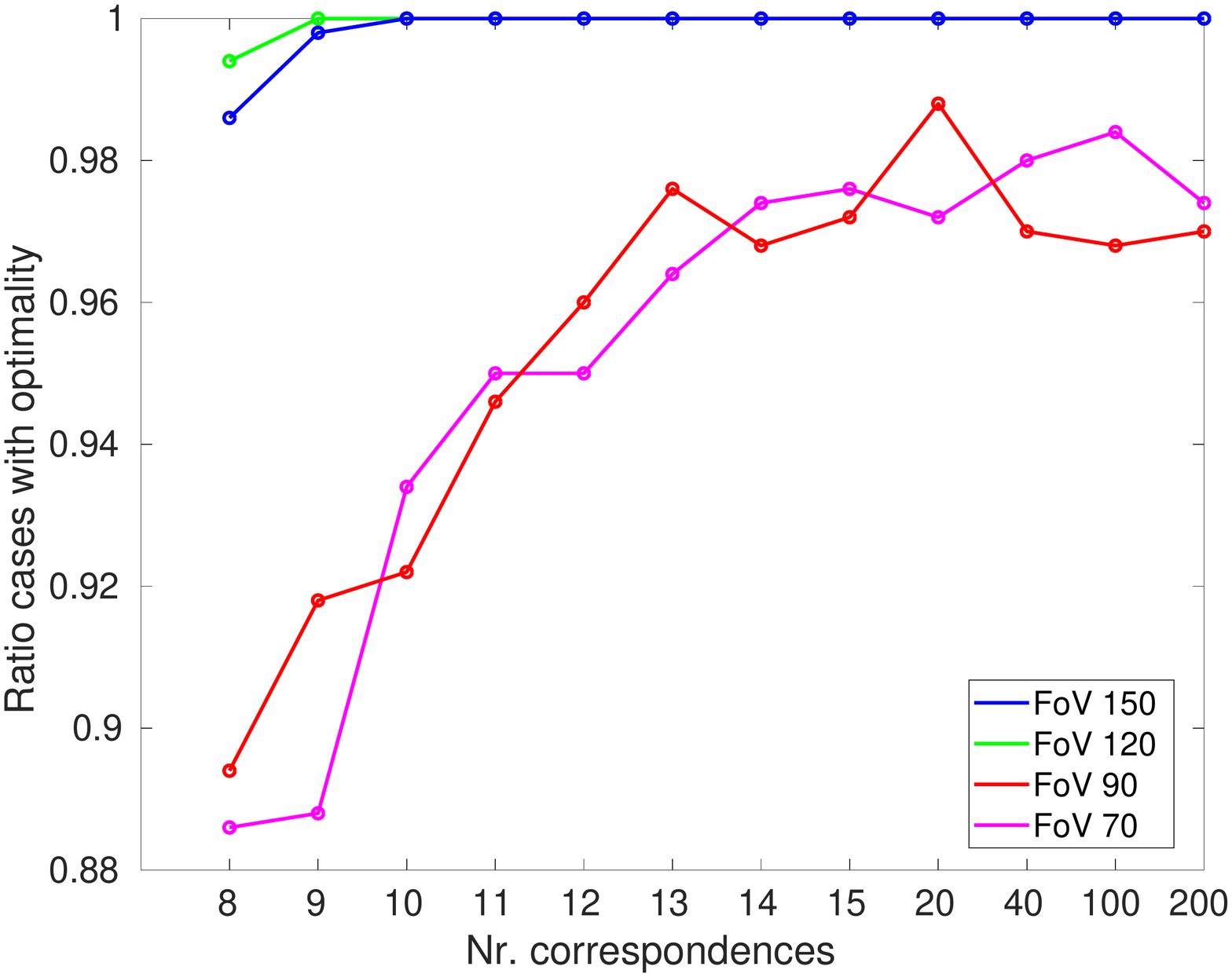}
     \caption{\hspace*{-1.5cm}}
     \label{fig:opt-cases-fov}
\end{subfigure}
\begin{subfigure}{0.24\textwidth}
     \hspace*{0.3cm}
     \includegraphics[width=\BigfigureSize]{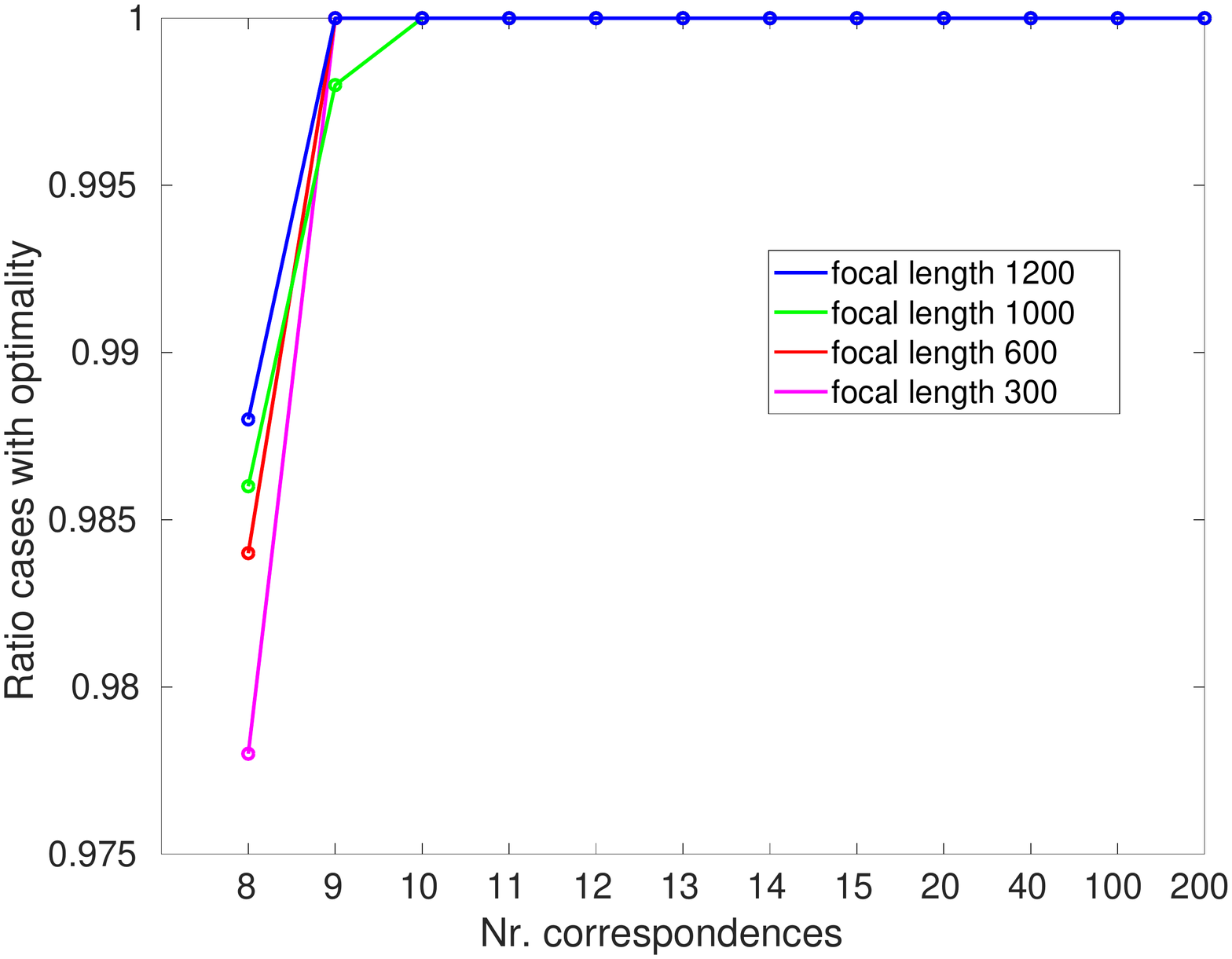}
     \caption{\hspace*{-2.cm}}
     \label{fig:opt-cases-focal}
\end{subfigure}

\begin{subfigure}{0.24\textwidth}
     \hspace*{-1.2cm}
     \includegraphics[width=\BigfigureSize]{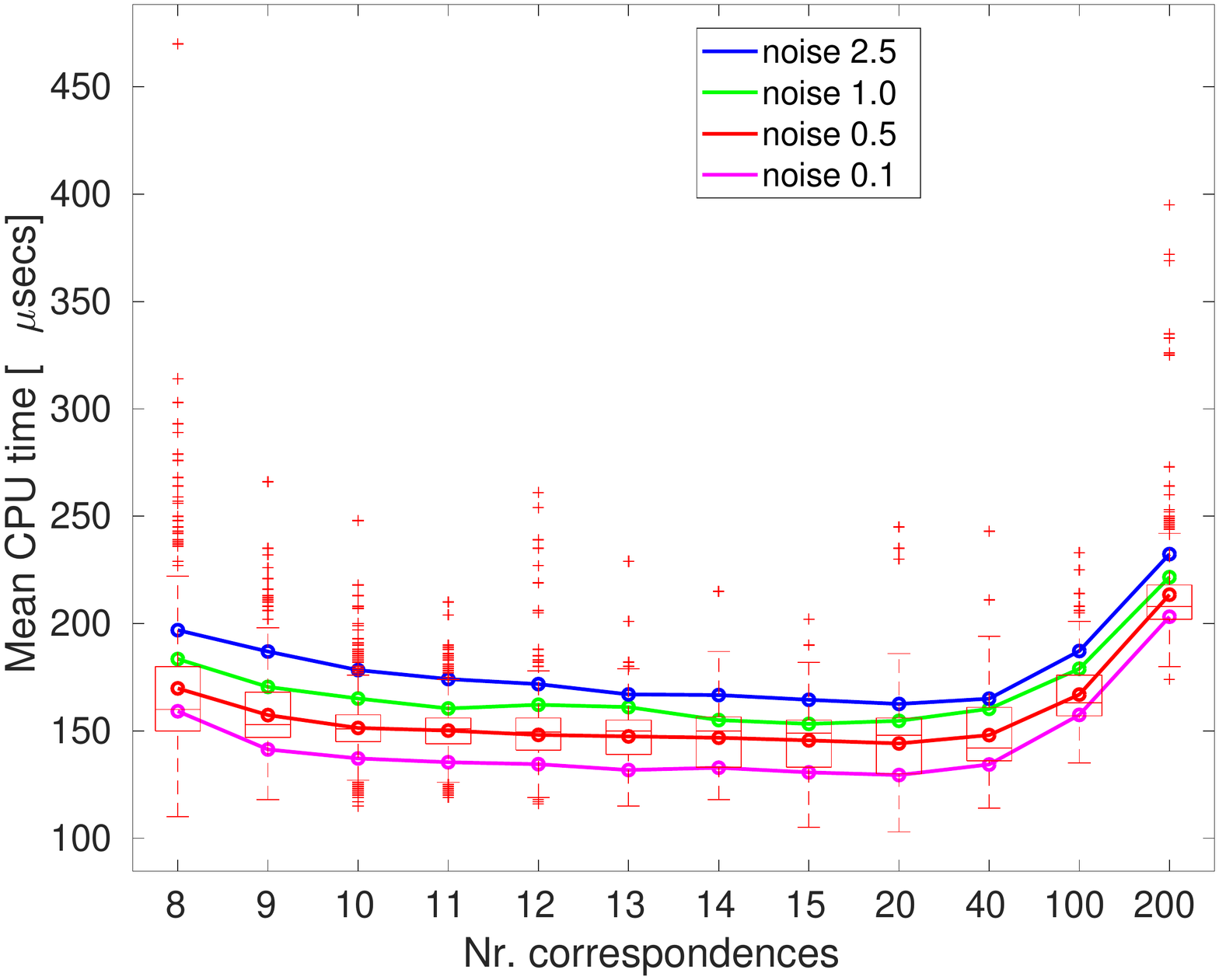}
     \caption{\hspace*{1.6cm}}
     \label{fig:time-noise}
\end{subfigure}
\begin{subfigure}{0.24\textwidth}
     \hspace*{-0.7cm}
     \includegraphics[width=\BigfigureSize]{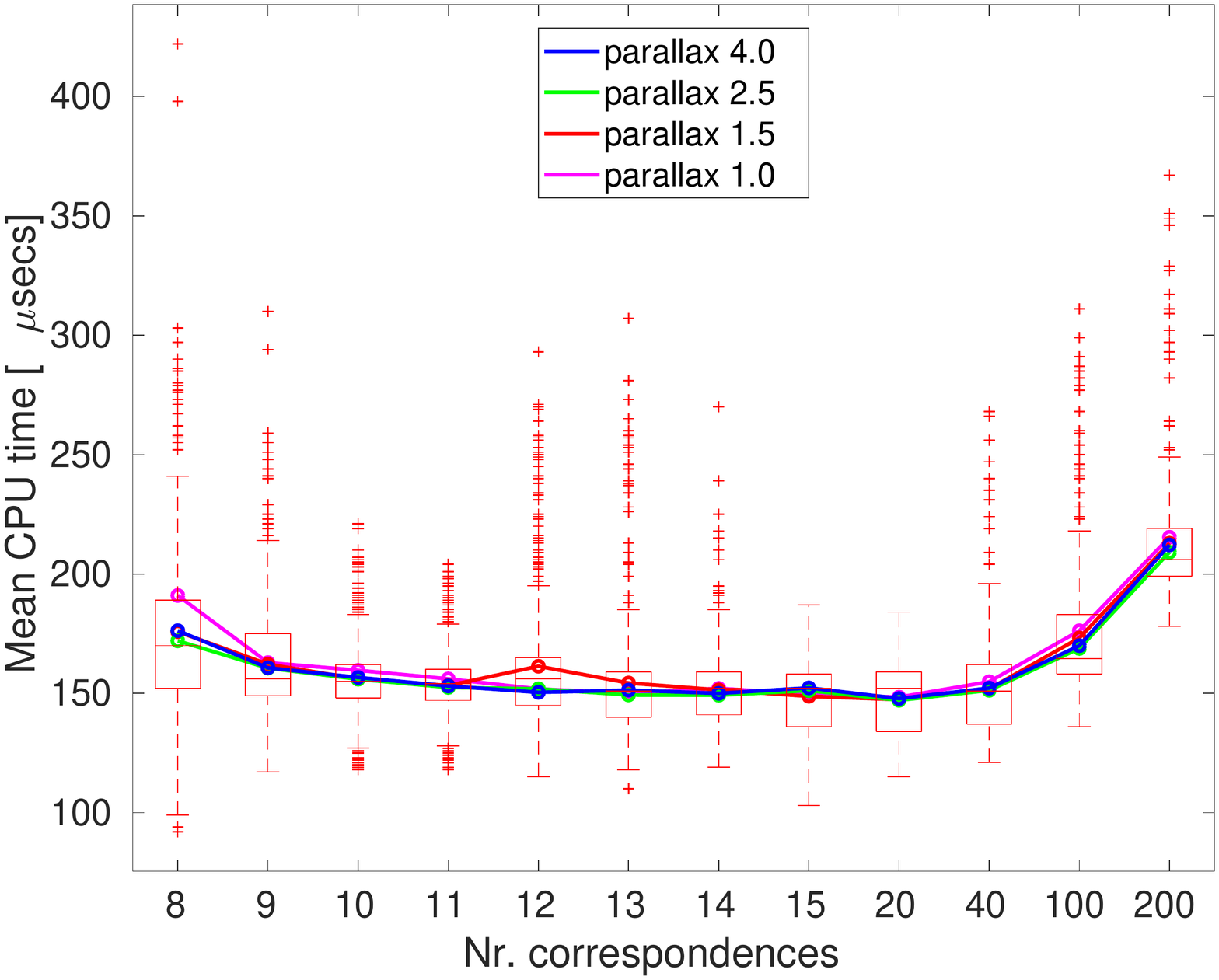}
     \caption{\hspace*{-0.2cm}}
     \label{fig:time-parallax}
\end{subfigure}
\begin{subfigure}{0.24\textwidth}
     \hspace*{-0.1cm}
     \includegraphics[width=\BigfigureSize]{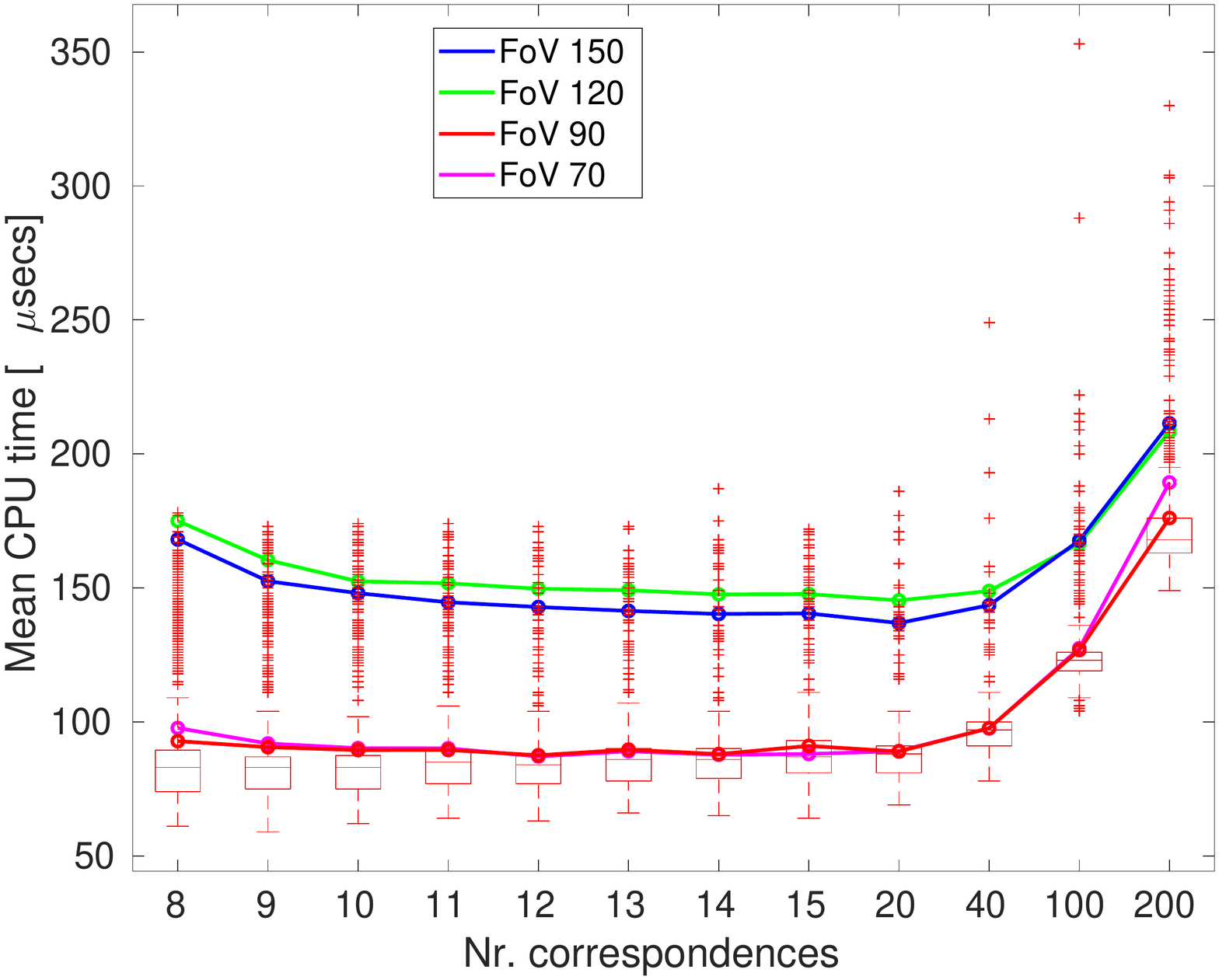}
     \caption{\hspace*{-1.5cm}}
     \label{fig:time-fov}
\end{subfigure}
\begin{subfigure}{0.24\textwidth}
     \hspace*{0.3cm}
     \includegraphics[width=\BigfigureSize]{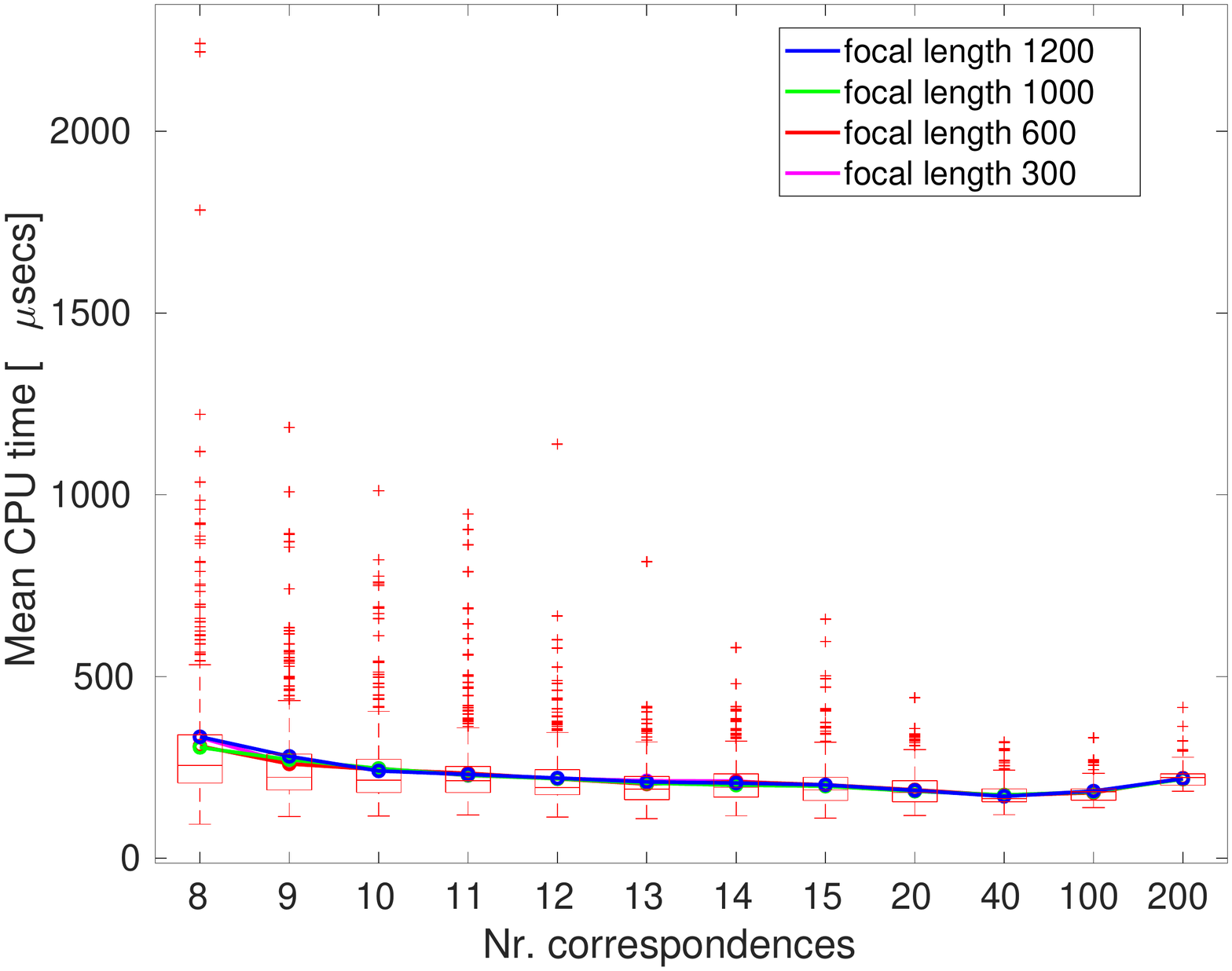}
     \caption{\hspace*{-2cm}}
     \label{fig:time-focal}
\end{subfigure}
    \caption{Percentage of cases (first row) in which we could certify the optimality of the solution (note the difference for the Y-axis scale between figures);
    and averaged computation time (second row, solid line) in $\mu secs$ required by the whole proposed pipeline. We include (second row) the boxplots with the results for noise $0.5$ \pixels,~maximum parallax $1.5$~\meters~,~\FOV~$90$~\degrees~and focal length $f = 600~\pixels$.}
    \label{fig:exp-influence-data}
\end{figure*}

\subsection{Experiments on Synthetic data} 
\label{sec:synt}
We design four types of experiments,
each of them target to show a different aspect of our proposal:
\begin{itemize}
    \item In this first set (Section~\eqref{sec:exp-synt-params})
    we show the performance of the current implementation
    with the simplification of the essential matrix manifold introduced
    in Section~\eqref{sec:essential-manifold}.
    We also prove empirically that the proposed preconditioner
    in Section~\eqref{sec:essential-manifold}
    does reduce the number of iterations of the RTR solver.
    \item In this second set (Section~\eqref{sec:exp-synt-comparison})
    we compare the proposed certifiable pipeline
    with state-of-the-art methods~\cite{kneip2013direct}
    in both performance and computation time.
    \item In this third set (Section~\eqref{sec:exp-synt-BA})
    we compare the performance of Bundle-adjustment (minimization of reprojection error)
    when initialized with minimal solvers and our proposal.
    \item In this fourth set (Section~\eqref{sec:exp-synt-GNC})
    we introduce outliers and
    compare the performance of our robust pipeline (Section~\eqref{sec:GNC})
    against minimal solvers embedded into a RANSAC framework,
    which can be considered as the most common approach to robustification.  
\end{itemize}

\smallskip
We employ the same procedure to generate the random data for all our synthetic experiments,
which is the same procedure used in previous works
~\cite{garcia2020certifiable}, \cite{briales2018certifiably}. 
We summarize it here for completeness:
We place the first camera frame at the origin
(identity orientation and zero translation) and
generate a set of random 3D points within a frustum
with depth ranging from one to eight meters measured from the first camera frame and inside its Field of View (FoV).
Then, we generate a random pose for the second camera
whose translation magnitude is bounded above by $\maxParallax$.
We also enforce that all the 3D points lie within the second camera's FoV.
We create the correspondences as unit bearing vectors and
add noise by assuming a spherical camera,
computing the tangential plane at each bearing vector
(point on the sphere) and
introducing a random error sampled from the standard uniform distribution,
considering a focal length of 800 pixels for both cameras.

\medskip
\subsubsection{Performance of the current implementation without outliers} \label{sec:exp-synt-params}

In order to showcase the performance of our proposal and
the computational improvements with respect to our previous work~\cite{garcia2020certifiable},
we design three batch of experiments:

\smallskip
\textbf{Influence of Parameters}:
We show first the performance of the proposed certifiable pipeline for instances of the \RPp\xspace with different parameters and
initialized with the \eightpt~algorithm.
We use the following default parameters and
change one of them in each batch of experiments:
\FOV~to $100~\degrees$,
translation parallax $\maxParallax = 2.0 \meters$, 
focal length of $f = 800~\pixels$ and
noise level $0.5~\pixels$.
We consider four level of noise 
$\sigma = \{0.1, 0.5, 1.0, 2.5\}~\pixels$,
four values for the maximum parallax  
$||\trans||_2 = \{1.0, 1.5, 2.5, 4.0\}~\meters$, 
four \FOV~angles $\FOV~\in \{ 70, 90, 120, 150 \}~\degrees$ (focal length fixed) 
and four focal lengths $f \in \{300, 600, 1000, 1200\}~\pixels$ (image size fixed).
For each configuration of parameters,
we generate random problem instances as explained above
with number of correspondences in $N \in \{ 8, 9, 10, 11, 12, 13, 14, 15, 40, 100, 150, 200 \}$.
Further, for each combination of parameter/number of correspondences, 
we generate $500$ random instances.

Figures~\eqref{fig:opt-cases-noise},\eqref{fig:opt-cases-parallax},\eqref{fig:opt-cases-fov} 
and \eqref{fig:opt-cases-focal} 
show the percentage of cases with \emph{certified} optimal solutions 
(Algorithm~\eqref{alg:certification}) 
for the experiments with different level of noise, parallax, 
\FOV~and focal length, respectively.
We plot the averaged computation time in $\mu secs$ for the three batches of experiments
in Figures~\eqref{fig:time-noise},\eqref{fig:time-parallax}, \eqref{fig:time-fov} and \eqref{fig:time-focal}.
We also include the results as boxplots 
\footnote{
All the boxplots that appear in this manuscript 
follow the same (matlab) format: 
central mark indicates median, the bottom and top edges indicate
the 25th and 75th percentile, respectively and the whiskers denoted the data that is not
considered as outlier. The latter are plotted individually with the ‘+‘ symbol.
}
for the experiments with noise $0.5~\pixels$,
maximum translation magnitude of $1.5~\meters$, 
$\FOV~90~\degrees$ 
and focal length $f = 600~\pixels$~(same color than the mean values).

The simplified characterization of the essential matrix manifold chosen for this implementation 
(Section~\eqref{sec:essential-manifold})
does not seem to affect the overall performance of the proposed certifiable pipeline,
as it is reflected by the high percentage of cases with certified optimal solutions.
As expected, the cases with low number of correspondences (up to $9-11$) present
the worst behavior,
although the percentage of optimal solutions goes up to $94\%$. 
From scenes with more than $11$ correspondences,
we obtain $100\%$ 
of certified cases for all the different parameters (noise, translation and \FOV),
except those with narrow \FOV~($70-90\ \degrees$).
For the latter, we observe that with more than $12$ correspondences,
we are able to certify as optimal more than $90\%$ of the solutions.
Further, we see that the averaged computation time for all the instances is under $250 \ \mu secs$,
even in challenging instances with high noise, narrow \FOV~and small focal length 
(a large signal-to-noise ratio).

\smallskip
\textbf{Influence of Initialization}:
Here we show the influence of the initialization on the performance of the proposed pipeline.
We generate random instances with the default parameters given above and
initialize the pipeline with the trivial identity matrix (\textsc{Iden}),
a random matrix (\textsc{Rand}),
the estimation obtained with the $5$-point algorithm (\fivept)~\cite{nister2004efficient},
with the $7$-point algorithm (\sevenpt)
and the $8$-pt algorithm (\eightpt).
We generate $500$ problem instances.

\begin{figure}
    \begin{subfigure}{0.49\textwidth}
     \centering
     \includegraphics[width=0.9\textwidth]{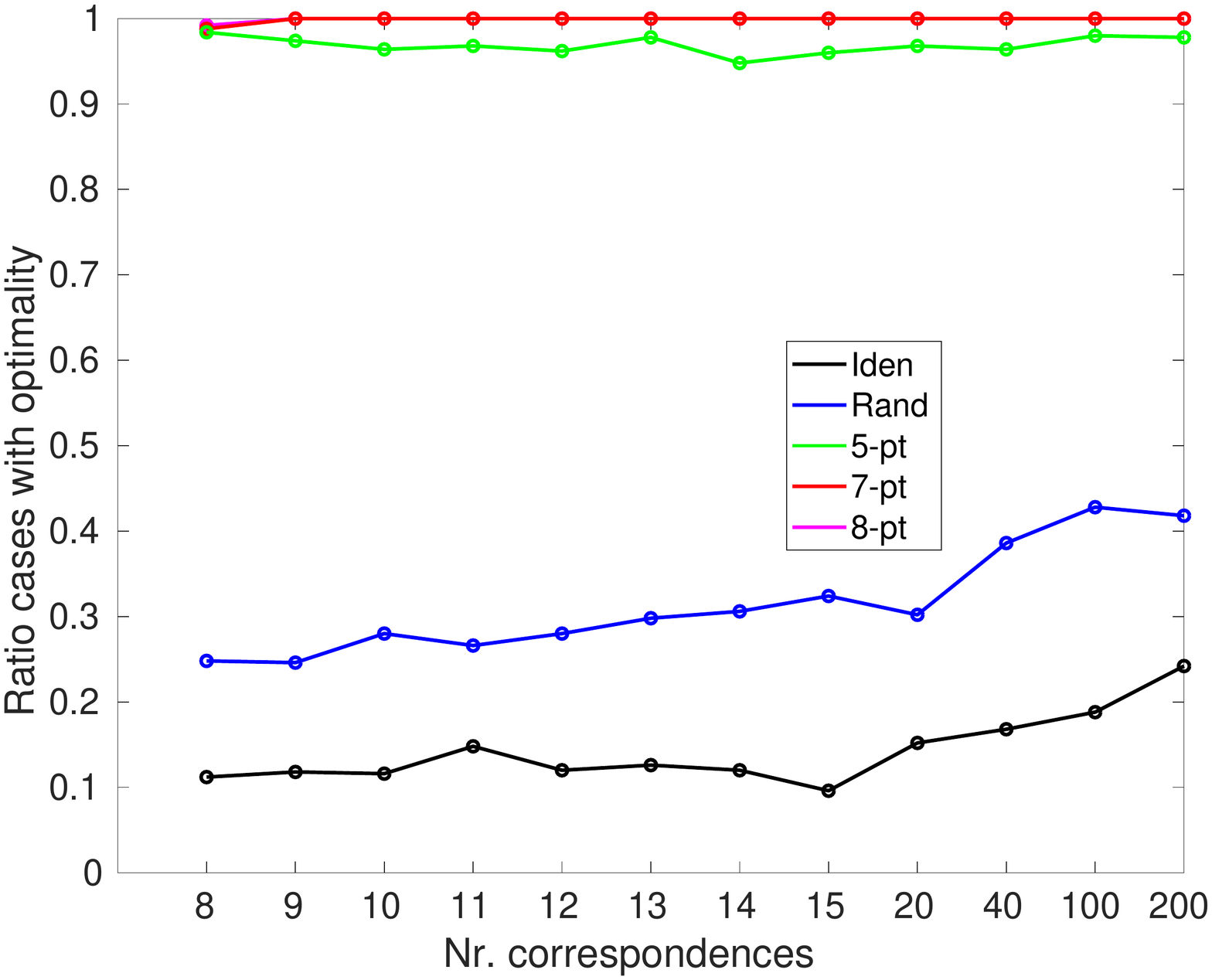}
     \caption{}
     \label{fig:opt-cases-init}
\end{subfigure}
\begin{subfigure}{0.49\textwidth}
     \centering
     \includegraphics[width=0.9\textwidth]{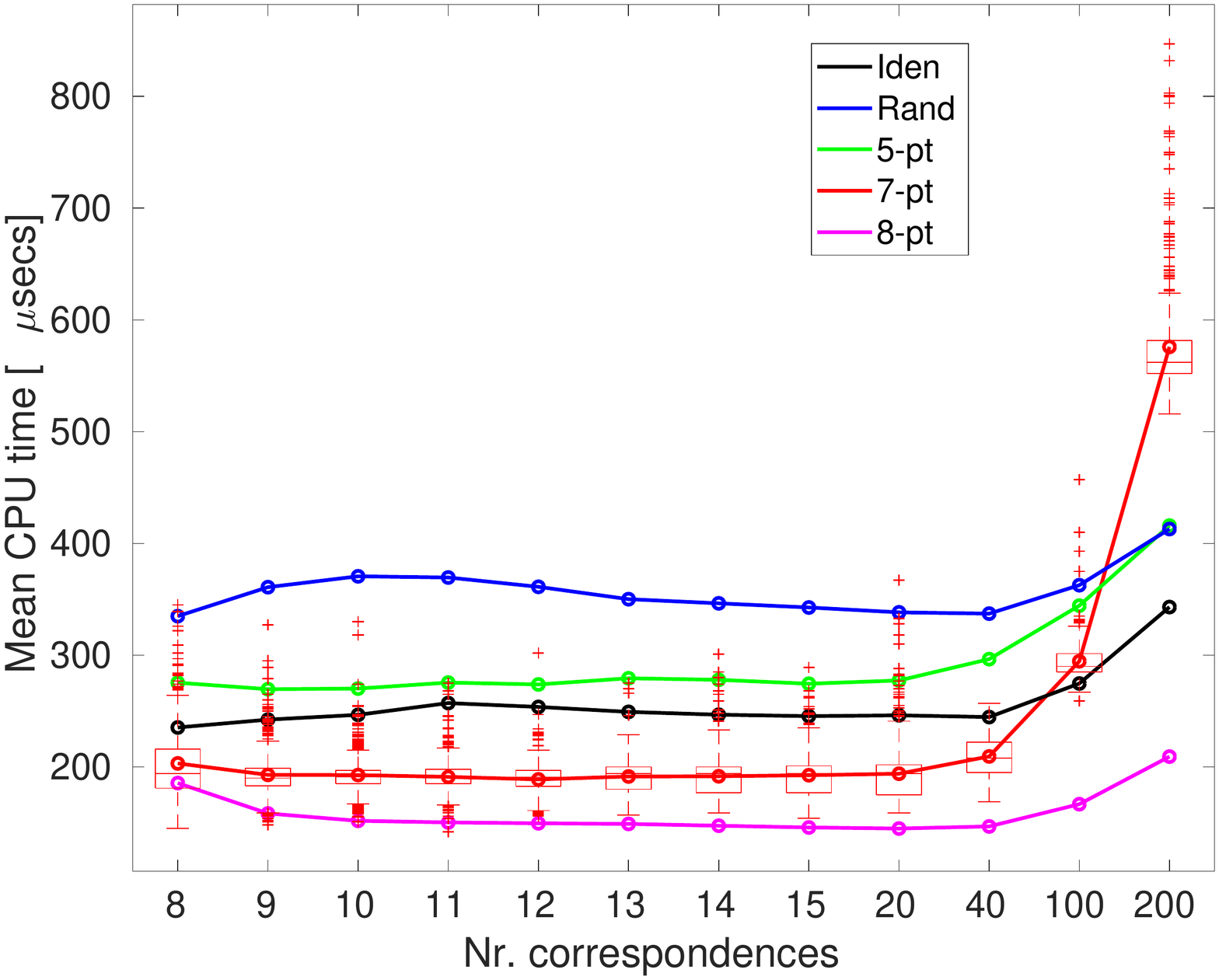}
     \caption{}
     \label{fig:time-init}
\end{subfigure}
    \caption{Percentage of cases (\eqref{fig:opt-cases-init}) in which we certify the solution as optimal 
    (the graphics for \sevenpt and \eightpt overlap);  
    and averaged computation time (solid line, \eqref{fig:time-init}) required by the certifiable pipeline. We include as boxplot the time for the optimizations initialized with the \sevenpt (boxplot, \eqref{fig:time-init}). 
    Observe that the instability of the \fivept may affect 
    the performance of the iterative solver, 
    which is reflected by the percentage of optimal solutions 
    not being $100\%$ in \eqref{fig:opt-cases-init}. 
    Figures reported the experiments with default parameters 
    (noise $0.5~\pixels$ and focal length $800~\pixels$). 
    }
    \label{fig:times-init}
\end{figure}

Figure~\eqref{fig:opt-cases-init} depicts the percentage of cases with certified optimal solution.
We see that the ratio of optimal solutions
from the \fivept, \sevenpt and \eightpt are similar ($>90\%$),
while being slightly higher for the last two initializations.
On the other hand,
the solutions obtained with the random and
identity matrices tend to percentages below $40\%$ and $30\%$, respectively.
This last result reflects a considerable difference with respect to~\cite{garcia2020certifiable},
in which we obtain an increasing percentage of optimal solutions (up to $95\%$)
with the number of correspondences for these initializations.
The simplification of the essential matrix manifold in this work,
while it does not seem to affect the refinements 
initialized with the \fivept, \sevenpt or \eightpt algorithms,
it hinders the optimizations with poor initializations.
In practice, however, one should never initialize the algorithm with these matrices;
even a ``bad`` estimation (\eg high noise and \eightpt)
is shown to attain the global optimum with high probability.
In terms of computational time, we observe that for all the cases the complete
certifiable pipeline takes less than
$400 \mu secs$ to return a solution.

\smallskip
\textbf{Influence of the Preconditioner}
The last batch of experiments within this set is devoted to the proposed preconditioner in Section~\eqref{sec:essential-manifold}.
We generate random instances of the \RPp\xspace with the default parameters
and varying level of noise and number of correspondences.
The proposed certifiable pipeline is initialized with the \eightpt estimation
and the optimization is performed with and without the preconditioner.
\begin{figure}
\begin{subfigure}{0.49\textwidth}
     \centering
     \includegraphics[width=0.9\textwidth]{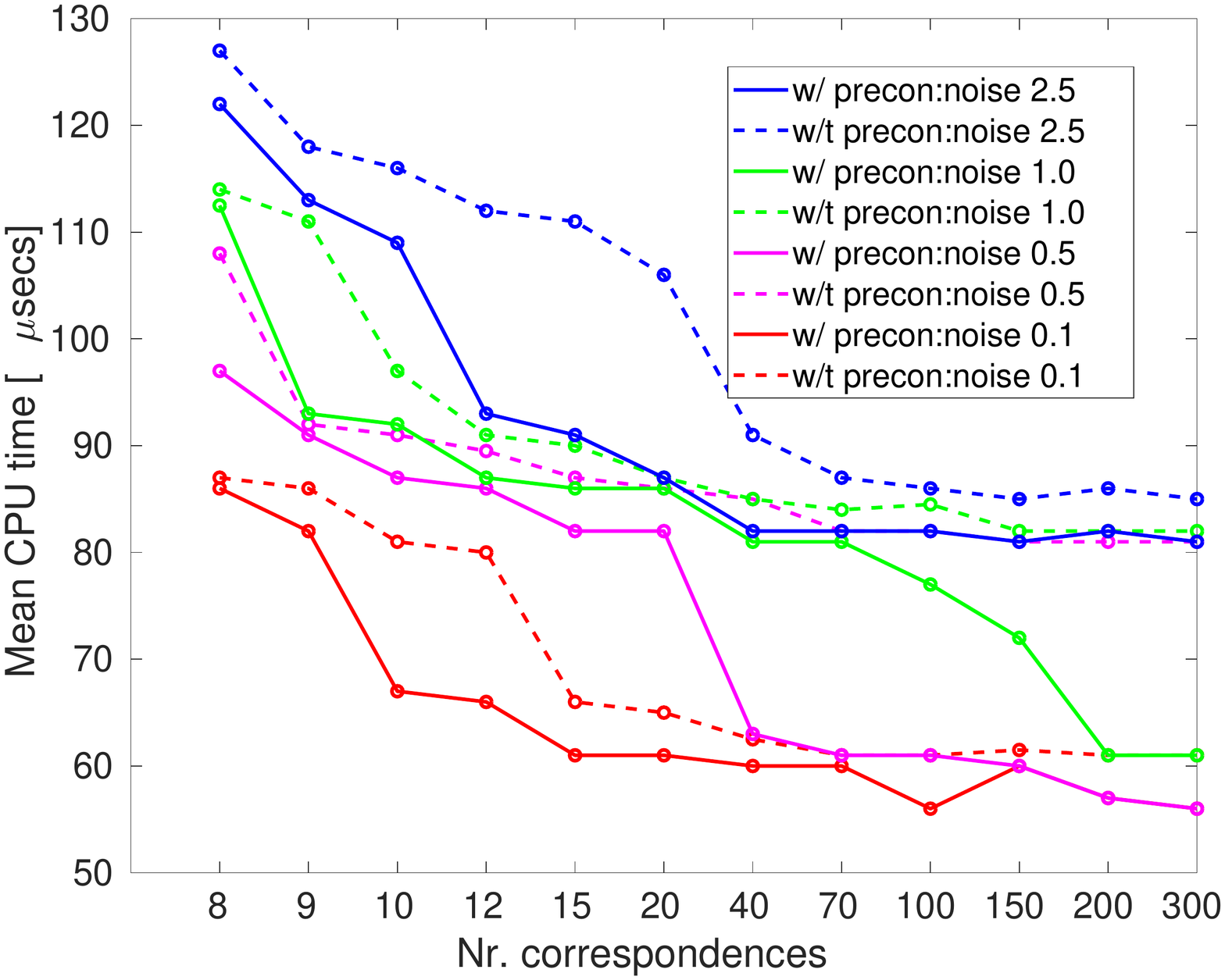}
     \caption{}
     \label{fig:time-with-precon}
\end{subfigure}
    \begin{subfigure}{0.49\textwidth}
     \centering
     \includegraphics[width=0.9\textwidth]{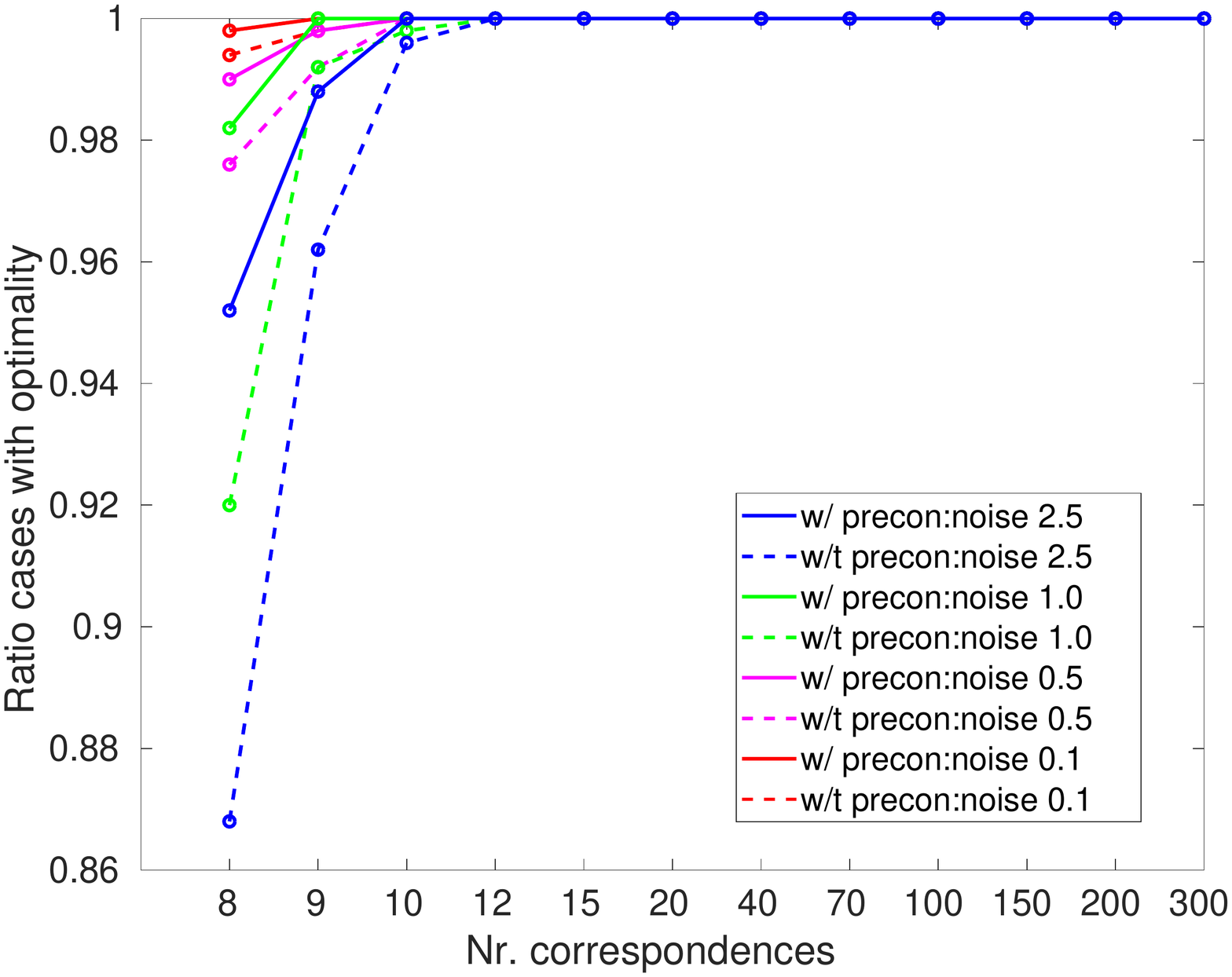}
     \caption{}
     \label{fig:opt-cases-precon}
\end{subfigure}
    \caption{Averaged computation time \eqref{fig:time-with-precon} required by the certifiable pipeline with (solid line) and without preconditioner (dashed line);
    and percentage of cases \eqref{fig:opt-cases-precon} in which we certify the solution as optimal with (solid line) and without (dashed line) preconditioner
    for different levels of noise.}
    \label{fig:times-precon}
\end{figure}

Figure~\eqref{fig:time-with-precon} depicts the averaged computation time in microseconds required by our proposal to produce a solution for each level of noise and number of correspondences.
In terms of computation time,
the main advantage of using this preconditioner is seen for those problems
with a large number of correspondences. 
It turns out, however, that its effect goes beyond its main purpose and
it is also reflected in the percentage of optimal solutions,
as it is shown in
Figure~\eqref{fig:opt-cases-precon} for those problems with less than 12 correspondences.

\begin{figure}[h]
    \begin{subfigure}{0.49\textwidth}
     \centering
     \includegraphics[width=0.9\textwidth]{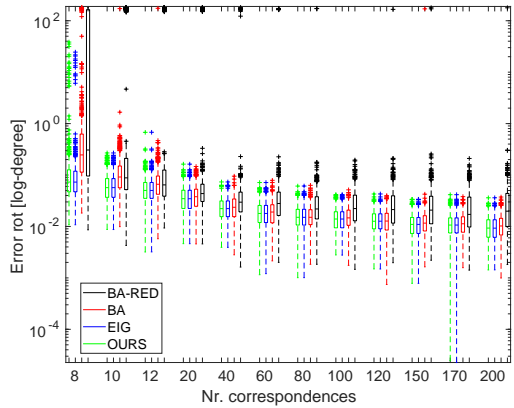}
     \caption{}
     \label{fig:error-rot-methods}
\end{subfigure}
\begin{subfigure}{0.49\textwidth}
     \centering
     \includegraphics[width=0.9\textwidth]{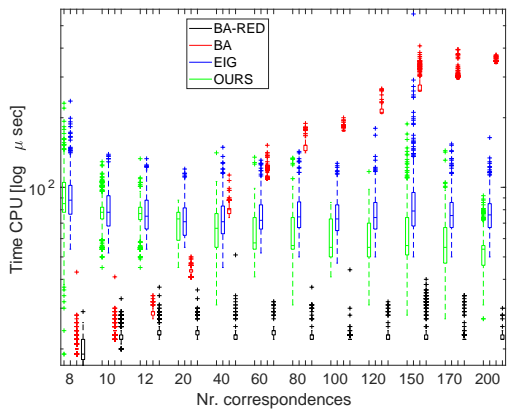}
     \caption{}
     \label{fig:time-methods}
\end{subfigure}
    \caption{Error in rotation in (log) \degrees~\eqref{fig:error-rot-methods};
    and computation time \eqref{fig:time-methods} in (log) $\mu secs$
    for our proposed pipeline \OURS,
    the eigen-formulation in \cite{kneip2013direct} \EIG,
    the standard non-linear optimization \BA;
    and its reduced version with only 10 points \BARED.}
    \label{fig:comparison-methods}
\end{figure}

\medskip 
\textbf{Computational cost}: 
Here we break down the computational cost 
of the certifiable pipeline in 
Section~\eqref{sec:proposed-pipeline} 
by considering its three main steps:
(1) initialization; (2) refinement; and (3) certification. 
The first stage depends on 
which solver is used for the initialization; 
for the \eightpt  
we measure less than $60\ \mu secs$ 
in average, considering also 
the data matrix $\matC$ generation 
and the computation of the preconditioner. 
The refinement stage 
depends on the quality of the initial guess, 
which a priori cannot be measured. 
We observe, however, 
that with the \eightpt \xspace 
as initialization, 
this stage takes less than $100\ \mu secs$ 
despite noise, number of correspondences and \FOV. 
Last, we consider the two main steps 
for the certification: 
(a) computing the dual candidates 
(solving equation~\eqref{eq:systeminlambdahat}); 
and 
(b) checking the spectral condition 
of the blocks of the Hessian $\HessE, \HessT$. 
Both steps take an average of $40 \ \mu secs$ 
for relaxation. 
In practice, two certifiers 
are able to certify most of the solutions. 
Considering this case, the computational time 
of the certification stage rise to $150 \ \mu secs$. 
Therefore, the computational cost 
of the complete certifiable pipeline 
tends to remain below  $300 \  \mu secs$.

\bigskip
\subsubsection{Comparison with state-of-the-art estimators without outliers} \label{sec:exp-synt-comparison}
In these experiments, we compare the proposed certifiable pipeline with the state-of-the-art non-minimal methods.
We denote by \EIG~the estimator proposed in \cite{kneip2013direct},
by \BA~the non-linear iterative minimization of the reprojection error
and by \BARED~the same non-linear optimization, but restricted to ten random correspondences.
We employ the implementation for these methods as it is provided by the computer vision library \opengv~\cite{kneip2014opengv}.
Recall that none of these methods is able to certify optimality.

We generate random instances with the default parameters given above and for each combination, we generate 500 instances.
In this case, the number of correspondences are 
$N = \{8, 10, 12, 20, 40, 60, 80, 100, 120, 150, 170, 200   \}$. 
We initialize all the algorithms with the same point $(\rot, \trans)$ obtained from the \eightpt.

Figure~\eqref{fig:error-rot-methods} shows the rotation error in logarithmic scale,
measured as the geodesic distance between the estimated rotation and the ground truth.
Note that \EIG~and \OURS~have similar error, of the order of \BA.
Figure~\eqref{fig:time-methods} depicts the computation time required by the optimization stage
in log-microseconds for the different methods.

While the error attained by \OURS,~\EIG~and \BA~are similar for problem instances
with $N \geq 20$,
Figure~\eqref{fig:time-methods} shows that \OURS~required less time to achieve
the same solution than the other two.
Since \BARED~only employs 10 correspondences,
it requires less CPU time although attains the largest error among all the methods.

\begin{figure}[h]
    \begin{subfigure}{0.49\textwidth}
     \centering
     \includegraphics[width=0.9\textwidth]{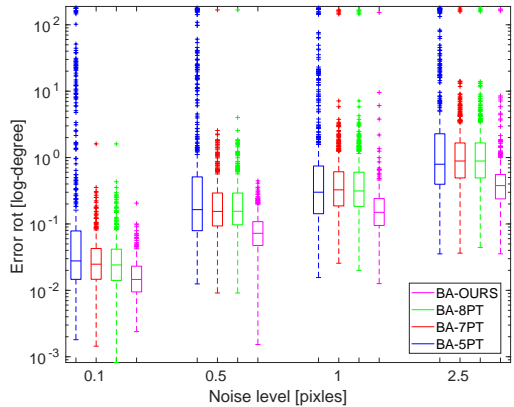}   \caption{}
     \label{fig:error-rot-BA-inits}
\end{subfigure}
\begin{subfigure}{0.49\textwidth}
     \centering
     \includegraphics[width=0.9\textwidth]{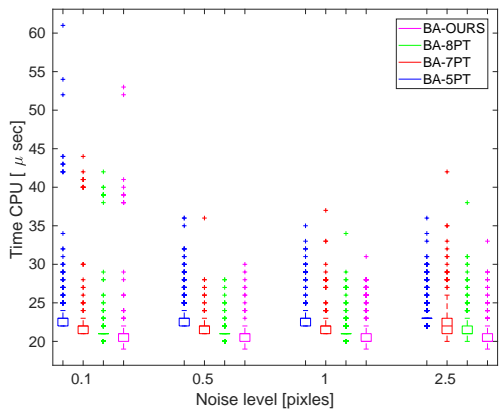}     \caption{}
     \label{fig:time-BA-inits}
\end{subfigure}
    \caption{Error in rotation in (log) \degrees~\eqref{fig:error-rot-methods};
    and computation time \eqref{fig:time-methods} in $\mu secs$
    for Bundle-adjustment initialized with
    the \fivept, the \sevenpt, the \eightpt and our proposed pipeline \OURS.}
    \label{fig:comparison-BA-inits}
\end{figure}

\subsubsection{Influence in Bundle-Adjustment} \label{sec:exp-synt-BA}
In this set of experiments we compare the performance,
both in error and time,
of the non-linear BA initialized with the previous minimal solvers and our proposal, 
since these minimal solvers are usually employed as initialization.
We compare the \fivept~(denoted as BA-5PT), the \sevenpt~(BA-7PT) and 
the \eightpt~(BA-8PT)
with our proposal (BA-OURS).
We generate random problem instances with the default parameters and vary the level of noise.
We fix the number of correspondences in $9$ and
feed each method with their minimal number of points
(our approach is executed with the $9$ points).
The multiple solutions are disambiguate with the remaining points.
We create 500 random problem instances for each level of noise.

Figure~\eqref{fig:error-rot-BA-inits} depicts the error in rotation 
(log-\degrees) \wrt the ground truth
for each solution returned by BA with the different initialization.
While the \sevenpt~and \eightpt~perform similarly,
\OURS~attains the lowest error despite the noise.
In terms of time, Figure~\eqref{fig:time-BA-inits} 
shows the computation time required
by BA (in log $\mu secs$) for all the algorithms.
Similarly, \sevenpt and \eightpt require comparable times, 
while \OURS~is faster.
The computational times reported in this set of experiments 
did not include the initialization stage, 
only the time required by the BA to converge. 
We observe that in general our proposed pipeline increase the 
accuracy of BA and reduce the time to covergence.

\bigskip
\subsubsection{Performance of the robust certifiable pipeline and comparison with state-of-the-art methods with outliers} \label{sec:exp-synt-GNC}

This last set of experiments on synthetic data 
is aimed to show the resiliency of our 
robust proposed pipeline 
(Algorithm~\eqref{alg:GNC}) 
in Section~\eqref{sec:GNC} against outliers: 
we apply the algorithm 
with the Tukey's biweight 
(\textsc{Tukey}) 
and Geman-McClure 
(\textsc{GM}) 
loss functions.
We also compare it with the non-iterative solvers, 
\ie \fivept,~\sevenpt~and \eightpt,
embedded in a RANSAC framework.

We generate random problem instances with the default parameters.
In order to help the methods,
we employ $200$ points and increase the \FOV~up to $150$~\degrees.
We introduce outliers by substituting the correspondence associated with the second frame
by a random (unitary) vector.
We increase the percentage of outliers with a fixed step of $5~\%$ and
for each of them, we generate $100$ random problem instances.

\begin{figure}
    \begin{subfigure}{0.49\textwidth}
     \centering
     \includegraphics[width=0.9\textwidth]{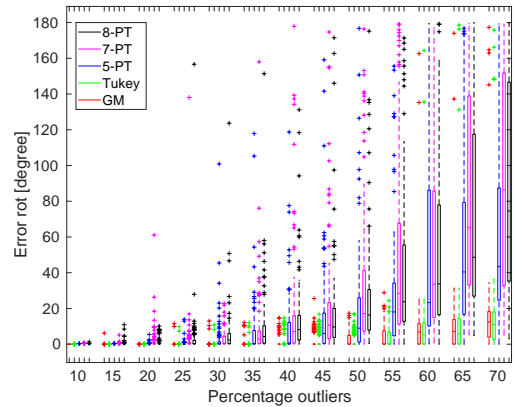}
     \caption{}
     \label{fig:outliers-result-error}
\end{subfigure}
\begin{subfigure}{0.49\textwidth}
     \centering
     \includegraphics[width=0.9\textwidth]{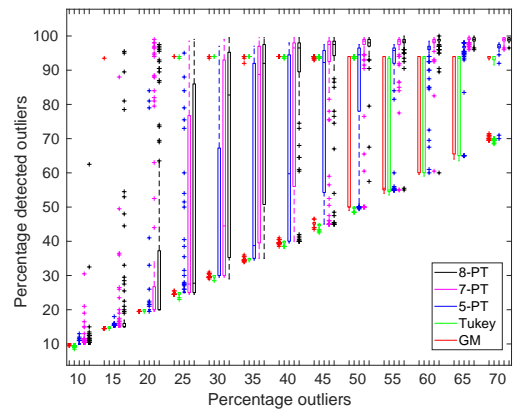}
     \caption{}
     \label{fig:outliers-result-inliers}
\end{subfigure}
    \caption{Error in rotation in \degrees~\eqref{fig:outliers-result-error};
    and  percentage of detected inliers \eqref{fig:outliers-result-inliers}
    for the different robust methods compared.}
    \label{fig:outliers-result}
\end{figure}

Figure~\eqref{fig:outliers-result-error} shows the error in rotation (\degrees)
for each compared method.
While the approaches based in RANSAC fail with $30\%$ of outliers
(the one with \fivept~performs better than \sevenpt~and \eightpt~though),
the proposed approach is able to return a solution with low error 
up to $50\%$ of bad correspondences.
Further, up to $70\%$ of outliers (high ratio of outliers),
the error attained by our proposal remains below $40~\degrees$
(with a mean value of $11~\degrees$),
while the RANSAC paradigms go up to $180~\degrees$ 
(minimum mean value is $62~\degrees$).

Figure~\eqref{fig:outliers-result-inliers} shows
the percentage of detected outliers (Y-axis) for each method as a function of the percentage of generated outliers (X-axis).
These results provide an explanation on the errors depicted in \eqref{fig:outliers-result-error}:
when the number of detected inliers decrease
(see \eg our method with more than $55\%$ outliers),
the solution returned attain highest errors.

\smallskip
\textbf{Computational cost}: 
Last, we summarize here the computational cost 
of Algorithm~\eqref{alg:GNC} 
and compare it against the average values 
for the RANSAC paradigms. 
The proposed algorithm 
requires $60-80$ outer iterations 
and $3-4$ inner iterations 
for any rate of outliers 
(except for $0\%$ of outliers, 
case in which it only runs one iteration) 
and the set of parameters employed. 
The relative pose estimation 
is the slowest step in the Algorithm, 
taking an average of $300~\mu secs$. 
In total, the robust pipeline requires 
$20-30~msecs$ to estimate a solution. 
On the other hand, RANSAC-based schemes 
take an average of $1-2~msecs$, 
being slower when the \fivept\xspace 
is used as minimal solver 
and/or 
when the number of outliers is large.

\subsection{Experiments on Real Data} \label{sec:exp-real}
To conclude this experimental validation, 
we evaluate our robust certifiable proposal 
(Algorithm~\eqref{alg:GNC}) on real data. 
To cover a wide regimen of real scenarios  
we select three available datasets, 
with both indoor and outdoor scenes, 
ground truth 
and intrinsic calibration parameters: 
\ETH~\cite{schops2017multi}, 
\TUM~\cite{sturm12iros} and 
\CVPR~\cite{strecha2008benchmarking}. 
The sequences employed here are listed in 
the~\suppl~(Section~(D)). 
For all of them, 
we generate the correspondences 
by extracting and matching 100 SURF features~\cite{bay2006surf}. 
We compute the equivalent unitary bearing vectors 
by applying the pin-hole camera model 
with the provided intrinsic parameters 
for each frame.
For each pair of images, we run the minimal solvers 
(\fivept, \sevenpt and \eightpt) 
embedded into a RANSAC scheme 
(\FIVER, \SEVENR, \EIGHTR) 
and our proposed robust pipeline (\OURS).

\begin{figure}[h]
    \centering
    \includegraphics[width=0.9\linewidth]{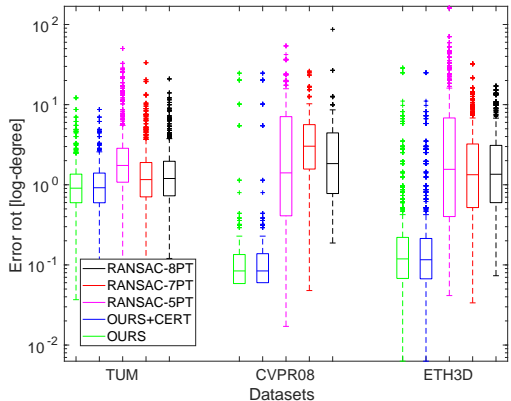}
    \caption{Error in rotation (log scale) for the different datasets in 
    which we evaluate the performance of our robust proposal in 
    Algorithm~\eqref{alg:GNC}, denoted as \textsc{OURS} and the 
    minimal solvers embedded into RANSAC paradigms: 
    \fivept~(\textsc{RANSAC-5PT}), 
    \sevenpt~(\textsc{RANSAC-7PT}) and 
    \eightpt~(\textsc{RANSAC-8PT}). 
    We also show the error in rotation for those cases in which we could 
    certify optimality (\textsc{OURS-CERT}). 
    }
    \label{fig:exp-real-error}
\end{figure}

Figure~\eqref{fig:exp-real-error} depicts the error in rotation 
(in terms of geodesic distance) 
for all the tested methods. 
For our proposal, we also include the error 
for those solutions that 
were certified as optimal (\OURSCERT). 
We observe how our robust pipeline 
achieves better results in general 
than the most employed robust solvers. 
See also how the optimal solutions 
tend to attain the lower errors. 
However, an optimal solution does not 
guarantee that all the outliers are discarded: 
the certification is performed for a concrete problem instance 
and solution. 
Last, 
the percentage of certified optimal solutions for each dataset 
were: 
\ETH~$98.5379\%$, 
\TUM~$82.6201\%$ and 
\CVPR~$99.0741\%$. 
We break down the percentage for each sequence 
and provide the error in rotation 
in the 
\suppl~(E).

\section{Conclusions and Future Work}
In this work we extended our previous work in~\cite{garcia2020certifiable} 
and proposed an efficient and 
robust certifiable relative pose estimation. 
We stated the \RPp\xspace as as optimization problem 
that minimizes the squared, normalized epipolar error 
over the set of normalized essential matrices. 
We provided a family of closed-form dual candidates 
derived from the six different relaxations of the 
set of essential matrices. 
These optimality certifiers 
were incorporated 
as final stage of an 
efficient, certifiable 
essential matrix estimation pipeline 
that first estimates 
a solution to the \RPp\xspace 
by iteratively refining on the product 
space of 3D rotations and 2-sphere 
an initial guess for this problem.  
We derived the Euclidean quadratic model 
of the \RPp \xspace 
required by the iterative solver. 
To speed up the convergence of the solver, 
we proposed a suitable preconditioner 
that also 
increased the number of certified solutions. 
Our proposal was proved through extensive experiments 
on synthetic data 
to be more accurate than 
the state-of-the-art closed-form methods, 
and faster than the iterative solvers. 
We showed that the gold standard approach 
for the \RPp, 
the 2-view Bundle-adjustment, 
converged faster and 
attained a lower error 
when initialized with the output of our algorithm 
in comparison with other suitable initializations.

We integrated our certifiable approach 
into the robust paradigm formed by the 
combination of Graduated Non-convexity 
and the Black-Rangarajan duality between 
robust and line process. 
Our robust pipeline was proved to be 
less sensible to high rates of outliers 
(up to $50\%$) 
in comparison with 
the common RANSAC approaches 
with the closed-form solvers. 
Last, our evaluation on real data 
confirmed the observations derived from the 
previous experiments, 
and more than $80\%$ of optimal solutions 
were certified for the three datasets considered. 
We made the code publicly available as 
an additional contribution, 
and also provided a generic library 
that implemented the robust paradigm here defined 
that was not restricted to the \RPp.

As future work, 
we consider the extension of our certifiable stage 
to consider other minimal characterizations of the 
set of essential matrices. 
Further enhancements on the verification algorithm 
are being considered for implementation.

\section*{Acknowledgements}
This work was supported by 
the grant program FPU18/01526 
funded by the Spanish Government 
and the research project WISER (DPI2017-84827-R).


\bibliographystyle{unsrt}

\newpage
\clearpage
\appendices

\onecolumn

\setcounter{MaxMatrixCols}{20}

\section{Jacobian of the constraints matrices} 
\label{app:sec:jac}

 In this Section we provide the Jacobian of the constraints, 
 denoted by $\matJ(\xVec)$ 
 for the family of relaxations considered in this work. 
 Each relaxation has an associated $12 \times 6$ Jacobian matrix 
 of full rank. 
 The explicit form of the Jacobian 
 associated with the $i$-th relaxation
 is obtained by dropping the 
 $(i+1)$-th column of the matrix 
 \begin{align}
    &\tilde{\matJ}(\xVec) = \nonumber\\
    &\begin{pmatrix}
    0  & \eVec_1  &\eVec_7/2 & 0 & 0 & 0 & \eVec_4/ 2\\
    0   &  0 & 0 & \eVec_4 & \eVec_7/2 & 0 & \eVec_1/2\\
    0 & 0 & \eVec_1/2 & 0 & \eVec_4 / 2 & \eVec_7 & 0 \\
    0  & \eVec_2 &   \eVec_8/2 & 0 & 0 & 0 & \eVec_5/ 2\\
    0  & 0  & 0 & \eVec_5 &  \eVec_8/2 & 0 &  \eVec_2/2\\
    0  & 0   & \eVec_2/2 & 0 & \eVec_5 / 2 & \eVec_8 & 0\\
    0  & \eVec_3  &\eVec_9/2 & 0 & 0 & 0 & \eVec_6 / 2\\
    0  & 0 &  0 & \eVec_6 & \eVec_9/2 & 0 & \eVec_3/2\\
    0  &  0 &  \eVec_3/2 &  0 & \eVec_6 / 2 & \eVec_9 & 0\\
    \tVec_1  & 0 &   \tVec_3/ 2 & -\tVec_1 & 0 & -\tVec_1 & \tVec_2/ 2\\
    \tVec_2  & - \tVec_2  & 0 & 0 & \tVec_3 /2 & - \tVec_2 & \tVec_1/ 2\\
    \tVec_3  &  -\tVec_3 & \tVec_1/ 2 & -\tVec_3 &  \tVec_2/2 & 0 & 0 \\
    \end{pmatrix}
\end{align}

 Recall that the $i$-th relaxation 
 is obtained by dropping the 
 $(i+1)$-th constraint in $\SetEssentialMatrices$.

 For the relaxation chosen in the main manuscript 
 (we drop $h_7 \equiv \eVec_1^T \eVec_2 + t_1 t_2 = 0$), 
 the Jacobian that appears in 
 Equation~\eqref{eq:systeminlambdahat} has the form:  
 \begin{equation}
    \matJ (\xVec) \doteq 
    \begin{pmatrix}
    0  & \eVec_1  &\eVec_7/2 & 0 & 0 & 0 \\
    0  &  0 & 0 & \eVec_4 & \eVec_7/2 & 0 \\
    0  & 0  & \eVec_1/2 & 0 & \eVec_4 / 2 & \eVec_7 \\
    0 & \eVec_2 &   \eVec_8/2 & 0 & 0 & 0 \\
    0 & 0  & 0 & \eVec_5 &  \eVec_8/2 & 0 \\
    0 & 0   & \eVec_2/2 & 0 & \eVec_5 / 2 & \eVec_8 \\
    0 & \eVec_3  &\eVec_9/2 & 0 & 0 & 0 \\
    0 &  0 &  0 & \eVec_6 & \eVec_9/2 & 0 \\
    0 &  0  & \eVec_3/2 & 0 & \eVec_6 / 2 & \eVec_9 \\
    \tVec_1  & 0 &   \tVec_3/ 2 & -\tVec_1 & 0 & -\tVec_1 \\
     \tVec_2 &  - \tVec_2  & 0 & 0 & \tVec_3 /2 & - \tVec_2 \\
    \tVec_3 &  -\tVec_3 & \tVec_1/ 2 & -\tVec_3 &  \tVec_2/2 & 0 \\
    \end{pmatrix}
\end{equation}


\section{Expression of the cost function based on the epipolar error in terms of the rotation and translation components of the essential matrix} \label{app:explicit-expressions-coeff-euclidean}
In this Section we will first develop the explicit expressions for the matrices $\Mt$ and $\Mr$, 
showing that the objective functions in Theorem \eqref{th:eqexpressions} are all equivalent. 
Then, the relation between these and the previous works \cite{kneip2013direct} and \cite{briales2018certifiably} are explicitly derived. 
We want to point out that the equivalence between the cost functions 
was first given by Briales \etal in \cite{briales2018certifiably} 
by manipulating the algebraic error employed in each optimization problem.

Let us first recall the chain of equalities given in \eqref{th:eqexpressions}:
\begin{align}
        \frac{1}{2}\eVec^T \matC \eVec &= \frac{1}{2}\trans^T \Mt \trans = \label{eq:kneip}\\
        &= \frac{1}{2}\rVec^T \Mr \rVec \label{eq:rotation} 
\end{align}
and 
the expression given for the coefficient matrix $\matC$ 
\begin{equation}
    \boldsymbol{C} = \sum_{i=1} ^N   (\obsip\kron \boldsymbol{f}_i)(\obsip \kron \obsi)^T. 
    \label{eq:definitionC}
\end{equation}

Consider the definition of the essential matrix in \eqref{eq:Me:[t]xR} and the identity:
\begin{equation}
    \matY = \matC \matX \matB^T \Leftrightarrow \VEC{\matY} = (\matB \kron \matC) \VEC{\matX}, \label{eq:kronequ}
\end{equation}
where all the matrices are of compatible dimension. We can now express $\Essential = \cross{t} \rot$ in two different useful forms:
\begin{align}
    \VEC{\Essential} &= (\iden{3} \kron \cross{t}) \VEC{\rot} \label{eq:vecr}\\
    \VEC{\Essential} &= (\rot^T \kron \iden{3}) \vCross{\trans} \label{eq:veccrosst}
\end{align}

\subsection{Objective cost as a function of the translation component.}

Let us define the matrix $\matB \in \Reals{9 \times 3}$ such that:
\begin{equation}
    \vCross{\trans} = \matB \trans. \label{eq:matrixB}
\end{equation}
We can now introduce \eqref{eq:veccrosst} and \eqref{eq:matrixB} into the original cost function:

\begin{align}
& \vCross{\trans}^T (\rot^T \kron \iden{3})^T \matC (\rot^T \kron \iden{3}) \vCross{\trans} = \label{eq:orignalt}\\
 & = \trans^T \matB^T (\rot^T \kron \iden{3})^T\matC \underbrace{(\rot^T \kron \iden{3}) \matB}_{\Tilde{\rot}} \trans = \trans^T \underbrace{\Tilde{\rot}^T \matC \Tilde{\rot}}_{\Mt} \trans,
\end{align}

where $\Tilde{\rot}^T = -[ \cross{r_1} , \cross{r_2} , \cross{r_3} ] ^T \in \Reals{9 \times 3}$ with $\{\rVec_i\}_{i=1}^3$ the columns of $\rot$ and $\Mt \in \symm{3}$ by construction. 
The objective function $\frac{1}{2} \trans^T \Mt\trans$ is quadratic in $\rot$ and $\trans$ and allows us to derive the gradient and Hessian \wrt $\trans$ in a simpler way.

\smallskip
\textbf{Relation with the objective function in~\cite{kneip2013direct}}:

Some algebraic manipulation in \eqref{eq:orignalt} yields the same objective function employed by Kneip and Lynen in~\cite{kneip2013direct}, as follows:
\begin{align}
    & \vCross{\trans}^T (\rot^T \kron \iden{3})^T \matC (\rot^T \kron \iden{3}) \vCross{\trans} = \\
    & \vCross{\trans}^T (\rot^T \kron \iden{3})^T \big( \sum_{i=1}^N(\obsip\kron \obsi)  (\obsip \kron \obsi) ^T \big)   (\rot^T \kron \iden{3}) \vCross{\trans} = \\
    &\overset{(1)}{=} \sum_{i=1}^N( \vCross{\trans}^T (\rot \kron \iden{3}) ( \obsip\kron \obsi)  (\obsip \kron \obsi) ^T    (\rot \kron \iden{3})^T \vCross{\trans} ) = \\
    &\overset{(2)}{=}\sum_{i=1}^N( \vCross{\trans}^T (\rot \obsip \kron \obsi)   (\rot \obsip\kron \obsi) ^T \vCross{\trans} ) = \\
    &\overset{(3)}{=} \sum_{i=1}^N (\obsi^T \cross{t} (\rot \obsip))^T (\obsi^T \cross{t} (\rot \obsip)) = \\
    &\overset{(4)}{=} \sum_{i=1}^N \big (\obsi^T \rot \obsip^T \trans \big)^T \big(\obsi^T \vCross{\rot \obsip}^T \trans\big) = \\
    &\overset{(5)}{=} \sum_{i=1}^N \big( \vCross{\rot \obsip} \obsi )^T \trans\big)^T \big( \vCross{\rot \obsip}\obsi )^T \trans\big) = \\
    &\overset{(6)}{=} \sum_{i=1}^N \big(  (\obsi \times \rot \obsip)^T \trans\big)^T \big(  (\obsi \times \rot \obsip)^T \trans\big) = \\
    &\overset{(7)}{=} \trans^T \underbrace{\big( \sum_{i=1}^N (\obsi \times \rot \obsip)  (\obsi \times \rot \obsip)^T\big)}_{\Mt}  \trans, \label{eq:develmt}
\end{align}
which is the objective function to be minimized in~\cite{kneip2013direct}.

The performed steps are:
\begin{enumerate}
    \item Introduce the definition of $\matC$ given in \eqref{eq:definitionC}. 
    \item "Move" the constant terms inside the sum over $i$.
    \item Apply the identity: $(\matA \kron \matB)(\matC \kron \matD) = (\matA \matC \kron \matB \matD)$.
    \item Use the expression in \eqref{eq:kronequ} to obtain a single term.
    \item Consider that $\matB^t \matA ^t = (\matA \matB)^t$.
    \item Apply the anticommutative property of the cross product, \ie $\aVec \times \bVec = -\bVec \times \aVec$.
    \item Develop the transposed parenthesis and "move" outside the sum over $i$ the constant term $\trans$.
\end{enumerate}


\bigskip
\subsection{Objective function as a function of the rotation component.} 

Similarly, we first develop the form for $\Mr$
and then relate our results with those described by Briales \etal in \cite{briales2018certifiably}.

We introduce \eqref{eq:vecr} in our original cost function and use $\rot = \VEC{\rot}$. The matrix $\Mr$ is obtained as
\begin{equation}
    \rVec^T (\iden{3} \kron \cross{t}) ^T \matC \underbrace{(\iden{3} \kron \cross{t})}_{\Tilde{\matT}} \rVec =
    \rVec^T \underbrace{\Tilde{\matT} ^T \matC \Tilde{\matT}}_{\Mr} \rVec. \label{eq:matrixmr}
\end{equation}
Similarly, the cost function defined by $\Mr$ is quadratic in the entries of $\rot, \trans$ and yields a simple expression for the computation of the gradient and Hessian \wrt $\rot$.

\smallskip 
\textbf{Relation with the objective function in~\cite{briales2018certifiably}}:

Let us introduce the definition of $\matC$ in \eqref{eq:matrixmr} and manipulate the expression as follows:
\begin{align}
    & \rVec^T(\iden{3} \kron \cross{t}) ^T \big( \sum_{i=1}^N(\obsip \kron \obsi)  (\obsip \kron \obsi) ^T \big) (\iden{3} \kron \cross{t})\rVec = \\
    &\overset{(1)}{=} \rVec^T \big( \sum_{i=1}^N(\iden{3} \kron \cross{t}) ^T (\obsip\kron \obsi)  (\obsip \kron \obsi) ^T  (\iden{3} \kron \cross{t}) \big) \rVec = \\
    &\overset{(2)}{=}  \rVec^T \big( \sum_{i=1}^N(\obsip \kron \cross{t}^T \obsi) (\obsip \kron \cross{t}^T \obsi)^T \big) \rVec = \\
    &\overset{(3)}{=}  \rVec^T \underbrace{\big( \sum_{i=1}^N(\obsip \kron \cross{f_i} \trans) (\obsip \kron \cross{f_i} \trans)^T \big)}_{\Mr} \rVec = \label{eq:mrexp}\\
    &\overset{(4)}{=}  \rVec^T  \sum_{i=1}^N\Big( (\obsip \kron \cross{f_i} \sum_{j, k = 1}^3 \big(t_j \eVec_j)\big) (\obsip \kron \cross{f_i} \sum_{j, k = 1}^3 \big(t_k \eVec_k)\big) \Big) \rVec = \\
    &\overset{(5)}{=}  \rVec^T  \sum_{i=1}^N\Big( (\obsip \kron \sum_{j = 1}^3 \big( \cross{f_i} t_j \eVec_j)\big) (\obsip \kron \sum_{k = 1}^3 \big( \cross{f_i} t_k \eVec_k)\big) \Big) \rVec = \\
    &\overset{(6)}{=}   \rVec^T  \sum_{i=1}^N\Big( \sum_{j, k = 1}^3 \big((\obsip \kron \cross{f_i} t_j\eVec_j) (\obsip \kron \cross{f_i}t_k \eVec_k)^T \big)\Big) \rVec = \\
    &\overset{(7)}{=}  t_j \rVec^T  \sum_{j, k=1}^3 \underbrace{ \Big( \sum_{i = 1}^N \big(\obsip \kron (\obsi \times \eVec_j)) (\obsip \kron (\obsi \times \eVec_k))^T \Big)}_{\matC_{jk}} \rVec t_k, \label{eq:quadjesus}
\end{align}
which is the cost function employed in \cite{briales2018certifiably}.

The main performed step are:
\begin{enumerate}
    \item "Move" the constant terms in $\trans$ inside the sum over $i$.
    \item Apply the relation $(\matA \kron \matB)(\matC \kron \matD) = (\matA \matC \kron \matB \matD)$.
    \item Use the anticommutative property of the cross product. 
    \item The cross product with a 3D vector can be expressed as a matrix multiplication, \ie, it's linear: $\cross{a} \bVec = \cross{a} b_1  \eVec_1 + \cross{a} b_2 \eVec_2 + \cross{a} b_3 \eVec_3$, where $\eVec_i$ denotes the canonical vector in $\Reals{3}$.
    \item Apply the associative property of the kronecker product, \ie $\aVec \kron (\bVec + ... \matC) = \aVec \kron \bVec + ...+ \aVec \kron \matC$.
    \item "Move" the constant, scalar terms $t_j, t_k$ outside the sums. Finally, swap the order of the sums.
\end{enumerate}


\section{Terms of the quadratic model for the cost function in terms of the rotation and translation} \label{app:explicit-expressions-quadratic-model}

In this Section we derive the Euclidean gradient and Hessian of the cost function considered in this work \wrt the points on the manifold $\rotSphereSpace$.

Since the considered manifold is a direct product, its tangent bundle \TM~\cite{absil2009optimization} is expressed as: 
$\text{\TM} = \TanR \times \Tant$, where $\TanR$ and $\Tant$ 
are the tangent bundles of $\rotSpace$ and $\sphere$, respectively~\cite{ma2001optimization}. 
This implies that we can write any vector field $\dot{Y}$ 
as the composition of its components in the two subspaces 
$\TanR ,\Tant$ 
\ie, 
$\dot{Y} = (\dot{Y}_{\rotSpace}, \dot{Y}_{\sphere}) \in \TanR \times \Tant$, 
which simplifies the derivation of them. 

\smallskip
\textbf{Gradient}:

Considering the cost function as a function restricted to the embedded manifold \cite{absil2009optimization}, the relation between the Euclidean gradient and its Riemannian counterpart is simplify to:

\begin{equation}
    \RGradfRt  = \big(\ProjR(\gradfR), \Projt (\gradft)\big),
\end{equation}

where $\ProjR(\bullet)$ and $\Projt(\bullet)$ denote the orthogonal projection operator onto the tangent spaces of $\rotSpace$ and $\sphere$ at the points $\rot, \trans$, respectively \cite{absil2009optimization}:
\begin{align}
    &\ProjR: \TanR (\Reals{3 \times 3}) \longrightarrow \TanR (\rotSpace) \\
    & \ProjR(\matX) = \rot \text{skew}(\rot^T \matX), 
\end{align}
where $\text{skew}(\matA)$ extracts the skew-symmetric part of the matrix $\matA$, \ie $\text{skew}(\matA) = \frac{1}{2} ( \matA - \matA^T)$  and
\begin{align}
    &\Projt: \Tant (\Reals{3}) \longrightarrow  \Tant (\sphere) \\
    & \Projt(\xVec) = \xVec - (\trans^T \xVec) \trans
\end{align}

The Euclidean gradient $\gradfRt = (\gradfR, \gradft) $ reads:
\begin{equation}
    \gradfR = \Mr{r} \quad \gradft = \Mt \trans.
\end{equation}

\smallskip 
\textbf{Hessian-vector product}:

Similarly, the Hessian counterparts are derived by considering the embedded manifold:
\begin{align}
    \RHessR (\rot,\trans) [\VR,\Vt] &= \ProjR \Big(  \HessfR[\VR, \Vt] - \VR^T \text{sym} \big ( \rot \gradfR \big ) \Big) \\
    \RHesst (\rot, \trans) [\VR, \Vt] &= \Projt \Big(  \Hessft [\VR, \Vt]\big) - \big (\trans^T \gradft \big )\Vt \Big),
\end{align}
where $\text{sym}(\matA)$ extracts the symmetric part of the matrix $\matA$ \ie, $\text{sym}(\matA) = \frac{1}{2} (\matA + \matA^T)$.

The action of the Euclidean Hessian on the vector ($\VR, \Vt$) is calculated by applying multivariate calculus as
\begin{equation}
    \HessfRt [\VR, \Vt] = (\HessfR[\VR, \Vt], \Hessft [\VR, \Vt]).
\end{equation}

The component for the rotation is computed as:
\begin{align}
    \HessfR[\VR, \Vt] &= \HessRR [\VR] + \HessRt[ \trans] \\
    \HessRR [\VR] &= \Mr \Vr  \\
    \HessRt[ \trans] &= \Mtr \Vt,
\end{align}
where $\MtrRT \doteq \MtrRT$ will be defined later.

The component for the translation is:
\begin{align}
    \Hessft [\VR, \Vt] &= \Hesstt[\Vt] + \HesstR [ \VR] \\
    \Hesstt [\Vt] &= \Mt \Vt \\
    \HesstR  [\Vt] &= \Mrt \VR,
\end{align}
with $ \Mrt \doteq \MrtRT$ and $\MrtRT = \MtrRT^T$ by symmetry of the Hessian.

\smallskip 
\textbf{Derivation of $\MrtRT$}

See that 
\begin{align}
    \Mrt = \frac{\partial (\partial \fRt)}{ \partial \rVec \partial \trans} = \frac{\partial (\Mt \trans)}{ \partial \rVec} \in \Reals{3 \times 9},
\end{align}
and 
\begin{equation}
    \Mt \trans = \Tilde{\rot}^T \matC \Tilde{\rot} \trans =   \Tilde{\rot}^T \matC \Tilde{\matT} \rVec \in \Reals{3},
\end{equation}
where we have used the definition of $\eVec$ given in~\eqref{eq:veccrosst} 
for the second equality 
(in the first one we substituted the definition of $\Mt$ in []). 

Recall that $\Tilde{\rot}^T \in \Reals{3 \times 9}$ 
and let $\Tilde{\rVec}_i \in \Reals{1 \times 9}$ be the $i$-th row of the matrix, 
\ie,
\begin{align}
    \Tilde{\rot}^T = 
    \begin{pmatrix}
    \Tilde{\rVec}_1 \\
    \Tilde{\rVec}_2 \\
    \Tilde{\rVec}_3.
    \end{pmatrix}
\end{align}

See that we can express each row as $\Tilde{\rVec}_i = \rVec^T \matB_i$, 
where $\matB_i \in \Reals{9 \times 9}$ are sparse matrices 
(shown in Equation~\eqref{eq:Mrt-sparse-matrices}). 
Then, the $i$-th row of $\Mtr$ is given by 
\begin{equation}
    \Tilde{\rVec}_i \matC \Tilde{\matT} \rVec = \rVec^T \matB_i \matC \Tilde{\matT} \rVec,
\end{equation}
and its derivate \wrt $\rVec$ is $2 \matB_i \matC \Tilde{\matT} \rVec \in \Reals{9 \times 1}$. 
The gradient $\Mrt$ is obtained by stacking the derivative of each row.

\smallskip
A similar procedure can be given for $ \Mtr \in \Reals{9 \times 3}$, 
by defining a set of matrices $\matA_i \in \Reals{3 \times 9}, \ i=1, \dots 9$ 
such that the $i$-th row of $\Tilde{\matT} \in \Reals{9 \times 9}$, $\Tilde{\trans}_1 \in \Reals{1 \times 9}$ can be expressed as $\Tilde{\trans}_1 = \trans^T \matA_i$. 

Then, 
\begin{equation}
    \Mtr = \frac{\partial (\Mr \rVec)}{\partial \trans},
\end{equation}
with 
\begin{equation}
    \Mr \rVec = \Tilde{\matT}^T \matC \Tilde{\matT} \rVec = \Tilde{\matT}^T \matC \Tilde{\rot} \trans.
\end{equation}

Its $i$-th row is given by $\trans^T \matA_i \matC \Tilde{\rot} \trans$ 
and its derivative \wrt $\trans$ by 
$2 \matA_i \matC \Tilde{\rot} \trans \in \Reals{3}$. 
Again, $\Mtr$ is obtained by stacking all the derivatives 
(nine in total) 
of the rows. 
We do not include the explicit expressions for $\matA_i$ 
given their similarity with $\matB_i$ 
(Equation~\eqref{eq:Mrt-sparse-matrices}).

Although it may not seem apparent, the expression of $\Mrt, \Mtr$ given above 
fulfills the relation $\Mtr = \Mrt^T$.

\begin{equation} 
 \hspace*{-1.5cm}
    \matB_1 = 
    \begin{pmatrix}
    0 & 0 & 0 & 0 & 0 & 0 & 0 & 0 & 0 \\
    0 & 0 & -1 & 0 & 0 & 0 & 0 & 0 & 0 \\
    0 & 1 & 0 & 0 & 0 & 0 & 0 & 0 & 0 \\
    0 & 0 & 0 & 0 & 0 & 0 & 0 & 0 & 0 \\
    0 & 0 & 0 & 0 & 0 & -1 & 0 & 0 & 0 \\
    0 & 0 & 0 & 0 & 1 & 0 & 0 & 0 & 0 \\
    0 & 0 & 0 & 0 & 0 & 0 & 0 & 0 & 0 \\
    0 & 0 & 0 & 0 & 0 & 0 & 0 & 0 & -1 \\
    0 & 0 & 0 & 0 & 0 & 0 & 0 & 1 & 0 \\
    \end{pmatrix} 
    , 
    \matB_2  = 
    \begin{pmatrix}
    0 & 0 & 1 & 0 & 0 & 0 & 0 & 0 & 0 \\
    0 & 0 & 0 & 0 & 0 & 0 & 0 & 0 & 0  \\
    -1 & 0 & 0 & 0 & 0 & 0 & 0 & 0 & 0 \\
    0 & 0 & 0 & 0 & 0 & 1 & 0 & 0 & 0 \\
    0 & 0 & 0 & 0 & 0 & 0 & 0 & 0 & 0 \\
    0 & 0 & 0 & -1 & 0 & 0 & 0 & 0 & 0 \\
    0 & 0 & 0 & 0 & 0 & 0 & 0 & 0 & 1 \\
    0 & 0 & 0 & 0 & 0 & 0 & 0 & 0 & 0 \\
    0 & 0 & 0 & 0 & 0 & 0 & -1 & 0 & 0 \\
    \end{pmatrix} 
    \matB_3  = 
    \begin{pmatrix}
    0 & -1 & 0 & 0 & 0 & 0 & 0 & 0 & 0 \\
    1 & 0 & 0 & 0 & 0 & 0 & 0 & 0 & 0  \\
    0 & 0 & 0 & 0 & 0 & 0 & 0 & 0 & 0 \\
    0 & 0 & 0 & 0 & -1 & 0 & 0 & 0 & 0 \\
    0 & 0 & 0 & 1 & 0 & 0 & 0 & 0 & 0 \\
    0 & 0 & 0 & 0 & 0 & 0 & 0 & 0 & 0 \\
    0 & 0 & 0 & 0 & 0 & 0 & 0 & -1 & 0 \\
    0 & 0 & 0 & 0 & 0 & 0 & 1 & 0 & 0 \\
    0 & 0 & 0 & 0 & 0 & 0 & 0 & 0 & 0 \\
    \end{pmatrix} 
    \label{eq:Mrt-sparse-matrices}
\end{equation}

\section{Experiments on Real Data: Datasets} \label{app:real-data-list}

The experiments in Section~\eqref{sec:exp-real} were performed in the following sequences: 

\begin{itemize}
    \item \textbf{ETH3D}: playground; relief; botanical\_garden; boulders; exhibition\_hall; living\_room; observatory; terrace; terrace\_2; terrains; statue.
    \item \textbf{CVPR08}: Herz-Jesu-P8; castle-P19; castle-P30; fountain-P11;  Herz-Jesu-P25; entry-P10; .
    \item \textbf{TUM}: rgbd\_dataset\_freiburg3\_long\_office\_household; rgbd\_dataset\_freiburg3\_structure\_texture\_far; \newline
    rgbd\_dataset\_freiburg3\_structure\_texture\_near; 
    rgbd\_dataset\_freiburg3\_large\_cabinet.
\end{itemize}

\section{Experiments on Real Data: Results} \label{app:real-data-perc} 

In this Section we break down the general 
statistics reported in the main document. 
Concretely, we show for each sequence of each dataset 
(\ETH, \TUM, \CVPR) 
the error in rotation measured in terms of geodesic distance 
\wrt the ground-truth 
and list the percentage of cases in which 
our verification in Algorithm~\eqref{alg:certification}
could certify the optimality of the solution. 

\begin{itemize}
    \item \textbf{ETH3D}: 
    playground ($100\%$); relief ($92\%$); botanical\_garden ($100\%$); 
    boulders ($100\%$); 
    exhibition\_hall ($94\%$); living\_room ($97.9167\%$); 
    observatory ($100\%$); terrace ($100\%$); terrace\_2 ($100\%$); 
    terrains ($100\%$); statue ($100\%$).
    \item \textbf{CVPR08}: Herz-Jesu-P8 ($100\%$); castle-P19 ($94.44\%$); 
    castle-P30 ($100\%$); fountain-P11 ($100\%$);  Herz-Jesu-P25 ($100\%$); 
    entry-P10 ($100\%$); .
    \item \textbf{TUM}: rgbd\_dataset\_freiburg3\_long\_office\_household ($98.7179 \%$); rgbd\_dataset\_freiburg3\_structure\_texture\_far ($75.8065\%$);  \newline
    rgbd\_dataset\_freiburg3\_structure\_texture\_near ($74.1379\%$); 
    rgbd\_dataset\_freiburg3\_large\_cabinet ($81.8182\%$).
\end{itemize}

\begin{figure}
         \centering
         \includegraphics[width=0.9\textwidth]{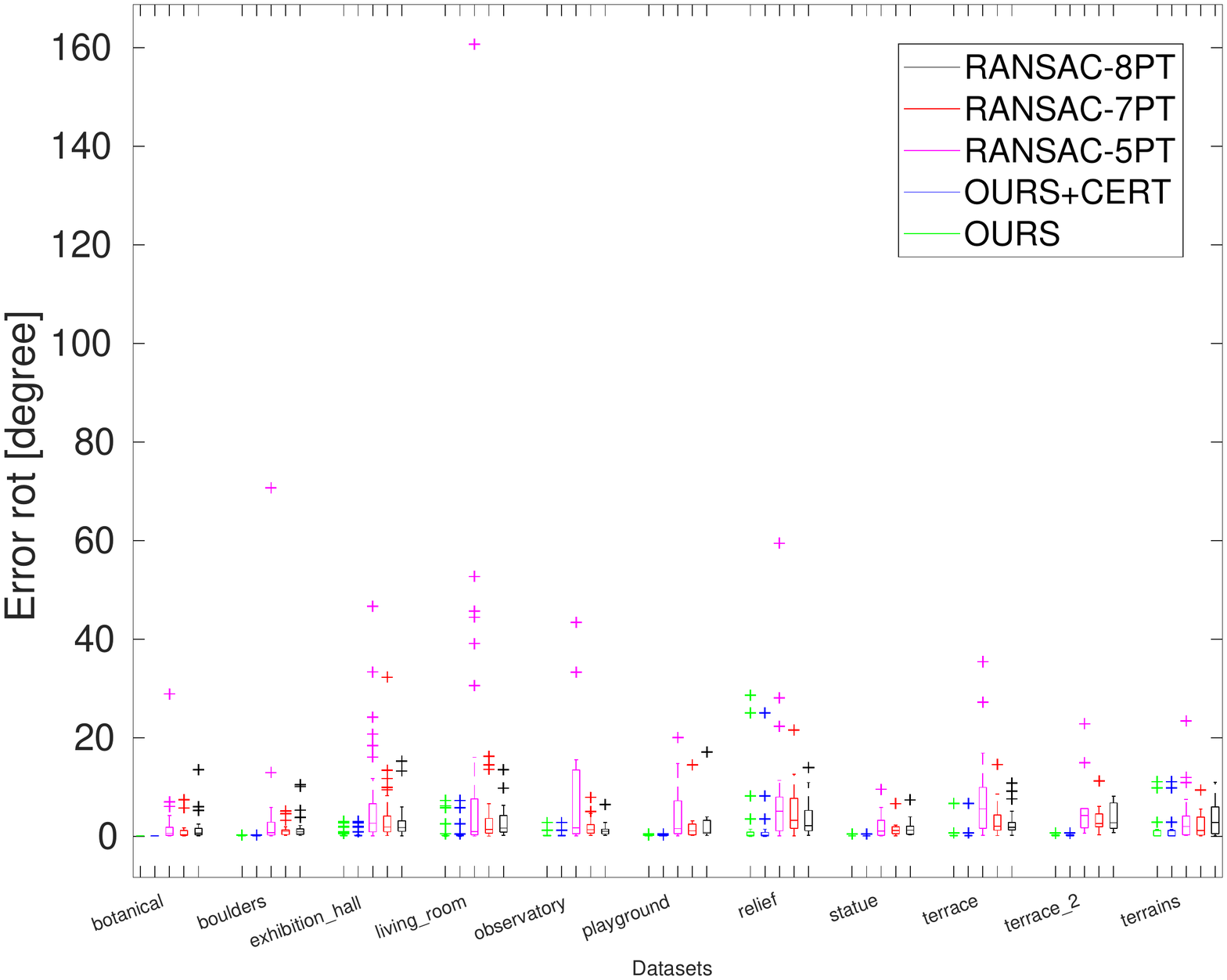}
         \caption{}
     \label{fig:app:rot-real-eth3d}
    \caption{Error in rotation \wrt the ground-truth 
    for the different sequences of the \ETH~dataset.}
\end{figure}

\begin{figure}
         \centering
         \includegraphics[width=0.9\textwidth]{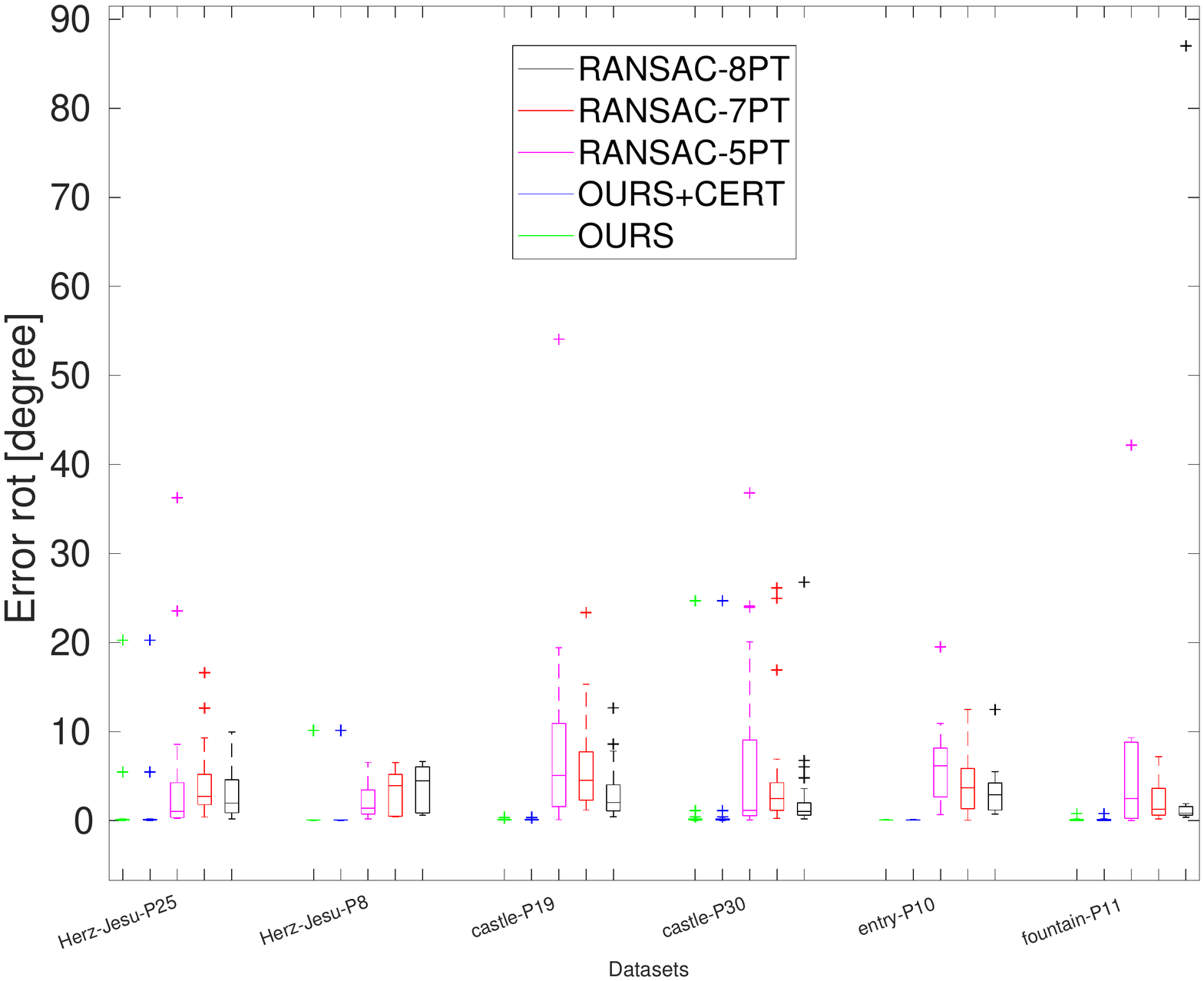}
         \caption{}
         \label{fig:app:rot-real-cvpr08}
    \caption{Error in rotation \wrt the ground-truth 
    for the different sequences of the \CVPR~dataset.}
\end{figure}

\begin{figure}
    
         \centering
         \includegraphics[width=0.9\textwidth]{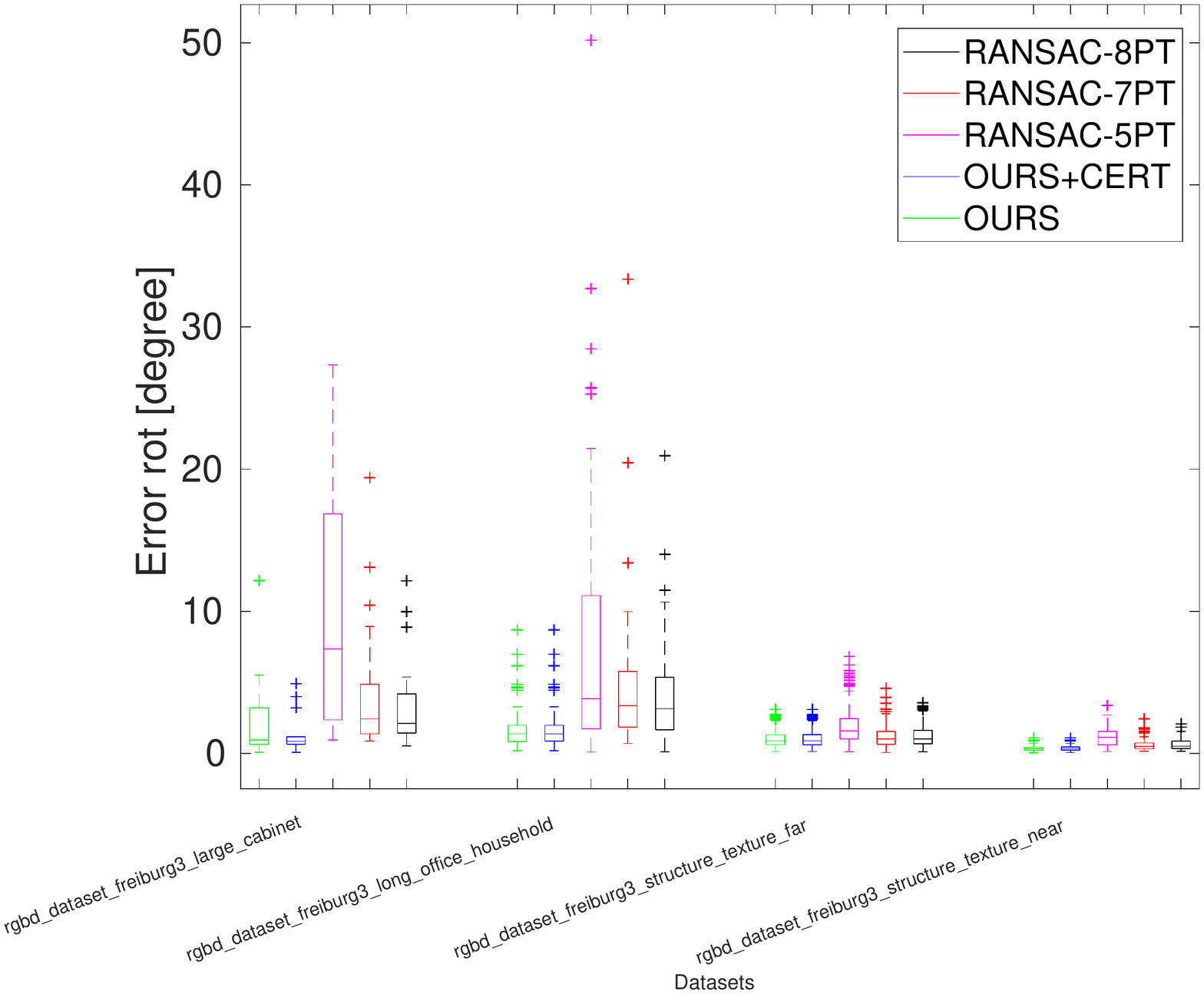}
         \caption{}
         \label{fig:app:rot-real-tum}
    \caption{Error in rotation \wrt the ground-truth 
    for the different sequences of the \TUM~dataset.}
\end{figure}

Figure~\eqref{fig:app:rot-real-eth3d}, \eqref{fig:app:rot-real-cvpr08} and 
\eqref{fig:app:rot-real-tum} depicts the error in rotation for 
the proposed robust pipeline in ALgorithm~\eqref{alg:GNC} (\OURS), 
the error for those cases with certifier optimal solution (\OURSCERT), 
and the minimal solvers (\fivept, \sevenpt and \eightpt) 
embedded into RANSAC paradigms 
(\FIVER, \SEVENR and \EIGHTR, respectively)

\end{document}